\documentclass{article}

\usepackage[margin=1.25in]{geometry}
\usepackage[utf8]{inputenc} % allow utf-8 input
\usepackage[T1]{fontenc}    % use 8-bit T1 fonts
\usepackage{url}            % simple URL typesetting
\usepackage{booktabs}       % professional-quality tables
\usepackage{amsfonts}       % blackboard math symbols
\usepackage{nicefrac}       % compact symbols for 1/2, etc.
\usepackage{microtype}      % microtypography
\usepackage[dvipsnames]{xcolor}

\usepackage{listings}
\usepackage{booktabs}
\usepackage[pdftex]{graphicx}
\usepackage{bbm}
\usepackage{pdflscape}
\usepackage{multirow}

\usepackage{amsmath, amsthm, amssymb}
\usepackage{pifont}
\usepackage[export]{adjustbox}

\usepackage[numbers,compress]{natbib}

\usepackage[small]{caption}
\usepackage[pdftex]{hyperref}
\usepackage{mathtools}

\usepackage{subcaption}    % subfigure is deprecated

\usepackage{xspace}
\usepackage{enumitem}

\definecolor{DarkGreen}{rgb}{0.1,0.5,0.1}
\definecolor{DarkRed}{rgb}{0.5,0.1,0.1}
\definecolor{DarkBlue}{rgb}{0.1,0.1,0.5}
\definecolor{Gray}{rgb}{0.2,0.2,0.2}

\hypersetup{
    unicode=false,          % non-Latin characters in Acrobat bookmarks
    pdftoolbar=true,        % show Acrobat toolbar?
    pdfmenubar=true,        % show Acrobat menu?
    pdffitwindow=false,      % page fit to window when opened
    pdfnewwindow=true,      % links in new window
    colorlinks=true,       % false: boxed links; true: colored links
    linkcolor=DarkBlue,          % color of internal links
    citecolor=DarkGreen,        % color of links to bibliography
    filecolor=DarkRed,      % color of file links
    urlcolor=DarkBlue,          % color of external links
    pdftitle={},
    pdfauthor={},
    pdfkeywords={language model, llm, calibration, benchmark, folktexts, datasets}, % list of keywords
}

\definecolor{codegreen}{rgb}{0,0.6,0}
\definecolor{codegray}{rgb}{0.5,0.5,0.5}
\definecolor{codepurple}{rgb}{0.58,0,0.82}
\definecolor{backcolor}{rgb}{0.95,0.95,0.92}
\definecolor{greyish}{rgb}{0.45,0.45,0.35}

\lstdefinestyle{mystyle}{
    backgroundcolor=\color{backcolor},   
    commentstyle=\color{codegreen},
    keywordstyle=\color{magenta},
    numberstyle=\small\color{codegray},
    stringstyle=\color{codepurple},
    basicstyle=\ttfamily\footnotesize,
    breakatwhitespace=false,         
    breaklines=true,                 
    captionpos=b,                    
    keepspaces=true,                 
    numbers=left,
    numbersep=5pt,                  
    showspaces=false,                
    showstringspaces=false,
    showtabs=false,                  
    tabsize=2,
    keepspaces=true, % keeps spaces in the code
    columns=fullflexible % adjusts spacing between characters    
}
\newcommand{\cX}{\mathcal{X}}

\newcommand{\cP}{\mathcal{P}}

\renewcommand{\Pr}{\mathbb{P}}

\newcommand{\folktexts}{{\tt folktexts}\xspace}

\usepackage{pythonhighlight}

\usepackage{dsfont}
\newcommand{\ind}{\mathds{1}}
\DeclareMathOperator{\E}{\mathbb{E}}

\usepackage{colortbl}

\usepackage{tcolorbox}

\usepackage[framemethod=tikz]{mdframed}

\definecolor{light-gray}{gray}{0.95}
\surroundwithmdframed[
  hidealllines=true,
  backgroundcolor=light-gray,
  innerleftmargin=15pt,
  innertopmargin=0pt,
  innerbottommargin=0pt]{lstlisting}

\usepackage[toc,page]{appendix}

\title{Evaluating language models as risk scores}

\usepackage{authblk}

\author[1,3]{Andr\'{e} F. Cruz\thanks{Corresponding author: {\tt andre.cruz@tuebingen.mpg.de}}}
\author[1,3]{{Moritz Hardt}}
\author[1,2,3]{Celestine Mendler-D\"{u}nner}

\affil[1]{Max Planck Institute for Intelligent Systems, T\"{u}bingen}
\affil[2]{ELLIS Institute Tübingen}
\affil[3]{Tübingen AI Center}

\date{}
\begin{document}

\maketitle

\begin{abstract}Current question-answering benchmarks predominantly focus on accuracy in realizable prediction tasks.
Conditioned on a question and answer-key, does the most likely token match the ground truth?
Such benchmarks necessarily fail to evaluate LLMs' ability to quantify ground-truth outcome uncertainty.
In this work, we focus on the use of LLMs as risk scores for unrealizable prediction tasks.
We introduce \folktexts, a software package to systematically generate risk scores using LLMs, and evaluate them against US Census data products.
A flexible API enables the use of different prompting schemes, local or web-hosted models, and diverse census columns that can be used to compose custom prediction tasks.
%
% We demonstrate the utility of \folktexts through a sweep of empirical insights into the statistical properties of 17 recent LLMs across five natural text benchmark tasks.
%
We evaluate 17 recent LLMs across five proposed benchmark tasks.
We find that zero-shot risk scores produced by multiple-choice question-answering have high predictive signal but are widely miscalibrated.
Base models consistently overestimate outcome uncertainty, while instruction-tuned models underestimate uncertainty and produce over-confident risk scores.
In fact, instruction-tuning polarizes answer distribution regardless of true underlying data uncertainty.
This reveals a general inability of instruction-tuned LLMs to express data uncertainty using multiple-choice answers.
% Additionally, while LLM predictive power reliably improves with model size, calibration does not.
%
A separate experiment using verbalized chat-style risk queries yields substantially improved calibration across instruction-tuned models.
%
% While standard realizable benchmarks can only reveal \textit{model uncertainty}, ...
These differences in ability to quantify data uncertainty cannot be revealed in realizable settings, and highlight a blind-spot in the current evaluation ecosystem that \folktexts covers.\looseness-1
\end{abstract}

%% Main paper body

\section{Introduction}

Fueled by the success of large language models (LLMs), it is increasingly tempting for practitioners to use such models for risk assessment and decision making in consequential domains~\cite{tamkin2023evaluating, kasneci2023chatgpt, thirunavukarasu2023large, gaebler2024auditing}. Given the CV of a job applicant, for example, some might prompt a model, what are the chances that the employee will perform well on the job? The true answer is likely uncertain. Some applicants of the same features will do well, others won’t. A good statistical model should faithfully reflect such outcome uncertainty.%~\looseness-1

Calibration is perhaps the most basic kind of uncertainty quantification to ask for. A calibrated model, on average, reflects the true frequency of outcomes in a population. Calibrated models must therefore give at least some indication of uncertainty. Fundamental to statistical practice across the board, calibration has also been a central component in the debate around the ethics and fairness of consequential risk scoring in recent years~\cite{chouldechova2017fair, pleiss2017fairness, corbett2023measure, barocas-hardt-narayanan}.
 
The evaluation of LLMs to date, however, has predominantly focused on accuracy metrics, often in realizable tasks where there is a unique correct label for each data point. Such benchmarks necessarily cannot speak to the use of language models as risk score estimators. A model can have high utility in well-defined question-answering tasks while being wildly miscalibrated.
In fact, 
while accuracy corresponds to knowledge over the expected answer, proper uncertainty quantification corresponds to knowledge over the variance over answers.
%
% Indeed, growing empirical evidence suggests that instruction-tuned language models tend to be over-confident in the probability that they assign to their preferred answer~\cite{mielke22reducing,achiam2023gpt}.

\begin{figure}[t!]
\centering
\includegraphics[width=\textwidth]{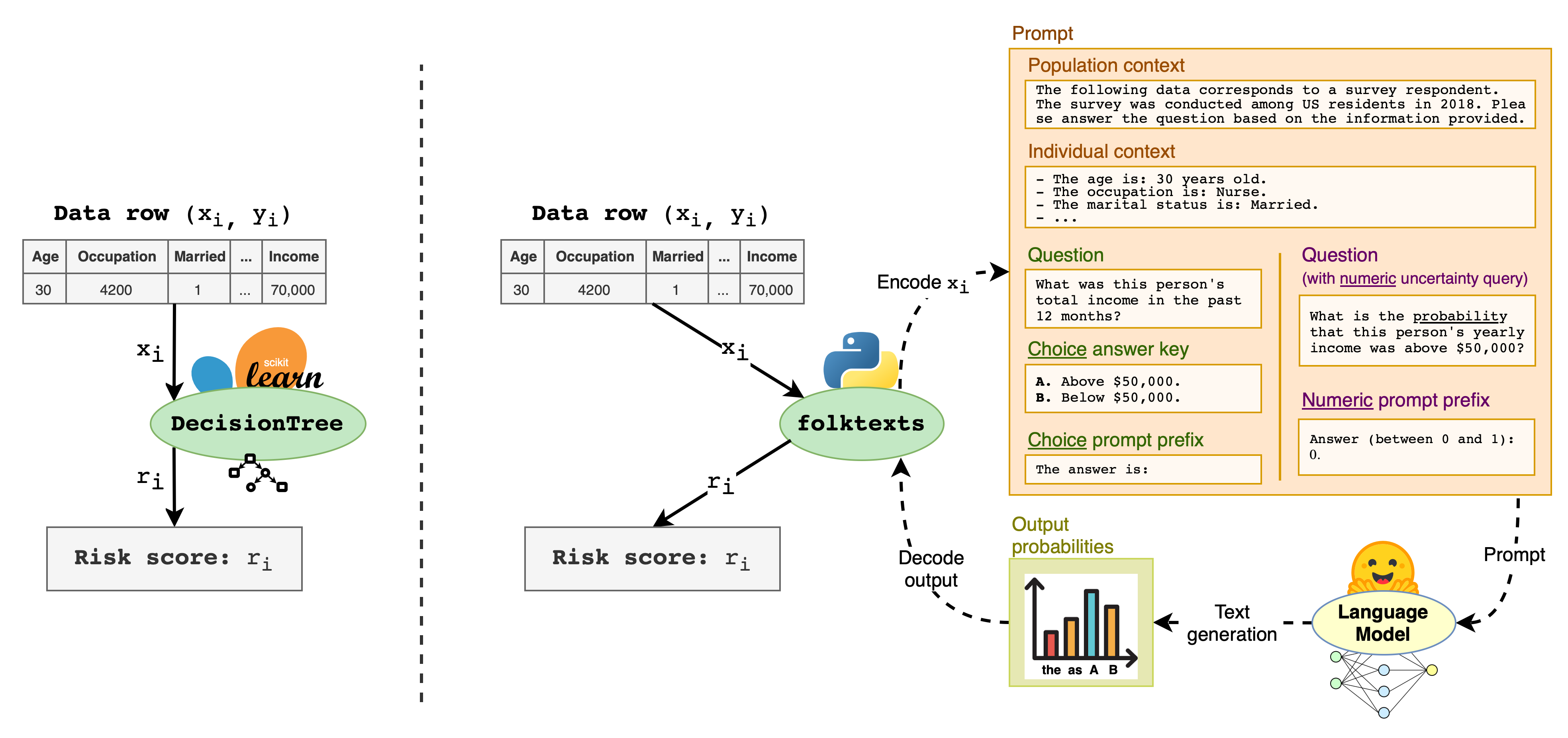}
\caption{Information flow from tabular data to risk scores, using a supervised classifier (\textit{left}) or a language model (\textit{right}). The \folktexts package maps language models to the traditional machine learning workflow.
}
\label{fig:folktexts_diagram}
\end{figure}

\subsection{Our contributions}

We contribute an open-source software package, called \folktexts,\footnote{\label{foot:implementation_url}Package and results available at: \url{https://github.com/socialfoundations/folktexts}} that provides datasets and tools to evaluate statistical properties of LLMs as risk scorers.
We show-case its functionalities with a sweep of new empirical insights on the risk scores produced by 17 recently proposed LLMs.~\looseness-1

The \folktexts package offers a systematic way to translate between the natural language interface and the standard machine learning type signature. It translates prediction tasks defined by numeric features $X$ and labels $Y$ into natural text prompts and extracts risk scores $R$ from LLMs. This opens up a rich repertoire of open-source libraries, benchmarks and evaluation tools to study their statistical properties.
Figure~\ref{fig:folktexts_diagram} illustrates the workflow for producing risk scores using LLMs.

% Classical metrics and tools for evaluating statistical properties of machine learning models rely on a standard type signature of numeric features $X$, labels $Y$, and risk scores $R$. As language models interface with the world strictly through natural text, they are incompatible with a rich repertoire of open-source libraries, benchmarks and evaluation tools.
% %
% The \folktexts package is designed to bridge this gap and offer a systematic way to translate between the natural language interface and the standard $(X, Y, R)$ statistical machine learning type signature.
% Figure~\ref{fig:folktexts_diagram} illustrates the workflow for producing risk scores using LLMs.

For benchmarking risk scores, we need ground truth samples from a known probability distribution.
Inspired by the popular folktables package~\citep{ding2021retiring}, 
\folktexts builds on US Census data products, specifically, the American Community Survey (ACS)~\citep{pums2018}, collecting information about more than 3.2 million individuals representative of the US population.
{\tt Folktexts} systematically constructs prediction tasks and prompts from the individual survey responses using the US Census codebook, and the ACS questionnaire as a reference.
Risk scores are extracted from the language model's output token probabilities using a standard question-answering interface.
The package offers five pre-specified question-answering benchmark tasks that are ready-to-use with any language model.
A flexible API allows for a variety of natural language tasks to be constructed out of 28 census features whose values are mapped to prompt-completion pairs (features detailed in Table~\ref{tab:col_to_text}).
% 
% These tasks can be customized through a flexible API where any census data attribute can be used either as a feature or as the target variable in a prediction task (see Table~\ref{tab:col_to_text} for the initial set of 28 mapped attributes).
Furthermore, evaluations can easily be performed over subgroups of the population to conduct algorithmic fairness audits.\looseness-1

\paragraph{Empirical insights.}
We contribute a sweep of empirical findings based on our package. We evaluate 17 recently proposed LLMs, with sizes ranging from 2B parameters to 141B parameters.
Our study demonstrates how inspecting risk scores of LLMs on underspecified prediction tasks reveals new insights that cannot be deduced from inspecting accuracy alone. The main findings are summarized as follows: 
\begin{itemize}
\item Models' output token probabilities have strong predictive signal\footnote{We measure predictive signal using the area under the receiver operating characteristic curve (AUC)~\citep{peterson1954theory}.} but are widely miscalibrated.

\item The failure modes of models are different: Multiple-choice answer probabilities generated by base models consistently overestimate outcome uncertainty, while instruction-tuned models underestimate uncertainty and produce over-confident risk scores.

\item Using a verbalized chat-style prompt results in materially different answer distributions, with significantly improved calibration for instruction-tuned models, accompanied by a small but consistent decrease in predictive power.

\item Instruction-tuning generally worsens calibration of multiple-choice answers, but improves calibration of verbalized numeric answers.

% \item As a result instruction-tuning worsens calibration error on tasks with high outcome uncertainty, despite generally improving predictive signal and accuracy.

\iffalse
\item Some models produce risk scores that are consistently biased in terms of protected categories. For example, among instruction-tuned models on the income-prediction task (where $Y=1$ is high income), we find that 
% high-income probability is consistently underestimated for the same risk score, 
$\Pr\left\{ Y=1 \mid R=r, S=\text{`Black'} \right\} > \Pr\left\{ Y=1 \mid R=r, S\neq\text{`Black'} \right\}$, where $r\in[0,1]$ is the predicted high-income score, and~$S$ denotes the census encoding of \emph{race}.
%
In words, the model is miscalibrated and underestimates high income within the Black population.~\looseness-1
\fi

\end{itemize}

We hope our package facilitates future investigations into statistical properties of LLMs, not only as a way to faithfully reflect real-world uncertainties, but also as a required stepping stone to trustworthy model responses.

\paragraph{Outline.} In Section~\ref{sec:background} we provide necessary background on risk scores and calibration in statistical machine learning. In Section~\ref{sec:llm-as-classifier} we extend this background to the application to language models, providing various design choices around prompting templates and ways to extract risk scores from language models. 
In Section~\ref{sec:results} we evaluate 17 recent LLMs on 5 proposed benchmark tasks and summarize empirical findings.

\subsection{Limitations}
\label{sec:limitations}

Predictive modeling and statistical risk scoring in consequential settings is a matter of active debate. Numerous scholars have cautioned us about the dangers of statistical risk scoring and, in particular, the potential of risk scores to harm marginalized and vulnerable populations~\cite{chouldechova2017fair,pasquale2015black,eubanks2018automating,benjamin2019race}.
Our evaluation suite is intended to help in identifying one potential problem with language models for risk assessment, specifically, their inability to faithfully represent outcome uncertainty. However, our metrics are not intended to be sufficient criteria for the use of LLMs in consequential risk assessment applications. The fact that a model is calibrated says little about the potential impact it might have when used as a risk score. Numerous works in the algorithmic fairness literature, for example, have discussed the limitations of calibration as a fairness metric, see, e.g.,~\cite{corbett2023measure,barocas-hardt-narayanan,kasy2021fairness}. There is also significant work on the limitations of statistical tools for predicting future outcomes and making decisions based on these predictions~\cite{perdomo2023difficult, perdomo2024relative, wang2024against}. Calibration cannot and does not address these limitations.

\subsection{Related work}

The use of LLMs for decision-making has seen increasing interest as of late.
\citet{hegselmann23tabllm} show that a {\tt Bigscience T0 11B} model~\citep{sanh2021multitask} surpasses the predictive performance (AUC) of supervised learning baselines in the very-few-samples regime.
The authors find that fine-tuning an LLM outperforms fitting a standard statistical model on a variety of tasks % (from 768 to 48K samples) 
up to training set sizes in the 10s to 100s of samples.
Related work by~\citet{tabletSlack23} shows that providing task-specific expert knowledge via instructions in context can lead to significant improvements in model predictive power.
\citet{tamkin2023evaluating} generate hypothetical individual information for a variety of decision scenarios, and analyze how language models' outputs change when provided different demographic data.
Some attributes are found to positively affect the model's decision (e.g., higher chance of approving a small business loan for minorities) while others affect it negatively (e.g., lower chance for older aged individuals).

A separate research thread considers how to leverage LLMs to model human population statistics. %, namely by predicting survey responses of different individuals.
\citet{Argyle_2023} evaluate whether GPT-3 can faithfully reproduce political party preferences for different US subpopulations, and conclude that model outputs can accurately reflect a variety of correlations between demographics and political preferences.
Other works use a similar methodology to model the distribution of human opinions on different domains~\citep{sanders2023demonstrations,horton2023large,brand2023using}.
\citet{aher2023using,dillion2023can} use LLMs to reproduce popular psychology experiments on human moral judgments, and confirm there's good alignment between human answers and LLM outputs.
%
% However, work by \citet{dominguez2023questioning} reveals a variety of biases in using multiple-choice LLM answers, and assert that `alignment' with different groups is purely determined by that group's entropy.
%
Literature on modelling human population statistics generally focuses on using LLMs to obtain accurate survey completions.
That is, given demographic information on an individual, what was their response to a specific survey question?
This methodology often ignores the fact that individuals described by the same set of demographic features will realistically give different answers.
An accurate model would not only provide the highest likelihood answer, but also a measure of uncertainty corresponding to the expected variability within a sub-population.
%
% In fact, 
% this measure of uncertainty is itself a type of knowledge, specifically, knowledge over the variance of the answer distribution.
%
Our work tackles this arguably neglected research avenue: Analyzing LLM risk scores instead of discrete token answers, and whether they are accurate and calibrated to human populations.

% Related work by \citet{tamkin2023evaluating} uses a similar tabular data set-up to study potential biases of the closed-source {\tt Claude 2.0} model.
% The authors generate hypothetical individual information for a variety of decision scenarios, and find that sensitive demographic information (such as age, race and gender) is actively used by the model for decision-making.
% Some attributes affect the decision positively (e.g., higher chance of approving a small business loan for minorities) while others affect it negatively (e.g., lower chance for older aged individuals).

% It is unclear to which degree using synthetic or unrealistic data can affect the studies' validity.
% Our proposed benchmarks enable similar studies to be conducted using real-world Census data.
%
% Moreover, related literature generally focuses on studying accurate class predictions by specific language models. We study over a dozen state-of-the-art language models, and discern which trends are actually consistent across all models, and which are constrained to a single model or model family.
% Further, we focus on a particularly neglected research avenue: analyzing LLM risk scores instead of discrete token answers, and whether they are accurate and calibrated.

\paragraph{Calibration.}
Calibration is a widely studied concept in the literature on forecasting in statistics and econometrics with a venerable history~\citep{brier50score, murphy1977reliability, degroot1983comparison, platt1999probabilistic,zadrozny2001obtaining}. % dawid1982well,gneiting07forecasts,cox1958two,foster98cal,niculescu2005predicting
Recent years have seen a surge in interest in calibration in the context of deep learning~\cite{guo2017calibration}.
Calibration of LLMs has been studied on diverse question-answering benchmarks, ranging from sentiment classification, knowledge testing, and mathematical reasoning to multi-task benchmarks~\citep[e.g.,][]{kadavath2022language,xiao22plm,jiang21cali,xiong2024can,chen2023close,hendrycks2021measuring, zellers2019hellaswag,clark2018think,cobbe2021training, zhang2024study}.
What differentiates our work is that we study calibration in naturally underspecified prediction tasks. This requires even the most accurate models to accurately reflect non-trivial probabilities over outcomes to be calibrated. 

% Practical constraints of evaluation, such as black-box access have been studied~\cite{xiong2024can}.
% There is a consensus that climbing the leader board on classical accuracy-driven  benchmarks~\citep{rajpurkar2016squad, wang2018glue, reddy2019coqa,ling2017program,srivastava2023beyond} is not sufficient to guarantee improvements in calibration. 

% Several works found that LLMs tend to be over-confident when asked to select the true answer among candidates, especially after instruction-tuning~\cite{mielke22reducing,achiam2023gpt}. Meaning, that they tend to assign high probability to whatever answer they generate. As a result, such models need to be accurate to be calibrated. To counter this source of potential confounding, we move away from realizable prediction tasks and suggest to assess calibration on under-specified questions with inherent uncertainty in the outcomes. This requires even the most accurate models to accurately reflect non-trivial probabilities over outcomes to be calibrated, beyond expressing high confidence.

To systematically construct such prediction tasks we resort to survey data.
Surveys have a long tradition in social science research as a tool for gathering statistical information about the characteristics and opinions of human populations~\cite{groves2009survey}. Survey data comes with carefully curated questionnaires, as well as ground truth data. The value of this rich data source for model evaluation has not remained unrecognized. Surveys have recently gained attention to study bias and alignment of LLMs~\citep{Argyle_2023,santurkar23opinion,durmus2023towards,dominguez2023questioning,scherrer23belief}, and inspecting systematic biases in multiple-choice responses~\cite{tjuatja2024llms}. Instead of using surveys to get insights about a models natural inclinations, we use them to test model calibration with respect to a given population.

Beyond task-calibration, calibration at the word and token-level has been explored in the context of language generation~\cite[e.g.,][]{ott2018analyzing}.  Others have focused on connections to hallucination~\cite{kalai2024calibrated}, and expressing uncertainty in natural language \cite{lin2022teaching,mielke22ling,band2024linguistic}. Related to our work, \citet{lin2024generating} emphasize the inherent outcome uncertainty in language generation. Focusing on generation at the word or token level poses the challenge of measuring a high-dimensional probability distribution. Considering  binary classification tasks has the practical advantage of circumventing this problem.

Calibration has also played a major role in the algorithmic fairness literature. Group-wise calibration has been proposed as a fairness criterion since the 1960s~\citep{cleary1968test,hutchinson201950}. In particular, it's been a notion central to an active debate about the fairness of risk scores in consequential decision making~\citep{chouldechova2017fair,pleiss2017fairness,corbett2023measure,barocas-hardt-narayanan}.
A recent line of work originating in algorithmic fairness studies \textit{multicalibration} as a strengthening of calibration~\citep{hebert18multical}.

% Recently, \cite{ye2024benchmarking} propose an uncertainty-aware evaluation metric, which rewards models for high confidence in correct answers. The authors demonstrate that taking probabilities into account can change the order in the leader board. Thus, climbing the leader board on current benchmarks does not necessarily improve model calibration. This emphasizes the need to add calibration to model evaluation.

% \cite{kadavath2022language} suggest that models are well calibrated on diverse multiple choice questions when asked to choose the correct answer.

\section{Preliminaries}
\label{sec:background}
%This section focuses on risk scores as the main point of study, and the 3 main metrics we derive from it (calibration, AUC, and accuracy)

This section provides necessary background on risk scores and the statistical evaluation of binary predictors.
Throughout, we assume a joint distribution~$\cP$ given by a pair of random variables $(X,Y)$ where $X$ is a set of features and $Y$ is the outcome to be predicted. In the applications we study, the features~$X$ typically form a text sequence and the outcome $Y$ is a discrete random variable that we would like to predict from the sequence. A parametric model $f_\theta(y|x)$ assigns a probability to each possible outcome $y$ given a feature vector $x$. The goal of a generative model is generally to approximate the conditional distribution~$\cP(y| x)$, where parameters are fit to a huge corpus of training data.~\looseness-1

We will focus on binary prediction throughout this work. Let $Y\in\{0,1\}$ be a random variable indicating the outcome of an event the learner wishes to predict. We use the shorthand  notation~$f_\theta(x)$ to denote the model's estimate of the probability that $Y=1$, given context $X=x$. Following standard terminology we will refer to $f_\theta:\cX\rightarrow [0,1]$ as \emph{score function}, and its output $f_\theta(x)$ as \emph{risk score}. There are various dimensions along which to evaluate risk scores by comparing them against samples of the reference population $\cP$.

\subsection{Calibration}
% Calibration is a basic aspect of uncertainty quantification.
We say a score function is \emph{calibrated} over a population $\cP$ if and only if for all  values~$r\in[0,1]$ with $\Pr\{f_\theta(X)=r\}>0$, we have %$\E[Y\mid f_\theta(X)=r]=r.$ Equivalently,
\begin{equation}
\Pr[ Y=1 \mid f_\theta(x) = r] = r. 
\label{eq:calibrated}
\end{equation}

% \[\mathrm{E}[ Y(x)=1 | \,\risk(x) = r] = r.\]
This condition asks that, over the set of all instances $x$ with score value $r$, an $r$ fraction of those instances must indeed have a positive label.
% This condition asks that, over the set of all instances $x$ that a model assigns a score value $r$, a fraction $r$ of instances indeed has positive outcome.
Importantly, calibration is defined with respect to a population, it does not measure a model's ability to discriminate between instances. A model that outputs the constant value~$\mu=\E[Y]$ on all instances is calibrated, by definition. In particular, we can always achieve calibration by setting $f_\theta(x)$ with the average value of $y$ among all instances~$x'$ in a given partition such that $f_\theta(x')=f_\theta(x).$

We use Expected Calibration Error (ECE) as the primary metric to empirically evaluate calibration. It is defined as the expected absolute difference between a classifier's confidence in its predictions and the accuracy on the same predictions. %, evaluated across bins of samples.
More formally, given $n$ triplets $(x_i, y_i, r_i)$ where $r_i$ denotes the model's score value for the corresponding data point $(x_i,y_i)$. The ECE is defined as 
\begin{equation}
\label{eq:ece}
    \mathrm{ECE} \coloneq \frac {1}{n} \sum_m  \left|\sum_{i\in B_m} y_i- \sum_{i\in B_m} r_i \right|,
\end{equation}
where data points are grouped into $M$ equally spaced bins $B_m$ according to their score values.
We use $M=10$ in our evaluation, which is a commonly used value~\citep{Pakdaman15cali}. %Since the number of bins can impact calibration measures~\cite{kumar19bins}, we expose it as a parameter in the implementation.  
We also provide a measure of ECE over quantile-based score bins, as well as Brier score~\citep{brier50score} to allow for a more complete picture.
Furthermore, we use reliability diagrams~\cite{degroot1983comparison} to aid a visual interpretation of calibration. These diagrams plot expected sample accuracy as a function of confidence.
Any deviation from a perfect diagonal represents miscalibration.

We can strengthen the calibration condition in \eqref{eq:calibrated} by requiring it in multiple subgroups of the population. Specifically, letting $G$ denote any discrete random variable, we can require the conditional calibration condition~$\Pr[ Y=1 \mid G=g\,,f_\theta(x) = r] = r$ for every setting~$g$ of the random variable~$G.$ This is often used to define fairness.

\subsection{Predictive performance}
%Florian's comment: Regarding the calibration vs accuracy section, there is quite a bit of literature on explicitly decomposing loss functions into calibration and some other property that might be worth mentioning 

When solving classification problems it's common practice to threshold risk scores to obtain a classifier. In our notation this corresponds to thresholding the risk scores $f_\theta(x)$:
\[c(x)=\ind\{f_\theta(x)>\tau\}.\] 
The classifier~$c$ that minimizes the misclassification error~$\E\ind\{c(X)\ne Y\}=\Pr\{c(X)\ne Y\}$ is given by~$c^*(x)=\ind\{f^*(x)>0.5\}$, where $f^*$ is the Bayes optimal scoring function. 
In the following, when we use \emph{accuracy} we refer to the fraction of correct predictions after thresholding. If not specified otherwise we use $\tau=0.5$. This corresponds to an argmax operator applied to the class probabilities. It is important to note that a classifier can achieve perfect accuracy even when derived from a suboptimal scoring function. Thus, accuracy alone provides an incomplete picture of a model's ability to express uncertainty.

\paragraph{Instance ranking.} The area under the receiver operating characteristic curve (AUC) is a rank-based measure of predictive model performance. It measures the probability that a randomly chosen positive observation ($Y=1$) will have a higher score than a randomly chosen negative observation ($Y=0$).
A high AUC value in no way reflects accurate or calibrated probability estimates, it relates only to the signal-to-noise ratio in the risk scoring function~\citep{peterson1954theory}.
Using the AUC metric allows us to neatly separate risk score calibration from their predictive signal, although both are crucial for accurate class predictions.

%Since there is always a calibrated constant score, calibration does not imply any non-trivial guarantee about accuracy. %The converse is true under some conditions. For example, the non-parametric model~$f$ that minimizes the squared loss $\E(f(X)-Y)^2$ is given by the conditional expectation~$f^*(x)=\E[Y\mid X=x].$ In fact, this model satisfies calibration in a strong sense: The score~$f^*$ is calibrated on any subdomain $\cX'\subset \cX$. This does not, in general, follow from the basic calibration property in~\eqref{eq:calibrated}. 

\iffalse
\paragraph{Extension to multiclass problems.}
Our discussion extends to the case where the target variable~$Y$ takes on more than two values. In this case, we assume that the model~$f_\theta(y|x)$ specifies a probability for each value $y$ given the input $x.$ We assume that the model is normalized so that $\sum_y f_\theta(y|x)=1.$ Equation~\ref{eq:calibrated} then corresponds to the condition
\[
\Pr\{Y=y\mid f_\theta(y|X)=r\} = r
\]
for all values~$y$ and~$r$ such that $\Pr\{f_\theta(y|X)=r\}>0.$ Calibration in the multiclass setting follows from optimality again. For example, if the model~$f^*$ minimizes the negative log likelihood, then $f^*(y|x)=\cP(y|x)$ and therefore it satisfies calibration.
\fi

\section{Evaluating language models as risk scores}
\label{sec:llm-as-classifier}

%% put this paragraph somewhere: 
%In practice, calibration of LLMs is typically evaluated over a set of ground truth prompt-response pairs, and predicted score values are binned to approximate~\eqref{eq:calibrated}. We will discuss the different calibration measures implemented in our package in Section~\ref{sec:metric}.

%TODO: In this section, put everything specific to the design of our benchmark, as well as things specific to LLMs, such as prompting etc.

We are interested in the ability of LLMs to express natural uncertainty in outcomes. Therefore, we construct unrealizable binary prediction tasks, and test the model's ability to reflect natural variations in underspecified \textit{individual} outcomes. Specifically, we prompt models with feature values $x$ to elicit risk scores $r$ and then evaluate these scores against ground truth labels $y$ (see Figure~\ref{fig:folktexts_diagram}).
% Recall the illustration of the work flow in Figure~\ref{fig:folktexts_diagram}. 

\subsection{Prediction tasks}
We construct natural language prediction tasks from the American Community Survey (ACS) Public Use Microdata Sample (PUMS).\footnote{\href{https://www.census.gov/programs-surveys/acs/microdata.html}{https://www.census.gov/programs-surveys/acs/microdata.html}} 
The data contains survey responses of about 3.2 million anonymous individuals, carefully curated to offer statistical insights into the population of the United States. We refer to the data as Census data. The Census data contains demographic attributes, as well as information related to income, employment, health, transportation, and housing.
Prediction tasks are defined by selecting a subset of attributes to define the features and one attribute to be the label. % $X$ or target $Y$, among the 28 Census attributes mapped to natural language prompts by \folktexts.
We threshold continuous target variables and bin multi-class predictions to obtain a binary classification task. 
Specifically, we test models on their ability to reflect natural variations in the outcome across the benchmark population.
To enable straightforward comparison with existing tabular benchmarks, we consider natural text analogues to the tasks in the popular {folktables} benchmark package~\citep{ding2021retiring}. %: ACSIncome, ACSEmployment, ACSPublicCoverage, ACSTravelTime, and ACSMobility.
%
% Importantly, our benchmark tasks relate to uncertain outcomes of \textit{individuals} from the US population.
Appendix~\ref{app:prediction_tasks} describes each task in further detail.
%
%Expecting zero-shot LLM answers to be optimal on unseen prediction tasks may be overly ambitious, we would reasonably expect calibrated outputs as a significantly lower bar.~\looseness-1

% \paragraph{LLMs instantiate multiple classifiers.}
% In a standard classification tasks calibration is typically measured with respect to the training data, defining $x$, as well as the distribution $\cP$. For a language model this is a bit different. We specify the population we want to measure explicitly in the prompt. The second moving part is that we can measure calibration over a fixed population given different evidence. Meaning, that we can use the same model but different features to assess calibration. If we provide no evidence than calibration asks for an accurate risk score on $\cP$. 

\paragraph{Natural uncertainty in risk scoring.} Prediction tasks on human populations typically come with natural outcome uncertainty, meaning that the target label is not uniquely determined by the input features (also known as \textit{aleatoric uncertainty}~\citep{der2009aleatory}). The fewer features are provided the higher the uncertainty in the outcome. We take this to our advantage to systematically evaluate calibration. Namely, to be calibrated, a risk score has to reflect both model uncertainty and uncertainty inherent to the prediction task. In fact, the optimal predictor would often output low-confidence answers.
In contrast, prevailing question-answering benchmarks have no data uncertainty, and thus require high confidence for an accurate model to be calibrated.
Underspecified, non-realizable prediction tasks allow us to circumvent such potential confounding between calibration and accuracy.

% In the special case of a realizable prediction task where the features $X$ fully determine the outcome $Y$, calibration and accuracy are related in that a fully confident predictor can only be calibrated if it is simultaneously optimal. 

\subsection{Extracting risk scores}
\label{sec:prompting}

To extract risk scores from LLMs, we map each inference problem to a natural text prompt. For a given data point, we specify the classification task in the prompt and extract class probabilities from the model's next token probabilities, similar to traditional question-answering interfaces. The prompt consists of three components (as shown in Figure~\ref{fig:folktexts_diagram}):
\begin{itemize}[leftmargin=5ex]
\item
\textbf{Instantiating population:} We first instantiate the population $\cP$ in the prompt context. This corresponds to the population represented by our reference data. 
We use third-person prompting: ``\texttt{\small The following data describes a survey respondent.~The survey was conducted among US \\residents in 2018.~Please answer the question based on the information provided.}''
This step is typically not needed in supervised classification tasks, because the training data implicitly defines the population. However, for LLMs this is particularly important for evaluating calibration outside the realizable setting, as risk scores cannot in general simultaneously be calibrated to different populations. Skipping this step could provide insights related to alignment~\cite{Argyle_2023,santurkar23opinion,durmus2023towards} rather than calibration.

%\cite{salewski2024context} show that such impersonation works.
\item
\textbf{Instantiating features:} Next we instantiate an individual. Each individual in the population corresponds to a row in our tabular dataset. We use the US Census codebook to construct a template for transforming attribute/value pairs from the dataset into meaningful natural text representations.
Consider the values $x_i=\{\mathrm{SEX:male}, \mathrm{AGEP:50}\}$ which would correspond to 
``\texttt{\small Information about this person:\textbackslash n - Gender is:~Male.\textbackslash n - Age is:~50 years old.}''
We use a bulleted list of short sentences to encode features.
Related literature~\citep{tamkin2023evaluating,hegselmann23tabllm,fang2024large,yu2023unified,gong20tablegpt} has studied different tabular data encodings in natural text, and found this simple approach to work best.
% Several works~\citep{tamkin2023evaluating,hegselmann23tabllm,liu2024confronting} have found that using a simple data encoding of the form ``[column] is [value]'' yields the best results, which we adopt in our work.

\item
\textbf{Querying outcome:}
% The arguably standard approach for eliciting predicting from language models is to use \textit{multiple-choice prompting}.
We use a standard \emph{multiple-choice} prompting format to elicit outcome predictions from LLMs. The framing of the question for an individual outcome is taken from the original multiple-choice Census questionnaire based on which the data was collected. All answers are presented as binary choices.
Querying about an individual's income would be: 
``\texttt{\small
Question:~What was this person's total income during the past 12 months?\textbackslash n A: Below \$50,000.\textbackslash n B: Above \$50,000.\textbackslash n
Answer:}''
The model's confidence on a given answer is given by the next token probabilities for {\tt A} and {\tt B}.
Additionally, we conduct experiments using a separate chat-style prompt that verbally queries for a numeric probability estimate (dubbed \emph{numeric} prompting).
The above income query would be:
``\texttt{\small
Question:~What is the probability that this person's yearly income is above \$50,000?\textbackslash n Answer (between 0 and 1):~}''
%
% Uncertainty can then be directly quantified using the numeric tokens generated in the following forward passes.
%
This more closely matches how real-world users interact with LLMs, and has been reported to improve uncertainty quantification~\citep{tian2023just}.

% \item
% \textbf{Answer key} and \textbf{uncertainty quantification:}
\end{itemize}

When using the constructed multiple-choice prompts, we query the models and extract scores from the next token probabilities for the choice labels \texttt{A,B} as $r_i = \Pr(\texttt{A})/(\Pr(\texttt{A})+\Pr(\texttt{B}))$, following the methodology of standard question-answering benchmarks~\cite{hendrycks2021measuring,santurkar23opinion}.
As LLMs are known to have ordering biases in multiple-choice question-answering~\cite{dominguez2023questioning,robinson2023leveraging}, we evaluate responses on all choice orderings and average the resulting scores.
When using numeric prompting, we prefix the answer with `{\tt 0.}' to improve the likelihood of a direct numeric response, and run two forward passes, selecting the highest likelihood numeric token at each iteration.
We refer to \citet{xiong2024can} for an overview on alternative design choices on how to elicit confidence scores from LLMs.

\subsection{The {\tt folktexts} package}
\label{sec:package}

The \folktexts package is designed to offer a flexible interface between tabular prediction tasks and natural language question-answering tasks in order to extract risk scores from LLMs. The package makes available the ACSIncome, ACSPublicCoverage, ACSMobility, ACSEmployment, and ACSTravelTime prediction tasks~\cite{ding2021retiring} as natural language benchmarks, together with various functionalities to customize the task definitions (e.g., use a different set of features to predict income) and subsample the reference data (e.g., predict income only among college graduates in California).
The set of attributes available to define the features and label can be found in the ACS PUMS data dictionary.\footnote{ \url{https://www.census.gov/programs-surveys/acs/microdata/documentation.html}}.
Additionally, \folktexts is compatible with open-source models running locally, as well as with closed-source models hosted through a web API. However, APIs must make available the next token probabilities for each forward pass instead of returning discrete text completions  --- the OpenAI API is one example that is compatible with \folktexts out of the box.

In addition to providing a reproducible way to extract risk scores from LLMs, \folktexts also offers pre-implemented evaluation metrics to benchmark and compare the calibration and accuracy of LLMs, as well as easy plotting of group-conditional calibration curves for a cursory view of potential biases. The package is easy to use within a python notebook, as a dependency, or directly from the command line. Further details on usage and design choices are available in Appendix~\ref{app:folktexts}.

\section{Empirical findings}
\label{sec:results}
We use \folktexts to evaluate several recently released models together with their instruction-tuned counterparts: the Llama 3 models~\citep{llama3modelcard}, including the 8B and the 70B versions, the Mistral 7B~\citep{jiang2023mistral}, the Mixtral 8x7B and 8x22B variants~\citep{jiang2024mixtral}, the Yi 34B~\citep{young2024yi}, and the Gemma~\citep{gemmateam2024gemma} 2B and 7B variants.
We also evaluate GPT 4o mini~\citep{achiam2023gpt} through the OpenAI API (note that no base model version is available).
Instruction-tuned models are marked `(i.t.)'.
%
% All LLM results are obtained via zero-shot prompting with no fine-tuning.
%
For comparison, results are also shown for a logistic regression (LR) model, and a gradient boosted decision trees model (XGBoost).
The XGBoost model~\citep{chen2016xgboost} is generally regarded as the state-of-the-art in tabular data tasks~\citep{borisov2024deep}. %\citep{grinsztajn2022tree}

In this section we focus on the ACSIncome prediction task, which is the default \folktexts benchmarking task. It consists in predicting whether a person's income is above or below \$50K from 10 demographic features, and closely emulates the popular UCI Adult prediction task~\cite{misc_adult_2}. 
The evaluation test set consists of 160K randomly selected samples from the 2018 Census data. A separate set of 1.5M samples is used to train the supervised LR and XGBoost models, LLMs are used as zero-shot classifiers without fine-tuning.
Both multiple-choice prompting and chat-style numeric prompting were used to obtain two separate risk score distributions for each model (as described in Section~\ref{sec:prompting}).
We focus on analyzing multiple-choice prompting results, as it is arguably the standard in LLM benchmarking~\citep{hegselmann23tabllm,sanh2021multitask,tabletSlack23,Argyle_2023,sanders2023demonstrations,horton2023large,hendrycks2021measuring,santurkar23opinion,durmus2023towards,dominguez2023questioning,scherrer23belief}. %, and allows for tighter adherence to Census questionnaire wording.
Appendix~\ref{app:additional_results} presents additional results and plots for the experiments analyzed in this section, including a more in-depth look into numeric prompting results, as well as results on alternative prediction tasks.
Experiments were ran on a cluster with Nvidia-A100-80GB GPUs, consuming an approximate total of 500 GPU hours.

\begin{table}[t!]
\centering
\footnotesize{\begin{tabular}{l|cccc|cccc}
\toprule
\multirow{3}{*}{\textbf{Model}} & \multicolumn{4}{c|}{\textbf{Multiple-choice prompting}} & \multicolumn{4}{c}{\textbf{Numeric risk prompting}} \\
 & \multirow{2}{*}{\textbf{ECE $\downarrow$}} & \multirow{2}{*}{\textbf{\shortstack{Brier \\score}} $\downarrow$} & \multirow{2}{*}{\textbf{\shortstack{AUC} $\uparrow$}} & \multirow{2}{*}{\textbf{\shortstack{Acc.} $\uparrow$}} & \multirow{2}{*}{\textbf{ECE $\downarrow$}} & \multirow{2}{*}{\textbf{\shortstack{Brier \\score}} $\downarrow$} & \multirow{2}{*}{\textbf{\shortstack{AUC} $\uparrow$}} & \multirow{2}{*}{\textbf{\shortstack{Acc.} $\uparrow$}} \\
 & & & & & & & & \\
\midrule
GPT 4o mini (it) & 0.24 & 0.24 & \cellcolor{cyan!17.6} 0.85 & 0.74 & \cellcolor{cyan!25.0} 0.05 & \cellcolor{cyan!25.0} 0.16 & \cellcolor{cyan!20.6} 0.83 & \cellcolor{cyan!24.4} 0.78 \\
Mixtral 8x22B (it) & 0.21 & \cellcolor{cyan!3.6} 0.22 & \cellcolor{cyan!11.2} 0.85 & \cellcolor{cyan!11.1} 0.76 & 0.11 & \cellcolor{cyan!15.5} 0.17 & \cellcolor{cyan!25.0} 0.84 & \cellcolor{cyan!18.9} 0.77 \\
Mixtral 8x22B & \cellcolor{cyan!13.2} 0.17 & \cellcolor{cyan!21.6} 0.19 & \cellcolor{cyan!16.5} 0.85 & 0.68 & 0.13 & \cellcolor{cyan!3.6} 0.18 & \cellcolor{cyan!9.6} 0.82 & \cellcolor{cyan!2.4} 0.74 \\
Llama 3 70B (it) & 0.27 & 0.27 & \cellcolor{cyan!25.0} 0.86 & 0.69 & 0.25 & 0.23 & \cellcolor{cyan!22.1} 0.84 & 0.67 \\
Llama 3 70B & 0.20 & \cellcolor{cyan!14.9} 0.20 & \cellcolor{cyan!20.8} 0.86 & 0.70 & 0.27 & 0.24 & \cellcolor{cyan!6.7} 0.82 & 0.54 \\
Mixtral 8x7B (it) & \cellcolor{cyan!16.8} 0.16 & \cellcolor{cyan!25.0} 0.18 & \cellcolor{cyan!19.7} 0.86 & \cellcolor{cyan!25.0} 0.78 & 0.10 & \cellcolor{cyan!15.5} 0.17 & \cellcolor{cyan!22.8} 0.84 & \cellcolor{cyan!12.8} 0.76 \\
Mixtral 8x7B & \cellcolor{cyan!10.6} 0.17 & \cellcolor{cyan!11.5} 0.21 & 0.83 & 0.65 & \cellcolor{cyan!4.0} 0.07 & \cellcolor{cyan!13.1} 0.17 & \cellcolor{cyan!3.7} 0.81 & \cellcolor{cyan!25.0} 0.78 \\
Yi 34B (it) & \cellcolor{cyan!1.3} 0.19 & \cellcolor{cyan!18.8} 0.19 & \cellcolor{cyan!19.7} 0.86 & 0.72 & 0.22 & 0.21 & 0.80 & 0.48 \\
Yi 34B & 0.25 & \cellcolor{cyan!2.5} 0.22 & \cellcolor{cyan!13.3} 0.85 & 0.62 & 0.15 & 0.19 & \cellcolor{cyan!17.7} 0.83 & 0.61 \\
Llama 3 8B (it) & 0.32 & 0.30 & \cellcolor{cyan!13.3} 0.85 & 0.62 & 0.23 & 0.23 & \cellcolor{cyan!0.1} 0.81 & 0.67 \\
Llama 3 8B & 0.25 & 0.26 & 0.81 & \cellcolor{orange!20.8} 0.38 & 0.14 & 0.24 & 0.63 & \cellcolor{orange!3.7} 0.40 \\
Mistral 7B (it) & 0.21 & \cellcolor{cyan!4.7} 0.22 & 0.83 & \cellcolor{cyan!16.5} 0.77 & 0.16 & 0.19 & \cellcolor{cyan!14.7} 0.83 & 0.70 \\
Mistral 7B & 0.20 & 0.23 & 0.80 & 0.73 & \cellcolor{orange!15.7} 0.36 & 0.32 & 0.75 & 0.49 \\
Gemma 7B (it) & \cellcolor{orange!14.7} 0.61 & \cellcolor{orange!4.7} 0.59 & \cellcolor{cyan!3.8} 0.84 & \cellcolor{orange!25.0} 0.37 & 0.33 & 0.30 & 0.78 & 0.42 \\
Gemma 7B & 0.24 & 0.27 & 0.76 & \cellcolor{orange!25.0} 0.37 & 0.15 & 0.20 & 0.80 & 0.73 \\
Gemma 2B (it) & \cellcolor{orange!25.0} 0.63 & \cellcolor{orange!25.0} 0.63 & 0.73 & \cellcolor{orange!25.0} 0.37 & 0.28 & 0.31 & \cellcolor{orange!25.0} 0.50 & \cellcolor{orange!25.0} 0.37 \\
Gemma 2B & \cellcolor{cyan!25.0} 0.14 & 0.25 & \cellcolor{orange!25.0} 0.62 & 0.45 & \cellcolor{orange!25.0} 0.37 & \cellcolor{orange!25.0} 0.37 & \cellcolor{orange!25.0} 0.50 & 0.63 \\
\midrule
LR & 0.03 & 0.18 & 0.79 & 0.74 & 0.03 & 0.18 & 0.79 & 0.74 \\
% GBM & 0.01 & 0.13 & 0.89 & 0.81 & 0.01 & 0.13 & 0.89 & 0.81 \\
XGBoost & 0.00 & 0.13 & 0.90 & 0.82 & 0.00 & 0.13 & 0.90 & 0.82 \\
\bottomrule
\end{tabular}}
% \footnotesize{\input{tables/multiple_choice_prompting/acsincome.tex}}
\caption{Zero-shot LLM results on the \textbf{ACSIncome} benchmark task, together with results for LR and XGBoost baselines fitted on 1.5M samples.
% Column ``{Acc. $\tau=t^\prime$}'' shows accuracy results after fitting a model binarization threshold on $n=100$ train samples, while ``Acc. $\tau=0.5$'' shows standard accuracy using the highest likelihood answer.
Table cells are colored using a continuous color map between the worst ({\color{orange} in orange}) and best ({\color{cyan} in cyan}) results of each column. %, where only the top 10\% and bottom 10\% of the scale is colored.
Models generally achieve high predictive signal (high AUC) but poor calibration (high ECE).
Numeric prompting leads to improved ECE but worse AUC.
}
\label{tab:acsincome_results}
\end{table}

\subsection{Benchmark results}

We perform a comprehensive evaluation of risk scores output by LLMs along multiple metrics. 
Results are summarized in Table~\ref{tab:acsincome_results}.

\paragraph{Multiple-choice prompting.}
We observe that a majority of LLMs (all of size 8B or larger) outperform the linear model baseline (LR) in terms of predictive power (AUC).
However, LLMs are clearly far from matching the supervised baselines with respect to calibration: all language models achieve very high calibration error (ECE), while baselines achieve near-perfect calibration.
Due to this high miscalibration, models struggle to translate scores with high predictive signal into high accuracy.
In fact, while most models achieve high AUC, most models struggle to surpass the supervised linear baseline in terms of accuracy.
We recall that AUC is agnostic to calibration, while accuracy on the maximum likelihood answer is not.
%
% This fact can lead to inaccurate answers out of a model, while at the same time having high predictive ability within its output token probabilities.
%
All of the Gemma models have worse than random accuracy, despite having clearly above random AUC (random would be 0.5).
Only the instruction-tuned Mistral models (7B, 8x7B, and 8x22B) outperform the linear model in terms of accuracy.
Interestingly, while larger models achieve higher AUC, calibration is not reliably improved by model size --- differences across model families are more pronounced than across model sizes.

Finally, we focus on comparing base models to their instruction-tuned counterparts, marked with `(it)'.
A striking trend is visible across the board: instruction-tuning generally worsens calibration (higher ECE) when using multiple-choice prompting.
At the same time, we generally see improvements in AUC and accuracy after instruction-tuning.
%
% The highest accuracy LLMs are all instruction-tuned Mistral models: Mixtral 8x7B (it), Mistral 7B (it), and Mixtral 8x22B (it), in descending ranking.
%
Appendix~\ref{app:results_on_additional_tasks} presents results on the four additional prediction tasks.
The same trend of instruction-tuning leading to worse calibration and higher AUC is broadly replicated.
However, performance across different tasks is somewhat inconsistent: LLMs generally have stronger predictive signal than a supervised linear model on the income prediction and travel-time prediction tasks, but consistently underperform the linear baseline on the address change task (ACSMobility).

\begin{figure}[t!]
    \centering
    \includegraphics[width=0.85\textwidth]{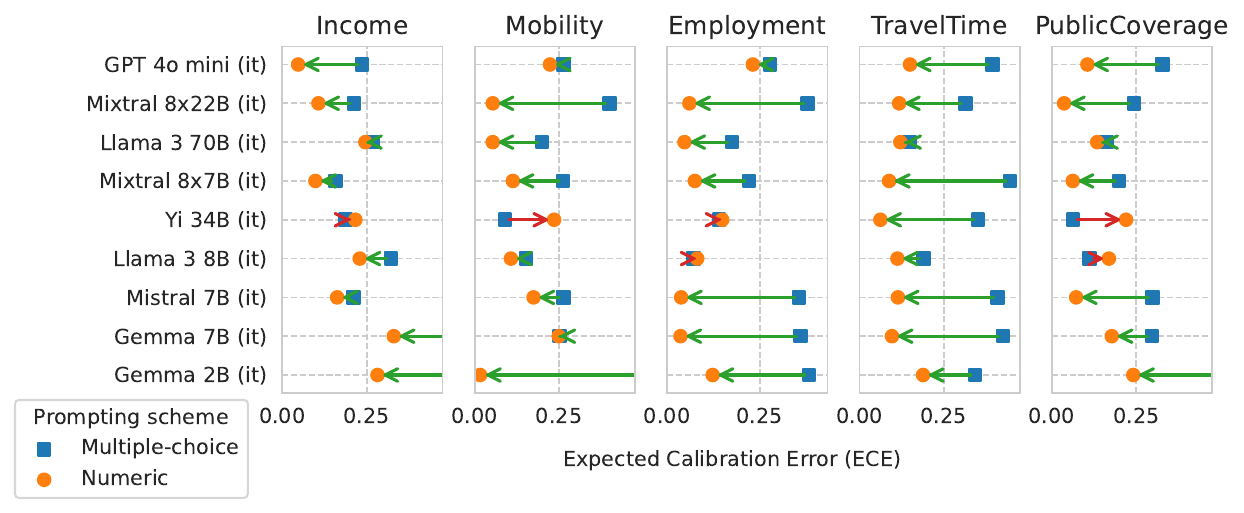}
    \caption{Change in calibration error (ECE) of instruction-tuned models when using numeric risk prompting (orange circles) versus multiple-choice prompting (blue squares). Improvement/deterioration is represented by green/red arrows, respectively. An overwhelming majority of model/task pairs see calibration improvements.
    }
    \label{fig:diff_prompts_ece_change_only_instruct}
\end{figure}

\paragraph{Numeric prompting.}
We observe broad improvements in calibration (lower ECE) and Brier score loss across instruction-tuned models when using numeric prompting.
Figure~\ref{fig:diff_prompts_ece_change_only_instruct} shows the change in ECE between both prompting schemes.
The trend is clear across most instruct models across all five benchmark tasks. The Yi 34B model is the exception, showing outlier results throughout the different experiments we conduct.
On the other hand, results for base models are less conclusive (see Figure~\ref{fig:diff_prompts_ece_change}).
At the same time, numeric prompting leads to small but consistent drops in predictive power of risk scores (AUC) on 4 out of 5 benchmark tasks, including ACSIncome.
Figure~\ref{fig:diff_prompts_auc_change} shows these changes visually for all tasks.
Finally, we point out that the {GPT 4o mini} model produces a surprisingly well-calibrated risk score distribution for the income prediction task ($\text{ECE}=0.05$).
In fact, for each benchmark task, at least one LLM is able to produce a remarkably well-calibrated score distribution (although a different model for different tasks).
This promising result points to the fact that some small amount of data will likely always be needed to properly evaluate LLMs capabilities to model human population statistics, but such modelling can often be done with high degree of confidence and a good understanding on which outputs might be wrong.
%
% Well calibrated distributions, when available, open the door to opportunities in modelling human population statistics with very little data.

\begin{figure}[t!]
    \includegraphics[height=11em]{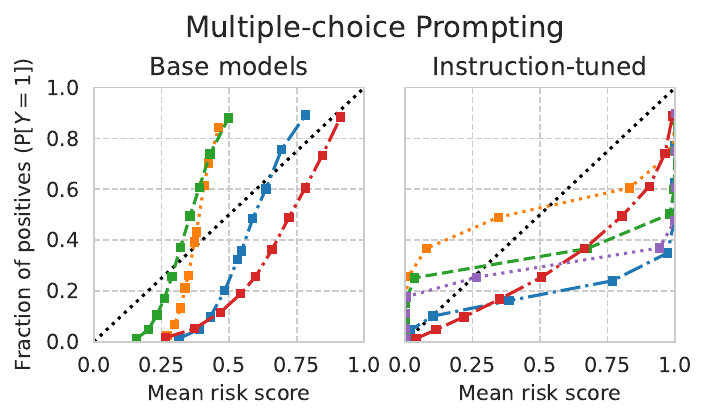}
    \includegraphics[height=11em]{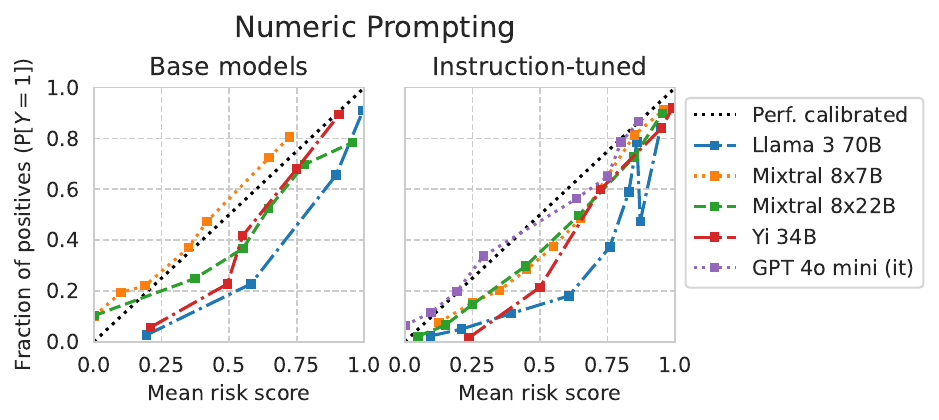}
    \caption{Calibration curves for base and instruction-tuned versions of the largest models studied, on the ACSIncome task. Curves are computed using $10$ quantile-based score bins.
    Risk scores were generated using multiple-choice-style prompting (\textit{left plots}) or numeric chat-style prompting (\textit{right plots}).
    }
    \label{fig:calibration_base_instr}
\end{figure}

\subsection{Score distribution}

To get additional insights into the difference between base models and their instruction-tuned counterparts, we inspect their risk score distributions more closely. 

\begin{figure}[t!]
    \centering
    \begin{subfigure}[b]{0.45\textwidth}
        \centering
        \includegraphics[width=\textwidth]{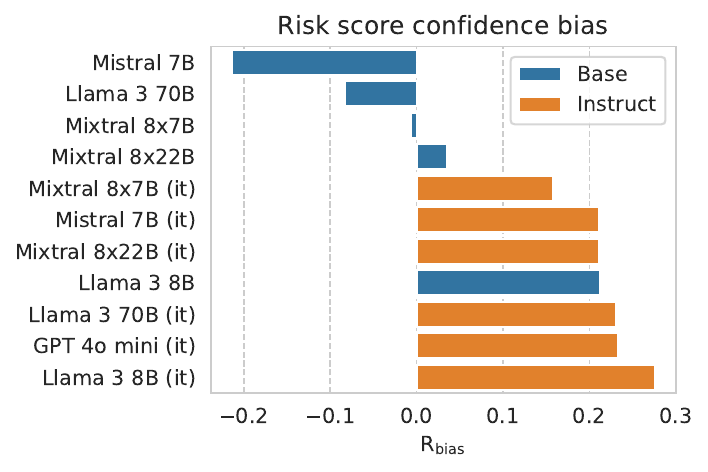}
        \caption{Multiple-choice prompting.}
        \label{fig:under_over_multiple_choice}
    \end{subfigure}
    \hfill
    \begin{subfigure}[b]{0.45\textwidth}
        \centering
        \includegraphics[width=\textwidth]{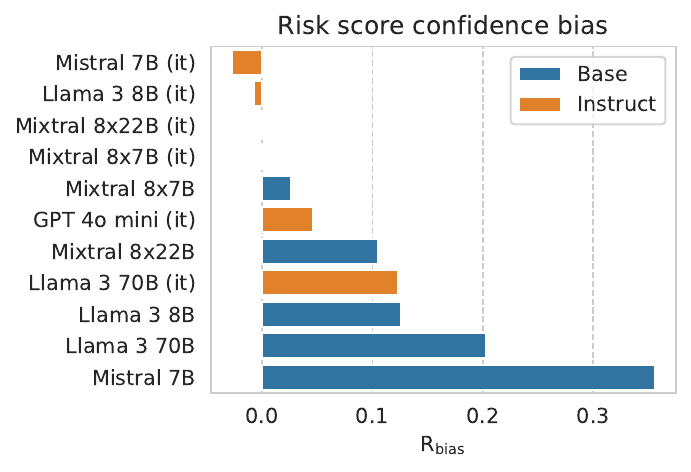}
        \caption{Numeric prompting.}
        \label{fig:under_over_numeric}
    \end{subfigure}
    \caption{Risk score confidence bias for all LLMs on the ACSIncome task. Negative values indicate under-confident risk scores (overestimating uncertainty), while positive values indicate over-confident risk scores (underestimating uncertainty).
    Instruction-tuned models are generally over-confident when using multiple-choice prompting (Fig.~\ref{fig:under_over_multiple_choice}), but this bias is considerably diminished when using numeric prompting (Fig.~\ref{fig:under_over_numeric}).
    }
    \label{fig:under_over_risk_scores}
\end{figure}

Figure~\ref{fig:calibration_base_instr} shows calibration curves of the largest LLMs studied, for both base and instruct variants and for both prompting schemes.
When using {multiple-choice prompting} (left-most plots of Fig.~\ref{fig:calibration_base_instr}), 
both base and instruction-tuned models have poor score calibration, but failure modes are entirely different: Base models output under-confident scores, while instruction-tuned models output over-confident scores.
To quantify this result, we introduce a measure of risk score \textit{confidence bias} similar to the ECE metric: 
\[\mathrm{R}_\mathrm{bias} \coloneq \sum_{m=1}^M \frac{|B_m|}{n} \left[ \mathrm{Conf}(B_m) - \mathrm{Acc}(B_m) \right],\] 
where $M$ is the number of score bins, $B_m$ is the set of samples in score bin $m$, $\mathrm{Acc}(.)$ is the accuracy on a given set of samples, and $\mathrm{Conf}(.)$ is the confidence on a given set of samples measured as the mean risk score for the highest likelihood class.
Figure~\ref{fig:under_over_multiple_choice} shows the risk score confidence bias results.
The two miscalibration modes are evident: confidence bias is higher for instruction-tuned models and lower (or even negative) for base models. That is, instruction-tuned models are generally over-confident in their predictions, outputting higher scores than their accuracy would warrant, while no such trend is visible for base models.
%
% Section~\ref{sec:subgroup_calibration} further analyzes subgroup calibration and its implications to algorithmic fairness.
%
On the other hand, when using {numeric prompting}, the two aforementioned failure modes are no longer evident, and the differences between base and instruction-tuned models are blurred (right-most plots of Fig.~\ref{fig:calibration_base_instr}).
In fact, Figure~\ref{fig:under_over_numeric} shows a trend reversal when using numeric prompting: Base models now show a higher over-confidence in their risk scores, while instruction-tuned models show approximately neutral score bias.

\begin{figure}[tp]
    \centering
    \includegraphics[width=0.97\textwidth]{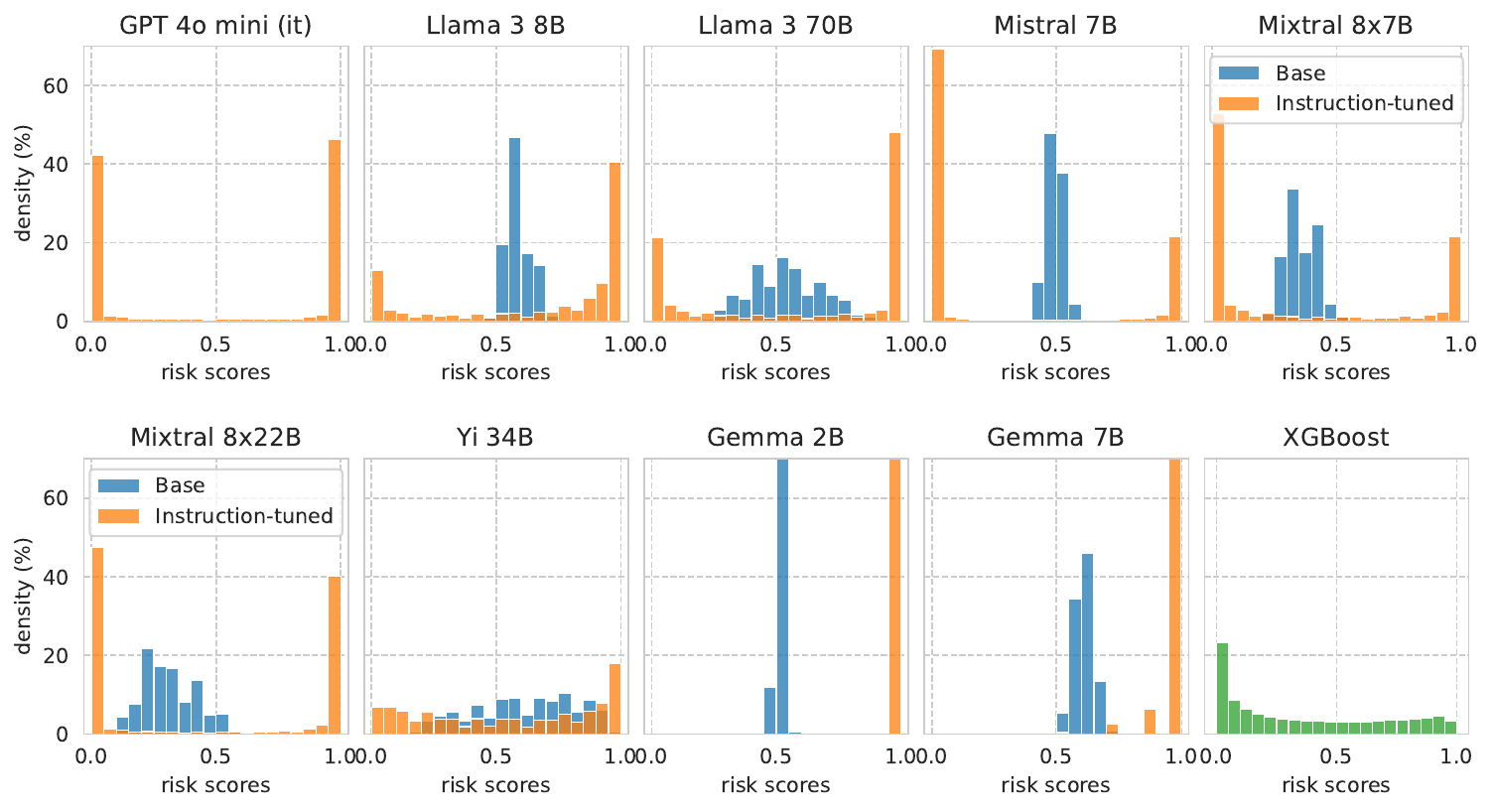}
    \caption{Risk score distribution for base and instruction-tuned model pairs on the ACSIncome task, using \emph{multiple-choice} prompting. After instruction-tuning, models exhibit high confidence, but worse calibration in general.
    The XGBoost scores showcase a perfectly calibrated distribution ($\text{ECE} \approx 0.00$).
    % See Table~\ref{tab:acsincome_results} for details.\looseness-1
    }
    \label{fig:score-dist}
\end{figure}

Figure~\ref{fig:score-dist} shows the risk score distribution for a variety of model pairs, produced using multiple-choice prompting.
The score distributions for base and instruct model variants are immediately distinguishable: base models consistently produce low-variance distributions centered around 0.5, while instruction-tuned variants often output scores near 0 or 1.
The same trend is visible on all model pairs, with {Yi 34B} showing the smallest difference between base and instruct variants. % (the Yi 34B model pair achieves somewhat outlier results throughout the paper, with smaller differences between model variants).
The score distributions produced by base/i.t. model pairs are markedly different, even among base/i.t. pairs achieving the exact same AUC; e.g., Llama 3 70B and Mixtral 8x22B.
% Even for the base/instruct pairs that achieve exactly the same AUC (Llama 3 70B, Mixtral 8x22B).
For the largest Llama (70B) and largest Mistral (8x22B) models, no predictive performance is gained by instruction tuning, but answers of the instruction-tuned models have significantly higher (over-)confidence and worse calibration.
In fact, while the base Llama 3 70B has an average of 0.07 under-confidence bias, the instruct variant produces risk scores on average 0.22 over-confident (see Figure~\ref{fig:under_over_multiple_choice}).
Note that the instruction-tuned Gemma models degenerate into predicting only positive outcomes with high confidence score, hence why they have the lowest accuracy, worst Brier score, and worst ECE.

Crucially, our evaluation reveals a previously unreported shortcoming of using multiple-choice prompting with instruction-tuned models: instruction-tuning polarizes score distributions, even if the true outcome has high entropy. % !!!
Standard realizable knowledge testing benchmarks can easily disguise this polarization phenomenon as improper quantification of \emph{model} uncertainty.
In fact, it seems evident that it is improper quantification of uncertainty in general, regardless of underlying uncertainty in the modelled distribution.
The following subsection goes further in-depth on the influence of data uncertainty in score distribution.
Appendix~\ref{app:numeric_prompting} analyzes numeric prompting results.

% \begin{figure}[t!]
%     \centering
%     \includegraphics[width=0.49\textwidth]
%     {imgs/multiple_choice_prompting/under_over_score.ACSIncome.multiple-choice-prompt.pdf}
%     \includegraphics[width=0.49\textwidth]{imgs/numeric_prompting/under_over_score.ACSIncome.numeric-prompt.pdf}
%     \caption{Risk score confidence bias for all LLMs on the ACSIncome task. Negative values indicate under-confident risk scores (overestimating uncertainty), while positive values indicate over-confident risk scores (underestimating uncertainty). For all but the Yi model pair, the instruct model shows higher over-confidence.%~\looseness-1
%     }
%     \label{fig:under_over_risk_scores}
% \end{figure}

% Empirical insights:
% - large LLMs have higher zero-shot predictive power (without any training) than linear models trained on the full dataset;
% - base models and instruction-tuned versions have the same predictive power, but the later are way more uncalibrated;
% - few-shot prompting (with a fixed set of representative examples, n=10) calibrates the model outputs by providing insight into the base rate;
    % - TODO: compare score distribution of instruction-tuned model, base-model, base-model + few-shot, and true score distributions; (should be in this order of closeness to true scores)
    % - would the outputs be equally calibrated by simply stating the base-rate? "The percentage of people that earn above 50K is 40%"

\subsection{Varying degree of uncertainty}

Next, we consider the dependence of multiple-choice risk scores on the available evidence. For this study we use the Mixtral 8x7B model, which achieves the best (lowest) Brier score among evaluated models (reflecting both high accuracy and high calibration).
We compute income prediction risk scores with increasing evidence: starting with only 2 features, and iteratively adding 2 features at a time (see results in Figure~\ref{fig:increasing_evidence}).
This sequence demonstrates how unrealizable, underspecified prediction tasks differ from realizable prediction tasks.
Predicting income based on an individual's place of birth (POBP) and race (RAC1P) is naturally not possible to a high degree of accuracy, forcing any calibrated model to output lower-confidence risk scores.
% Even with all 10 features, state-of-the-art supervised models only achieve 0.82 accuracy.
%
Indeed, both base and instruct variants correctly output lower confidence scores for the smaller feature sets when compared with the larger feature sets (compare left-most to right-most plots of Fig.~\ref{fig:increasing_evidence}).
However, instruction-tuning still leads to a clear polarization of risk score distribution, regardless of true data uncertainty: 
Score variance for base models is in range $\sigma \in [0.02, 0.06]$, while for i.t. models it's in range $\sigma \in [0.16, 0.41]$.
% While score variance for the base model goes from $\sigma=0.02$ to $\sigma=0.06$ with increasing number of features, the i.t. model score variance starts $\sigma=0.16$ and goes up to $\sigma=0.41$ under the same input distribution.
%
Appendix~\ref{app:varying_uncertainty} goes further in-depth on how score distributions change with varying data uncertainty. 
% Moreover, the score distribution for the base Mixtral 8x7B model is correctly centered around the true label prevalence of $\Pr\left\{ Y=1 \right\}=0.37$ on this task.
% \todo{comment on evaluation of exact risk scores -- higher feature overlap enables us to get a more accurate estimate of the ground-truth distribution of $Y|X=x$}

In addition to providing insights into risk scores, varying individual features in the prediction task also provides insights into LLM feature importance. Appendix~\ref{app:feature_importance} presents LLM feature importance results and discusses the main differences to traditional supervised learning models.

\begin{figure}[tp]
    \centering
    \includegraphics[height=11.3em]{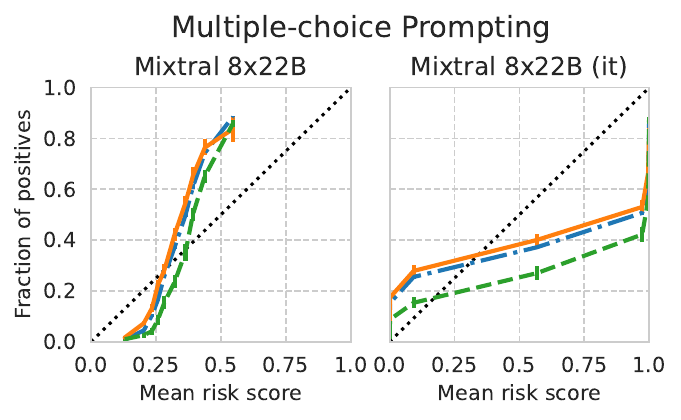}
    \includegraphics[height=11.3em]{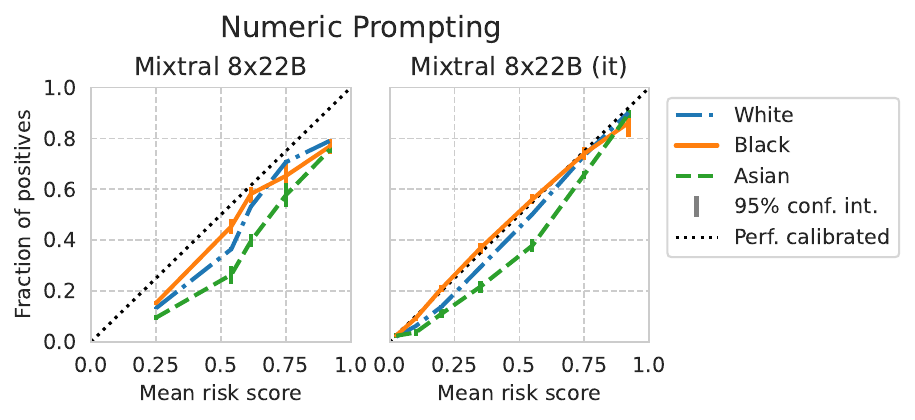}
    \includegraphics[height=11.3em]{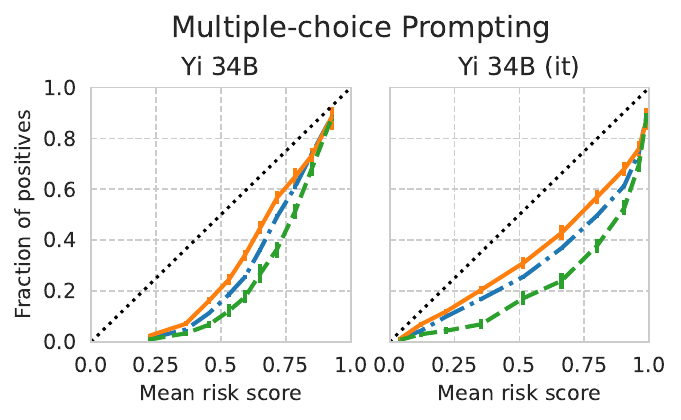}
    \includegraphics[height=11.3em]{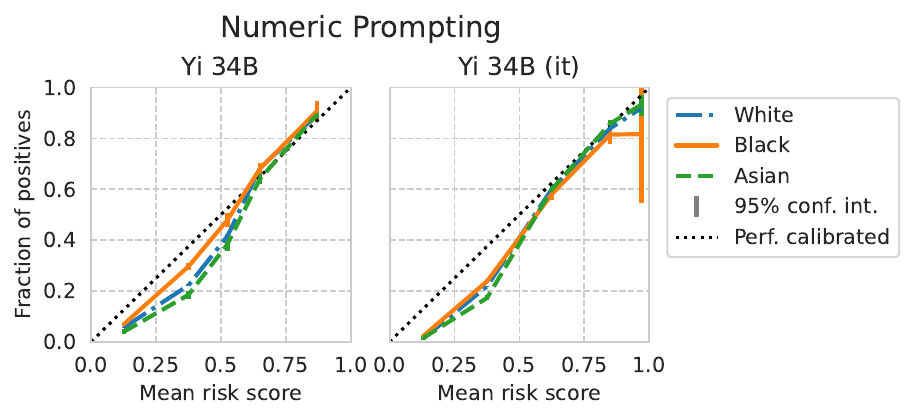}
    \caption{Calibration curves for the Mixtral 8x22B (\textit{top}) and the Yi 34B (\textit{bottom}) models, across different race subpopulations on the ACSIncome task.
    \textit{Left}: Multiple-choice-style prompting. \textit{Right}: Numeric or chat-style prompting.
    The models shown are the worst offenders in terms of subgroup-specific score bias.
    On average, Black individuals see a lower score ($x$ axis) for the same true probability of positive outcomes ($y$ axis).\looseness-1
    }
    \label{fig:subgroup-calibration-yi-mixtral}
\end{figure}

\subsection{Subgroup calibration}
\label{sec:subgroup_calibration}

Finally, we evaluate risk score calibration across different subpopulations, such as typically done as part of a fairness audit.
Figure~\ref{fig:subgroup-calibration-yi-mixtral} shows calibration curves for two sets of models on the ACSIncome task, evaluated on three subgroups specified by the instantiation of the race attribute in the data. We pick the three race categories with the largest representation in the ACS data.
Note that a positive prediction of $Y=1$ is arguably the advantageous outcome, as it corresponds to the high-income category (``Earns above \$50,000 per year'').
%
% For base models, calibration curves are generally similar across subgroups.
% However, for instruction-tuned models, 
%
On the two models shown, samples belonging to the `Black' population see consistently lower scores for the same positive label probability when compared to the `Asian' or `White' populations.
% In other words, for the same score $R=r$, the probability of a positive label $Y=1$ is higher for $S=\text{`Black'}$ individuals: $\Pr[Y=1|R=r, S=\text{`Black'}] \geq \Pr[Y=1|R=r, S \neq \text{`Black'}]$, where $S$ denotes the census encoding of \emph{race}.
%
The Mixtral 8x22B (it) and Yi 34B (it) models shown are the worst offenders, with an average risk score difference between Asian and Black groups of 0.17 and 0.13 respectively; i.e., a Black individual will receive on average a 0.17 or 0.13 lower score than an Asian individual for the same true probability of high-income $\Pr\{Y=1\}$.
This poses a higher bar for Black individuals to get a positive prediction of ``high-income''.
Note that the remaining models studied do not show such striking differences in group-conditional calibration.
In fact, this score bias can be reversed for some \textit{base} models, overestimating scores from Black individuals compared with other subgroups (results for all models shown in Appendix~\ref{app:additional_subgroup_cali_plots}). % presents results on all models together with further discussion on subgroup miscalibration.
Such differences in score calibrations arguably warrant a more in-depth analysis that escapes the scope of this paper.
We raise concerns regarding subgroup miscalibration, which should caution practitioners against using such scores in consequential domains without a comprehensive fairness audit.
%

% \section{Conclusion}
\section{Discussion}

% \anote{We should really focus on the fact that uncertainty is often disregarded in benchmarks, and that no other benchmark evaluates \textit{data} uncertainty, instead evaluating \textit{model} uncertainty...}

We introduced \folktexts, a software package that provides datasets and tools to evaluate risk scores produced by language models.
% We believe that \folktexts strengthens the evaluation ecosystem in a direction that was previously underserved.
%
%, specifically, the systematic evaluation of uncertainty quantification in LLMs. 
%
Unlike most existing LLM benchmarks, the datasets we introduced have inherent outcome uncertainty, making them useful as a basis for systematic evaluation of uncertainty quantification in LLMs.
While uncertainty on realizable tasks reflects only model uncertainty (i.e., whether the model is aware of its lack of knowledge), 
uncertainty on unrealizable tasks is itself a type of knowledge over the underlying data distribution. % --- one that LLMs seem unable to learn.
%
% Accurate risk score estimates can thus be used for data generation.

% Crucially, our evaluation reveals a previously unreported shortcoming of using multiple-choice prompting with instruction-tuned models: instruction-tuning polarizes score distributions, even if the true outcome has high entropy. % !!!
% %
% Standard realizable knowledge testing benchmarks can easily disguise this polarization phenomenon as improper quantification of \emph{model} uncertainty.
% In fact, it seems evident that it is improper quantification of uncertainty in general, regardless of underlying uncertainty in the modelled distribution.

Our empirical findings show that LLM risk scores produced using standard multiple-choice Q\&A generally have strong predictive signal, but are wildly miscalibrated. Such models may be good for knowledge testing, but lack adequate indicators of uncertainty, making them unsuitable for synthetic data generation.
Instruction-tuned models generally have worse calibration but slightly higher accuracy and AUC than their base-model counterparts on multiple-choice answers.
%
% Crucially, we demonstrate how inspecting calibration curves and score distributions is necessary to understand differences across models, such as the differences between base and instruct model variants. %, which were obfuscated when looking only at accuracy or single calibration metrics.
In fact, instruction-tuning leads to marked polarization of multiple-choice answer distribution, regardless of underlying true data uncertainty.
This reveals a general inability of instruction-tuned LLMs to quantify uncertainty using multiple-choice Q\&A.
At the same time, verbally querying models for numeric probability estimates considerably improves calibration of instruction-tuned models, but at a small cost in AUC.
%
% %
Going forward, we envision that future package extensions will include other uncertainty quantification methods, such as confidence intervals, or conformal prediction methods.

%% Paper acknowledgements
\section*{Acknowledgments}
We thank Florian Dorner, Mila Gorecki, and Ricardo Dominguez-Olmedo for invaluable feedback on an earlier version of this paper.
The authors thank the International Max Planck Research School for Intelligent Systems (IMPRS-IS) for supporting Andr\'e F. Cruz.

%% References
% \newpage
\bibliography{refs}

\begin{thebibliography}{75}
\providecommand{\natexlab}[1]{#1}
\providecommand{\url}[1]{\texttt{#1}}
\expandafter\ifx\csname urlstyle\endcsname\relax
  \providecommand{\doi}[1]{doi: #1}\else
  \providecommand{\doi}{doi: \begingroup \urlstyle{rm}\Url}\fi

\bibitem[Tamkin et~al.(2023)Tamkin, Askell, Lovitt, Durmus, Joseph, Kravec, Nguyen, Kaplan, and Ganguli]{tamkin2023evaluating}
Alex Tamkin, Amanda Askell, Liane Lovitt, Esin Durmus, Nicholas Joseph, Shauna Kravec, Karina Nguyen, Jared Kaplan, and Deep Ganguli.
\newblock Evaluating and mitigating discrimination in language model decisions.
\newblock \emph{arXiv preprint arXiv:2312.03689}, 2023.

\bibitem[Kasneci et~al.(2023)Kasneci, Se{\ss}ler, K{\"u}chemann, Bannert, Dementieva, Fischer, Gasser, Groh, G{\"u}nnemann, H{\"u}llermeier, et~al.]{kasneci2023chatgpt}
Enkelejda Kasneci, Kathrin Se{\ss}ler, Stefan K{\"u}chemann, Maria Bannert, Daryna Dementieva, Frank Fischer, Urs Gasser, Georg Groh, Stephan G{\"u}nnemann, Eyke H{\"u}llermeier, et~al.
\newblock Chatgpt for good? on opportunities and challenges of large language models for education.
\newblock \emph{Learning and individual differences}, 103:\penalty0 102274, 2023.

\bibitem[Thirunavukarasu et~al.(2023)Thirunavukarasu, Ting, Elangovan, Gutierrez, Tan, and Ting]{thirunavukarasu2023large}
Arun~James Thirunavukarasu, Darren Shu~Jeng Ting, Kabilan Elangovan, Laura Gutierrez, Ting~Fang Tan, and Daniel Shu~Wei Ting.
\newblock Large language models in medicine.
\newblock \emph{Nature medicine}, 29\penalty0 (8):\penalty0 1930--1940, 2023.

\bibitem[Gaebler et~al.(2024)Gaebler, Goel, Huq, and Tambe]{gaebler2024auditing}
Johann~D Gaebler, Sharad Goel, Aziz Huq, and Prasanna Tambe.
\newblock Auditing the use of language models to guide hiring decisions.
\newblock \emph{arXiv preprint arXiv:2404.03086}, 2024.

\bibitem[Chouldechova(2017)]{chouldechova2017fair}
Alexandra Chouldechova.
\newblock Fair prediction with disparate impact: A study of bias in recidivism prediction instruments.
\newblock \emph{Big data}, 5\penalty0 (2):\penalty0 153--163, 2017.

\bibitem[Pleiss et~al.(2017)Pleiss, Raghavan, Wu, Kleinberg, and Weinberger]{pleiss2017fairness}
Geoff Pleiss, Manish Raghavan, Felix Wu, Jon Kleinberg, and Kilian~Q Weinberger.
\newblock On fairness and calibration.
\newblock \emph{Advances in neural information processing systems}, 30, 2017.

\bibitem[Corbett-Davies et~al.(2023)Corbett-Davies, Gaebler, Nilforoshan, Shroff, and Goel]{corbett2023measure}
Sam Corbett-Davies, Johann~D Gaebler, Hamed Nilforoshan, Ravi Shroff, and Sharad Goel.
\newblock The measure and mismeasure of fairness.
\newblock \emph{The Journal of Machine Learning Research}, 24\penalty0 (1):\penalty0 14730--14846, 2023.

\bibitem[Barocas et~al.(2023)Barocas, Hardt, and Narayanan]{barocas-hardt-narayanan}
Solon Barocas, Moritz Hardt, and Arvind Narayanan.
\newblock \emph{Fairness and Machine Learning: Limitations and Opportunities}.
\newblock MIT Press, 2023.

\bibitem[Ding et~al.(2021)Ding, Hardt, Miller, and Schmidt]{ding2021retiring}
Frances Ding, Moritz Hardt, John Miller, and Ludwig Schmidt.
\newblock Retiring adult: New datasets for fair machine learning.
\newblock \emph{Advances in Neural Information Processing Systems}, 34, 2021.

\bibitem[Flood et~al.(2018)Flood, King, Rodgers, Ruggles, and Warren]{pums2018}
Sarah Flood, Miriam King, Renae Rodgers, Steven Ruggles, and J~Robert Warren.
\newblock \emph{Integrated public use microdata series, Current Population Survey: Version 6.0}.
\newblock Minneapolis, MN: IPUMS, 2018.

\bibitem[Peterson et~al.(1954)Peterson, Birdsall, and Fox]{peterson1954theory}
WWTG Peterson, T~Birdsall, and We~Fox.
\newblock The theory of signal detectability.
\newblock \emph{Transactions of the IRE professional group on information theory}, 4\penalty0 (4):\penalty0 171--212, 1954.

\bibitem[Pasquale(2015)]{pasquale2015black}
Frank Pasquale.
\newblock \emph{The black box society: The secret algorithms that control money and information}.
\newblock Harvard University Press, 2015.

\bibitem[Eubanks(2018)]{eubanks2018automating}
Virginia Eubanks.
\newblock \emph{Automating inequality: How high-tech tools profile, police, and punish the poor}.
\newblock St. Martin's Press, 2018.

\bibitem[Benjamin(2019)]{benjamin2019race}
Ruha Benjamin.
\newblock \emph{Race after technology: Abolitionist tools for the new Jim code}.
\newblock John Wiley \& Sons, 2019.

\bibitem[Kasy and Abebe(2021)]{kasy2021fairness}
Maximilian Kasy and Rediet Abebe.
\newblock Fairness, equality, and power in algorithmic decision-making.
\newblock In \emph{Proceedings of the 2021 ACM Conference on Fairness, Accountability, and Transparency}, pages 576--586, 2021.

\bibitem[Perdomo et~al.(2023)Perdomo, Britton, Hardt, and Abebe]{perdomo2023difficult}
Juan~Carlos Perdomo, Tolani Britton, Moritz Hardt, and Rediet Abebe.
\newblock Difficult lessons on social prediction from wisconsin public schools.
\newblock \emph{arXiv preprint arXiv:2304.06205}, 2023.

\bibitem[Perdomo(2024)]{perdomo2024relative}
Juan~Carlos Perdomo.
\newblock The relative value of prediction in algorithmic decision making.
\newblock In \emph{International Conference on Machine Learning}, 2024.

\bibitem[Wang et~al.(2024)Wang, Kapoor, Barocas, and Narayanan]{wang2024against}
Angelina Wang, Sayash Kapoor, Solon Barocas, and Arvind Narayanan.
\newblock Against predictive optimization: On the legitimacy of decision-making algorithms that optimize predictive accuracy.
\newblock \emph{ACM Journal on Responsible Computing}, 1\penalty0 (1):\penalty0 1--45, 2024.

\bibitem[Hegselmann et~al.(2023)Hegselmann, Buendia, Lang, Agrawal, Jiang, and Sontag]{hegselmann23tabllm}
Stefan Hegselmann, Alejandro Buendia, Hunter Lang, Monica Agrawal, Xiaoyi Jiang, and David Sontag.
\newblock Tabllm: Few-shot classification of tabular data with large language models.
\newblock In \emph{Proceedings of The 26th International Conference on Artificial Intelligence and Statistics}, volume 206, pages 5549--5581, 2023.

\bibitem[Sanh et~al.(2021)Sanh, Webson, Raffel, Bach, Sutawika, Alyafeai, Chaffin, Stiegler, Scao, Raja, Dey, Bari, Xu, Thakker, Sharma, Szczechla, Kim, Chhablani, Nayak, Datta, Chang, Jiang, Wang, Manica, Shen, Yong, Pandey, Bawden, Wang, Neeraj, Rozen, Sharma, Santilli, Fevry, Fries, Teehan, Biderman, Gao, Bers, Wolf, and Rush]{sanh2021multitask}
Victor Sanh, Albert Webson, Colin Raffel, Stephen~H. Bach, Lintang Sutawika, Zaid Alyafeai, Antoine Chaffin, Arnaud Stiegler, Teven~Le Scao, Arun Raja, Manan Dey, M~Saiful Bari, Canwen Xu, Urmish Thakker, Shanya~Sharma Sharma, Eliza Szczechla, Taewoon Kim, Gunjan Chhablani, Nihal Nayak, Debajyoti Datta, Jonathan Chang, Mike Tian-Jian Jiang, Han Wang, Matteo Manica, Sheng Shen, Zheng~Xin Yong, Harshit Pandey, Rachel Bawden, Thomas Wang, Trishala Neeraj, Jos Rozen, Abheesht Sharma, Andrea Santilli, Thibault Fevry, Jason~Alan Fries, Ryan Teehan, Stella Biderman, Leo Gao, Tali Bers, Thomas Wolf, and Alexander~M. Rush.
\newblock Multitask prompted training enables zero-shot task generalization, 2021.

\bibitem[Slack and Singh(2023)]{tabletSlack23}
Dylan Slack and Sameer Singh.
\newblock {TABLET}: Learning from instructions for tabular data.
\newblock \emph{arXiv}, 2023.

\bibitem[Argyle et~al.(2023)Argyle, Busby, Fulda, Gubler, Rytting, and Wingate]{Argyle_2023}
Lisa~P. Argyle, Ethan~C. Busby, Nancy Fulda, Joshua~R. Gubler, Christopher Rytting, and David Wingate.
\newblock Out of one, many: Using language models to simulate human samples.
\newblock \emph{Political Analysis}, 31\penalty0 (3):\penalty0 337–351, 2023.

\bibitem[Sanders et~al.(2023)Sanders, Ulinich, and Schneier]{sanders2023demonstrations}
Nathan~E Sanders, Alex Ulinich, and Bruce Schneier.
\newblock Demonstrations of the potential of ai-based political issue polling.
\newblock \emph{arXiv preprint arXiv:2307.04781}, 2023.

\bibitem[Horton(2023)]{horton2023large}
John~J Horton.
\newblock Large language models as simulated economic agents: What can we learn from homo silicus?
\newblock Technical report, National Bureau of Economic Research, 2023.

\bibitem[Brand et~al.(2023)Brand, Israeli, and Ngwe]{brand2023using}
James Brand, Ayelet Israeli, and Donald Ngwe.
\newblock Using gpt for market research.
\newblock \emph{Harvard Business School Marketing Unit Working Paper}, \penalty0 (23-062), 2023.

\bibitem[Aher et~al.(2023)Aher, Arriaga, and Kalai]{aher2023using}
Gati~V Aher, Rosa~I Arriaga, and Adam~Tauman Kalai.
\newblock Using large language models to simulate multiple humans and replicate human subject studies.
\newblock In \emph{International Conference on Machine Learning}, pages 337--371. PMLR, 2023.

\bibitem[Dillion et~al.(2023)Dillion, Tandon, Gu, and Gray]{dillion2023can}
Danica Dillion, Niket Tandon, Yuling Gu, and Kurt Gray.
\newblock Can ai language models replace human participants?
\newblock \emph{Trends in Cognitive Sciences}, 27\penalty0 (7):\penalty0 597--600, 2023.

\bibitem[Brier(1950)]{brier50score}
Glenn~W. Brier.
\newblock Verification of forecasts expressed in terms of probability.
\newblock \emph{Monthly Weather Review}, 78\penalty0 (1):\penalty0 1 -- 3, 1950.

\bibitem[Murphy and Winkler(1977)]{murphy1977reliability}
Allan~H Murphy and Robert~L Winkler.
\newblock Reliability of subjective probability forecasts of precipitation and temperature.
\newblock \emph{Journal of the Royal Statistical Society Series C: Applied Statistics}, 26\penalty0 (1):\penalty0 41--47, 1977.

\bibitem[DeGroot and Fienberg(1983)]{degroot1983comparison}
Morris~H DeGroot and Stephen~E Fienberg.
\newblock The comparison and evaluation of forecasters.
\newblock \emph{Journal of the Royal Statistical Society: Series D (The Statistician)}, 32\penalty0 (1-2):\penalty0 12--22, 1983.

\bibitem[Platt et~al.(1999)]{platt1999probabilistic}
John Platt et~al.
\newblock Probabilistic outputs for support vector machines and comparisons to regularized likelihood methods.
\newblock \emph{Advances in large margin classifiers}, 10\penalty0 (3):\penalty0 61--74, 1999.

\bibitem[Zadrozny and Elkan(2001)]{zadrozny2001obtaining}
Bianca Zadrozny and Charles Elkan.
\newblock Obtaining calibrated probability estimates from decision trees and naive bayesian classifiers.
\newblock In \emph{International Conference on Machine Learning}, page 609–616, 2001.

\bibitem[Guo et~al.(2017)Guo, Pleiss, Sun, and Weinberger]{guo2017calibration}
Chuan Guo, Geoff Pleiss, Yu~Sun, and Kilian~Q. Weinberger.
\newblock On calibration of modern neural networks.
\newblock In \emph{International Conference on Machine Learning}, 2017.

\bibitem[Kadavath et~al.(2022)Kadavath, Conerly, Askell, Henighan, Drain, Perez, Schiefer, Hatfield-Dodds, DasSarma, Tran-Johnson, et~al.]{kadavath2022language}
Saurav Kadavath, Tom Conerly, Amanda Askell, Tom Henighan, Dawn Drain, Ethan Perez, Nicholas Schiefer, Zac Hatfield-Dodds, Nova DasSarma, Eli Tran-Johnson, et~al.
\newblock Language models (mostly) know what they know.
\newblock \emph{arXiv preprint arXiv:2207.05221}, 2022.

\bibitem[Xiao et~al.(2022)Xiao, Liang, Bhatt, Neiswanger, Salakhutdinov, and Morency]{xiao22plm}
Yuxin Xiao, Paul~Pu Liang, Umang Bhatt, Willie Neiswanger, Ruslan Salakhutdinov, and Louis-Philippe Morency.
\newblock Uncertainty quantification with pre-trained language models: A large-scale empirical analysis.
\newblock In \emph{Findings of the Association for Computational Linguistics: EMNLP 2022}, pages 7273--7284, 2022.

\bibitem[Jiang et~al.(2021)Jiang, Araki, Ding, and Neubig]{jiang21cali}
Zhengbao Jiang, Jun Araki, Haibo Ding, and Graham Neubig.
\newblock How can we know when language models know? on the calibration of language models for question answering.
\newblock \emph{Transactions of the Association for Computational Linguistics}, 9:\penalty0 962--977, 2021.

\bibitem[Xiong et~al.(2024)Xiong, Hu, Lu, LI, Fu, He, and Hooi]{xiong2024can}
Miao Xiong, Zhiyuan Hu, Xinyang Lu, YIFEI LI, Jie Fu, Junxian He, and Bryan Hooi.
\newblock Can {LLM}s express their uncertainty? an empirical evaluation of confidence elicitation in {LLM}s.
\newblock In \emph{The Twelfth International Conference on Learning Representations}, 2024.

\bibitem[Chen et~al.(2023)Chen, Yuan, Cui, Liu, and Ji]{chen2023close}
Yangyi Chen, Lifan Yuan, Ganqu Cui, Zhiyuan Liu, and Heng Ji.
\newblock A close look into the calibration of pre-trained language models.
\newblock In \emph{61st Annual Meeting of the Association for Computational Linguistics, ACL 2023}, pages 1343--1367, 2023.

\bibitem[Hendrycks et~al.(2021)Hendrycks, Burns, Basart, Zou, Mazeika, Song, and Steinhardt]{hendrycks2021measuring}
Dan Hendrycks, Collin Burns, Steven Basart, Andy Zou, Mantas Mazeika, Dawn Song, and Jacob Steinhardt.
\newblock Measuring massive multitask language understanding.
\newblock In \emph{International Conference on Learning Representations}, 2021.

\bibitem[Zellers et~al.(2019)Zellers, Holtzman, Bisk, Farhadi, and Choi]{zellers2019hellaswag}
Rowan Zellers, Ari Holtzman, Yonatan Bisk, Ali Farhadi, and Yejin Choi.
\newblock Hellaswag: Can a machine really finish your sentence?
\newblock In \emph{Proceedings of the 57th Annual Meeting of the Association for Computational Linguistics}, 2019.

\bibitem[Clark et~al.(2018)Clark, Cowhey, Etzioni, Khot, Sabharwal, Schoenick, and Tafjord]{clark2018think}
Peter Clark, Isaac Cowhey, Oren Etzioni, Tushar Khot, Ashish Sabharwal, Carissa Schoenick, and Oyvind Tafjord.
\newblock Think you have solved question answering? try arc, the ai2 reasoning challenge.
\newblock \emph{arXiv preprint arXiv:1803.05457}, 2018.

\bibitem[Cobbe et~al.(2021)Cobbe, Kosaraju, Bavarian, Chen, Jun, Kaiser, Plappert, Tworek, Hilton, Nakano, et~al.]{cobbe2021training}
Karl Cobbe, Vineet Kosaraju, Mohammad Bavarian, Mark Chen, Heewoo Jun, Lukasz Kaiser, Matthias Plappert, Jerry Tworek, Jacob Hilton, Reiichiro Nakano, et~al.
\newblock Training verifiers to solve math word problems.
\newblock \emph{arXiv preprint arXiv:2110.14168}, 2021.

\bibitem[Zhang et~al.(2024)Zhang, Zhang, Yu, Madeka, Foster, Xing, Lakkaraju, and Kakade]{zhang2024study}
Hanlin Zhang, Yi-Fan Zhang, Yaodong Yu, Dhruv Madeka, Dean Foster, Eric Xing, Himabindu Lakkaraju, and Sham Kakade.
\newblock A study on the calibration of in-context learning.
\newblock \emph{Arxiv preprint arxiv:2312.04021}, 2024.

\bibitem[Groves et~al.(2009)Groves, Fowler, Couper, Lepkowski, Singer, and Tourangeau]{groves2009survey}
R.M. Groves, F.J. Fowler, M.P. Couper, J.M. Lepkowski, E.~Singer, and R.~Tourangeau.
\newblock \emph{Survey Methodology}.
\newblock Wiley, 2009.

\bibitem[Santurkar et~al.(2023)Santurkar, Durmus, Ladhak, Lee, Liang, and Hashimoto]{santurkar23opinion}
Shibani Santurkar, Esin Durmus, Faisal Ladhak, Cinoo Lee, Percy Liang, and Tatsunori Hashimoto.
\newblock Whose opinions do language models reflect?
\newblock In \emph{Proceedings of the 40th International Conference on Machine Learning}, 2023.

\bibitem[Durmus et~al.(2023)Durmus, Nyugen, Liao, Schiefer, Askell, Bakhtin, Chen, Hatfield-Dodds, Hernandez, Joseph, et~al.]{durmus2023towards}
Esin Durmus, Karina Nyugen, Thomas~I Liao, Nicholas Schiefer, Amanda Askell, Anton Bakhtin, Carol Chen, Zac Hatfield-Dodds, Danny Hernandez, Nicholas Joseph, et~al.
\newblock Towards measuring the representation of subjective global opinions in language models.
\newblock \emph{arXiv preprint arXiv:2306.16388}, 2023.

\bibitem[Dominguez-Olmedo et~al.(2024)Dominguez-Olmedo, Hardt, and Mendler-D{\"u}nner]{dominguez2023questioning}
Ricardo Dominguez-Olmedo, Moritz Hardt, and Celestine Mendler-D{\"u}nner.
\newblock Questioning the survey responses of large language models.
\newblock In \emph{ICLR 2024 Workshop on Reliable and Responsible Foundation Models}, 2024.

\bibitem[Scherrer et~al.(2023)Scherrer, Shi, Feder, and Blei]{scherrer23belief}
Nino Scherrer, Claudia Shi, Amir Feder, and David Blei.
\newblock Evaluating the moral beliefs encoded in {LLMs}.
\newblock In \emph{Advances in Neural Information Processing Systems}, volume~36, pages 51778--51809, 2023.

\bibitem[Tjuatja et~al.(2024)Tjuatja, Chen, Wu, Talwalkar, and Neubig]{tjuatja2024llms}
Lindia Tjuatja, Valerie Chen, Sherry~Tongshuang Wu, Ameet Talwalkar, and Graham Neubig.
\newblock Do {LLMs} exhibit human-like response biases? a case study in survey design.
\newblock \emph{Arxiv preprint arxiv:2311.04076}, 2024.

\bibitem[Ott et~al.(2018)Ott, Auli, Grangier, and Ranzato]{ott2018analyzing}
Myle Ott, Michael Auli, David Grangier, and Marc’Aurelio Ranzato.
\newblock Analyzing uncertainty in neural machine translation.
\newblock In \emph{International Conference on Machine Learning}, pages 3956--3965. PMLR, 2018.

\bibitem[Kalai and Vempala(2024)]{kalai2024calibrated}
Adam~Tauman Kalai and Santosh~S. Vempala.
\newblock Calibrated language models must hallucinate.
\newblock \emph{Arxiv preprint arxiv:2311.14648}, 2024.

\bibitem[Lin et~al.(2022)Lin, Hilton, and Evans]{lin2022teaching}
Stephanie Lin, Jacob Hilton, and Owain Evans.
\newblock Teaching models to express their uncertainty in words.
\newblock \emph{Transactions on Machine Learning Research}, 2022.
\newblock ISSN 2835-8856.

\bibitem[Mielke et~al.(2022)Mielke, Szlam, Dinan, and Boureau]{mielke22ling}
Sabrina~J. Mielke, Arthur Szlam, Emily Dinan, and Y-Lan Boureau.
\newblock Reducing conversational agents{'} overconfidence through linguistic calibration.
\newblock \emph{Transactions of the Association for Computational Linguistics}, 10:\penalty0 857--872, 2022.

\bibitem[Band et~al.(2024)Band, Li, Ma, and Hashimoto]{band2024linguistic}
Neil Band, Xuechen Li, Tengyu Ma, and Tatsunori Hashimoto.
\newblock Linguistic calibration of language models.
\newblock \emph{arXiv preprint arXiv:2404.00474}, 2024.

\bibitem[Lin et~al.(2024)Lin, Trivedi, and Sun]{lin2024generating}
Zhen Lin, Shubhendu Trivedi, and Jimeng Sun.
\newblock Generating with confidence: Uncertainty quantification for black-box large language models, 2024.

\bibitem[Cleary(1968)]{cleary1968test}
T~Anne Cleary.
\newblock Test bias: Prediction of grades of negro and white students in integrated colleges.
\newblock \emph{Journal of Educational Measurement}, 5\penalty0 (2):\penalty0 115--124, 1968.

\bibitem[Hutchinson and Mitchell(2019)]{hutchinson201950}
Ben Hutchinson and Margaret Mitchell.
\newblock 50 years of test (un) fairness: Lessons for machine learning.
\newblock In \emph{Proceedings of the conference on fairness, accountability, and transparency}, pages 49--58, 2019.

\bibitem[Hebert-Johnson et~al.(2018)Hebert-Johnson, Kim, Reingold, and Rothblum]{hebert18multical}
Ursula Hebert-Johnson, Michael Kim, Omer Reingold, and Guy Rothblum.
\newblock Multicalibration: Calibration for the ({C}omputationally-identifiable) masses.
\newblock In \emph{Proceedings of the 35th International Conference on Machine Learning}, volume~80, pages 1939--1948, 2018.

\bibitem[Pakdaman~Naeini et~al.(2015)Pakdaman~Naeini, Cooper, and Hauskrecht]{Pakdaman15cali}
Mahdi Pakdaman~Naeini, Gregory Cooper, and Milos Hauskrecht.
\newblock Obtaining well calibrated probabilities using bayesian binning.
\newblock \emph{Proceedings of the AAAI Conference on Artificial Intelligence}, 29\penalty0 (1), 2015.

\bibitem[Der~Kiureghian and Ditlevsen(2009)]{der2009aleatory}
Armen Der~Kiureghian and Ove Ditlevsen.
\newblock Aleatory or epistemic? does it matter?
\newblock \emph{Structural safety}, 31\penalty0 (2):\penalty0 105--112, 2009.

\bibitem[Fang et~al.(2024)Fang, Xu, Tan, Zhang, Hu, Qi, Nickleach, Socolinsky, Sengamedu, and Faloutsos]{fang2024large}
Xi~Fang, Weijie Xu, Fiona~Anting Tan, Jiani Zhang, Ziqing Hu, Yanjun Qi, Scott Nickleach, Diego Socolinsky, Srinivasan Sengamedu, and Christos Faloutsos.
\newblock Large language models({LLMs}) on tabular data: Prediction, generation, and understanding -- a survey.
\newblock \emph{Arxiv preprint arxiv:2402.17944}, 2024.

\bibitem[Yu et~al.(2023)Yu, Fu, Yu, Huang, and Li]{yu2023unified}
Bowen Yu, Cheng Fu, Haiyang Yu, Fei Huang, and Yongbin Li.
\newblock Unified language representation for question answering over text, tables, and images.
\newblock In Anna Rogers, Jordan Boyd-Graber, and Naoaki Okazaki, editors, \emph{Findings of the Association for Computational Linguistics: ACL 2023}, pages 4756--4765, 2023.

\bibitem[Gong et~al.(2020)Gong, Sun, Feng, Qin, Bi, Liu, and Liu]{gong20tablegpt}
Heng Gong, Yawei Sun, Xiaocheng Feng, Bing Qin, Wei Bi, Xiaojiang Liu, and Ting Liu.
\newblock {T}able{GPT}: Few-shot table-to-text generation with table structure reconstruction and content matching.
\newblock In \emph{Proceedings of the 28th International Conference on Computational Linguistics}, pages 1978--1988, 2020.

\bibitem[Tian et~al.(2023)Tian, Mitchell, Zhou, Sharma, Rafailov, Yao, Finn, and Manning]{tian2023just}
Katherine Tian, Eric Mitchell, Allan Zhou, Archit Sharma, Rafael Rafailov, Huaxiu Yao, Chelsea Finn, and Christopher~D Manning.
\newblock Just ask for calibration: Strategies for eliciting calibrated confidence scores from language models fine-tuned with human feedback.
\newblock In \emph{The 2023 Conference on Empirical Methods in Natural Language Processing}, 2023.
\newblock URL \url{https://openreview.net/forum?id=g3faCfrwm7}.

\bibitem[Robinson and Wingate(2023)]{robinson2023leveraging}
Joshua Robinson and David Wingate.
\newblock Leveraging large language models for multiple choice question answering.
\newblock In \emph{The Eleventh International Conference on Learning Representations}, 2023.

\bibitem[AI@Meta(2024)]{llama3modelcard}
AI@Meta.
\newblock Llama 3 model card.
\newblock 2024.
\newblock URL \url{https://github.com/meta-llama/llama3/blob/main/MODEL_CARD.md}.

\bibitem[Jiang et~al.(2023)Jiang, Sablayrolles, Mensch, Bamford, Chaplot, de~las Casas, Bressand, Lengyel, Lample, Saulnier, Lavaud, Lachaux, Stock, Scao, Lavril, Wang, Lacroix, and Sayed]{jiang2023mistral}
Albert~Q. Jiang, Alexandre Sablayrolles, Arthur Mensch, Chris Bamford, Devendra~Singh Chaplot, Diego de~las Casas, Florian Bressand, Gianna Lengyel, Guillaume Lample, Lucile Saulnier, Lélio~Renard Lavaud, Marie-Anne Lachaux, Pierre Stock, Teven~Le Scao, Thibaut Lavril, Thomas Wang, Timothée Lacroix, and William~El Sayed.
\newblock Mistral 7b.
\newblock \emph{Arxiv preprint arxiv:2310.06825}, 2023.

\bibitem[Jiang et~al.(2024)Jiang, Sablayrolles, Roux, Mensch, Savary, Bamford, Chaplot, Casas, Hanna, Bressand, et~al.]{jiang2024mixtral}
Albert~Q Jiang, Alexandre Sablayrolles, Antoine Roux, Arthur Mensch, Blanche Savary, Chris Bamford, Devendra~Singh Chaplot, Diego de~las Casas, Emma~Bou Hanna, Florian Bressand, et~al.
\newblock Mixtral of experts.
\newblock \emph{arXiv preprint arXiv:2401.04088}, 2024.

\bibitem[Young et~al.(2024)Young, Chen, Li, Huang, Zhang, Zhang, Li, Zhu, Chen, Chang, et~al.]{young2024yi}
Alex Young, Bei Chen, Chao Li, Chengen Huang, Ge~Zhang, Guanwei Zhang, Heng Li, Jiangcheng Zhu, Jianqun Chen, Jing Chang, et~al.
\newblock Yi: Open foundation models by 01. ai.
\newblock \emph{arXiv preprint arXiv:2403.04652}, 2024.

\bibitem[Team et~al.(2024)Team, Mesnard, Hardin, Dadashi, Bhupatiraju, Pathak, Sifre, Rivière, Kale, Love, Tafti, Hussenot, Sessa, Chowdhery, Roberts, Barua, Botev, Castro-Ros, Slone, Héliou, Tacchetti, Bulanova, Paterson, Tsai, Shahriari, Lan, Choquette-Choo, Crepy, Cer, Ippolito, Reid, Buchatskaya, Ni, Noland, Yan, Tucker, Muraru, Rozhdestvenskiy, Michalewski, Tenney, Grishchenko, Austin, Keeling, Labanowski, Lespiau, Stanway, Brennan, Chen, Ferret, Chiu, Mao-Jones, Lee, Yu, Millican, Sjoesund, Lee, Dixon, Reid, Mikuła, Wirth, Sharman, Chinaev, Thain, Bachem, Chang, Wahltinez, Bailey, Michel, Yotov, Chaabouni, Comanescu, Jana, Anil, McIlroy, Liu, Mullins, Smith, Borgeaud, Girgin, Douglas, Pandya, Shakeri, De, Klimenko, Hennigan, Feinberg, Stokowiec, hui Chen, Ahmed, Gong, Warkentin, Peran, Giang, Farabet, Vinyals, Dean, Kavukcuoglu, Hassabis, Ghahramani, Eck, Barral, Pereira, Collins, Joulin, Fiedel, Senter, Andreev, and Kenealy]{gemmateam2024gemma}
Gemma Team, Thomas Mesnard, Cassidy Hardin, Robert Dadashi, Surya Bhupatiraju, Shreya Pathak, Laurent Sifre, Morgane Rivière, Mihir~Sanjay Kale, Juliette Love, Pouya Tafti, Léonard Hussenot, Pier~Giuseppe Sessa, Aakanksha Chowdhery, Adam Roberts, Aditya Barua, Alex Botev, Alex Castro-Ros, Ambrose Slone, Amélie Héliou, Andrea Tacchetti, Anna Bulanova, Antonia Paterson, Beth Tsai, Bobak Shahriari, Charline~Le Lan, Christopher~A. Choquette-Choo, Clément Crepy, Daniel Cer, Daphne Ippolito, David Reid, Elena Buchatskaya, Eric Ni, Eric Noland, Geng Yan, George Tucker, George-Christian Muraru, Grigory Rozhdestvenskiy, Henryk Michalewski, Ian Tenney, Ivan Grishchenko, Jacob Austin, James Keeling, Jane Labanowski, Jean-Baptiste Lespiau, Jeff Stanway, Jenny Brennan, Jeremy Chen, Johan Ferret, Justin Chiu, Justin Mao-Jones, Katherine Lee, Kathy Yu, Katie Millican, Lars~Lowe Sjoesund, Lisa Lee, Lucas Dixon, Machel Reid, Maciej Mikuła, Mateo Wirth, Michael Sharman, Nikolai Chinaev, Nithum Thain, Olivier Bachem,
  Oscar Chang, Oscar Wahltinez, Paige Bailey, Paul Michel, Petko Yotov, Rahma Chaabouni, Ramona Comanescu, Reena Jana, Rohan Anil, Ross McIlroy, Ruibo Liu, Ryan Mullins, Samuel~L Smith, Sebastian Borgeaud, Sertan Girgin, Sholto Douglas, Shree Pandya, Siamak Shakeri, Soham De, Ted Klimenko, Tom Hennigan, Vlad Feinberg, Wojciech Stokowiec, Yu~hui Chen, Zafarali Ahmed, Zhitao Gong, Tris Warkentin, Ludovic Peran, Minh Giang, Clément Farabet, Oriol Vinyals, Jeff Dean, Koray Kavukcuoglu, Demis Hassabis, Zoubin Ghahramani, Douglas Eck, Joelle Barral, Fernando Pereira, Eli Collins, Armand Joulin, Noah Fiedel, Evan Senter, Alek Andreev, and Kathleen Kenealy.
\newblock Gemma: Open models based on gemini research and technology.
\newblock \emph{Arxiv preprint arxiv:2403.08295}, 2024.

\bibitem[Achiam et~al.(2023)Achiam, Adler, Agarwal, Ahmad, Akkaya, Aleman, Almeida, Altenschmidt, Altman, Anadkat, et~al.]{achiam2023gpt}
Josh Achiam, Steven Adler, Sandhini Agarwal, Lama Ahmad, Ilge Akkaya, Florencia~Leoni Aleman, Diogo Almeida, Janko Altenschmidt, Sam Altman, Shyamal Anadkat, et~al.
\newblock {GPT}-4 technical report.
\newblock \emph{arXiv preprint arXiv:2303.08774}, 2023.

\bibitem[Chen and Guestrin(2016)]{chen2016xgboost}
Tianqi Chen and Carlos Guestrin.
\newblock {XGBoost}: A scalable tree boosting system.
\newblock In \emph{Proceedings of the 22nd ACM SIGKDD International Conference on Knowledge Discovery and Data Mining}, KDD '16, page 785–794, New York, NY, USA, 2016. Association for Computing Machinery.
\newblock ISBN 9781450342322.
\newblock \doi{10.1145/2939672.2939785}.
\newblock URL \url{https://doi.org/10.1145/2939672.2939785}.

\bibitem[Borisov et~al.(2024)Borisov, Leemann, Seßler, Haug, Pawelczyk, and Kasneci]{borisov2024deep}
Vadim Borisov, Tobias Leemann, Kathrin Seßler, Johannes Haug, Martin Pawelczyk, and Gjergji Kasneci.
\newblock Deep neural networks and tabular data: A survey.
\newblock \emph{IEEE Transactions on Neural Networks and Learning Systems}, 35\penalty0 (6):\penalty0 7499--7519, 2024.
\newblock \doi{10.1109/TNNLS.2022.3229161}.

\bibitem[Becker and Kohavi(1996)]{misc_adult_2}
Barry Becker and Ronny Kohavi.
\newblock {Adult}.
\newblock UCI Machine Learning Repository, 1996.
\newblock {DOI}: https://doi.org/10.24432/C5XW20.

\bibitem[Breiman(2001)]{breiman2001random}
Leo Breiman.
\newblock Random forests.
\newblock \emph{Machine learning}, 45:\penalty0 5--32, 2001.

\end{thebibliography}
\bibliographystyle{unsrtnat}

%% NeurIPS checklist
% \newpage
% \input{paper-checklist}

%%%%%%%%%%%%%%%%%%%%%%%%%%%%%%%%%%%%%%%%%%%%%%%%%%%%%%%%%%%%

% Set appendix Figures/Tables as "A1, A2, ..."
\setcounter{table}{0}
\renewcommand{\thetable}{A\arabic{table}}
\setcounter{figure}{0}
\renewcommand{\thefigure}{A\arabic{figure}}

% Appendix
\clearpage
\appendix
\appendixpage
\section{Additional results}
\label{app:additional_results}

This appendix section shows additional results and corresponding plots to support the insights presented in Section~\ref{sec:results}.
Section~\ref{app:numeric_prompting} shows results using a chat-style verbalized numeric prompting scheme.
Section~\ref{app:results_on_additional_tasks} shows results on four extra benchmark tasks made available with \folktexts.
Section~\ref{app:additional_subgroup_cali_plots} extends the discussion on subgroup calibration and algorithmic fairness on the ACSIncome task.
Section~\ref{app:varying_uncertainty} goes further in-depth on how to use \folktexts to control data uncertainty in the benchmark prediction tasks.
Finally, Section~\ref{app:feature_importance} presents and discusses results on feature importance for LLM predictions.

\subsection{Additional results using verbalized numeric prompting}
\label{app:numeric_prompting}

\begin{figure}[p]
    \centering
    \includegraphics[width=0.89\linewidth]{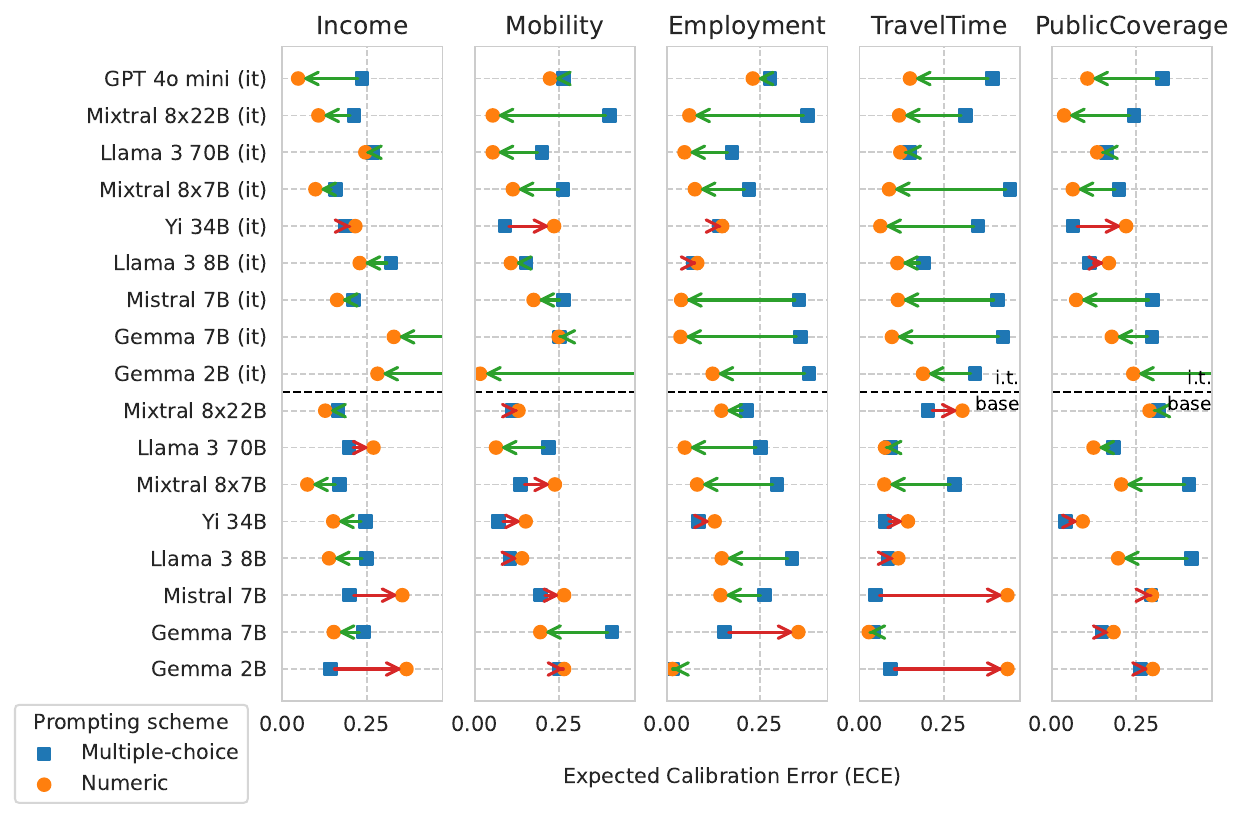}
    \caption{Change in calibration error (ECE) when using numeric risk prompting ({\color{Orange}$\bullet$}) versus multiple-choice prompting ({\footnotesize\color{NavyBlue}$\blacksquare$}).
    Instruction-tuned models (\textit{top rows}) show substantial calibration improvements, while base models (\textit{bottom rows}) show mixed results.
    Green/red arrows signal ECE improvement/degradation.
    }
    \label{fig:diff_prompts_ece_change}
\end{figure}

\begin{figure}[p]
    \centering
    \includegraphics[width=0.89\linewidth]{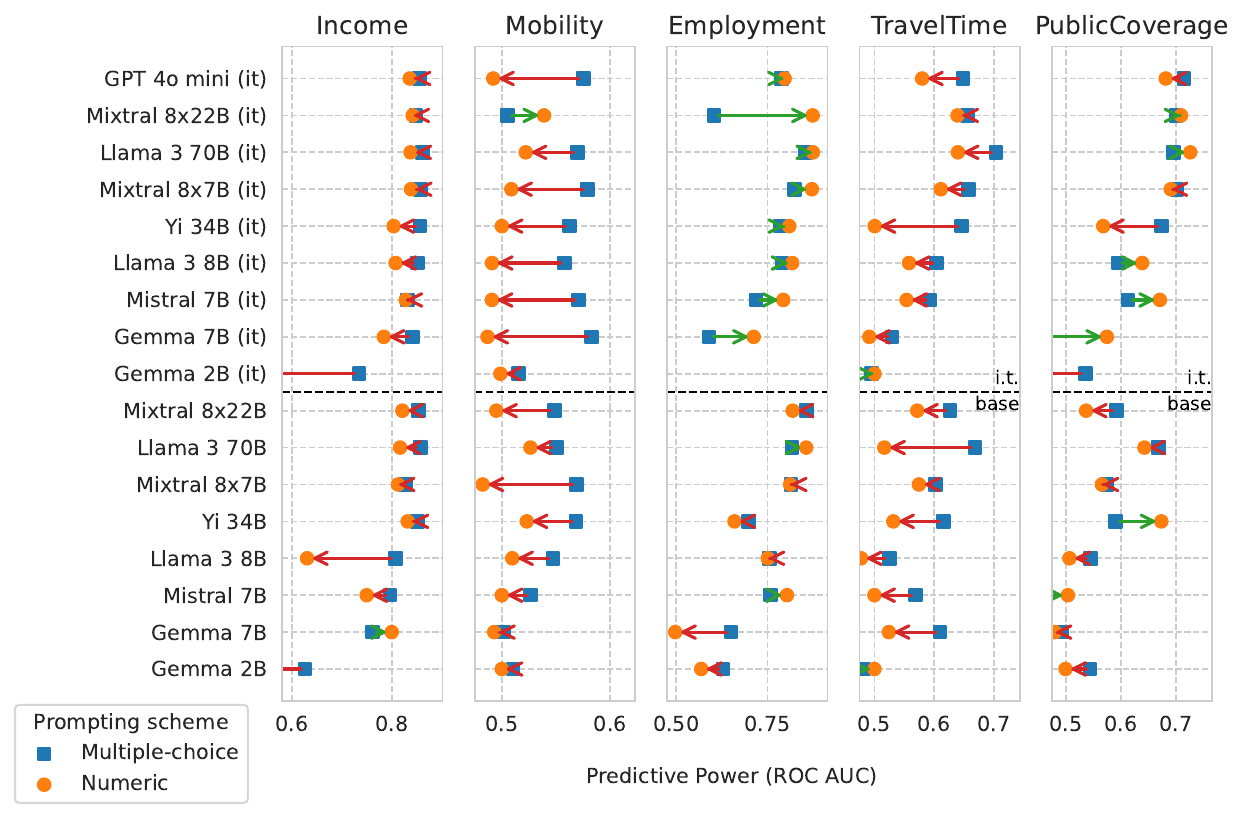}
    \caption{Change in predictive power (AUC) when using numeric risk prompting ({\color{Orange}$\bullet$}) versus multiple-choice prompting ({\footnotesize\color{NavyBlue}$\blacksquare$}).
    Both instruction-tuned models (\textit{top rows}) and base models (\textit{bottom rows}) generally achieve worse AUC with numeric prompting. Green/red arrows signal AUC improvement/degradation, respectively.
    }
    \label{fig:diff_prompts_auc_change}
\end{figure}

\begin{figure}[tbp]
    \centering
    \includegraphics[width=0.98\linewidth]{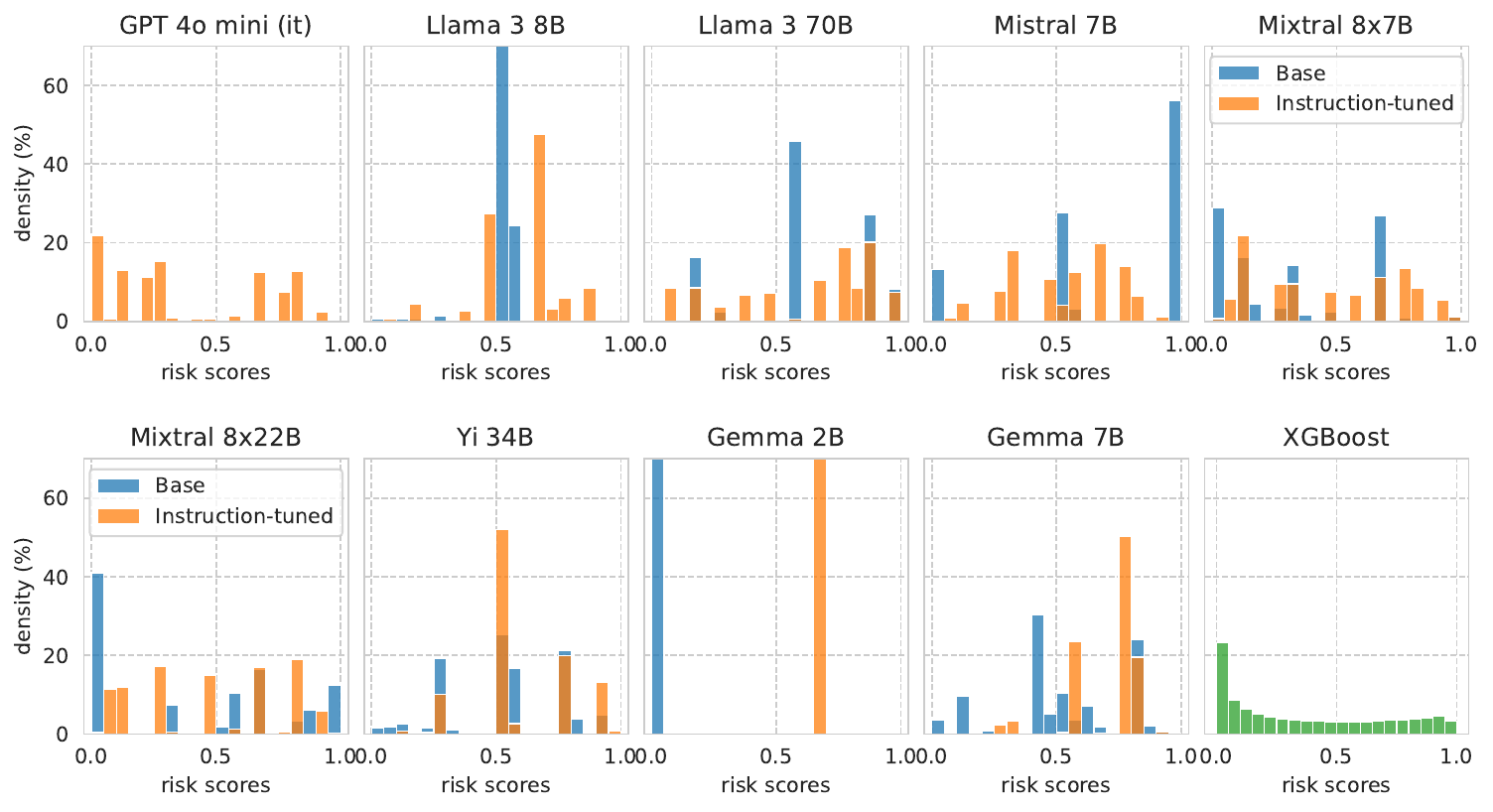}
    \caption{Distribution of risk scores produced using \textit{numeric prompting} on the ACSIncome benchmark task. A baseline score distribution that achieves $0.00$ calibration error is shown in green (XGBoost model).
    For each model, risk scores produced in this manner fall into only a few different possible values, contrasting with the neatly continuous distribution produced by multiple-choice prompting.
    This fact leads to numerous more ties among predicted risk scores, which can explain the reduced AUC performance with this prompting scheme. Nonetheless, calibration error is considerably smaller for instruction-tuned models. %, and for most base models.
    }
    \label{fig:score-dist-numeric-prompt}
\end{figure}

Figure~\ref{fig:diff_prompts_ece_change} shows the change in calibration error (ECE) between using multiple-choice prompting and verbalized numeric prompting, on all five benchmark tasks.
Instruction-tuned models (top rows) show ECE improvements on an overwhelming majority of model/task pairs, while base models (bottom rows) show less consistent results.
However, using numeric prompting comes at a consistent cost of diminished predictive power (AUC) of the risk scores, shown in Figure~\ref{fig:diff_prompts_auc_change}.
A majority of model/task pairs have worse AUC with numeric prompting, with the exception of the employment prediction task.
One potential explanation for this generalized decrease in AUC lies in the fact that numeric prompting generates a large number of tied risk scores.
Figure~\ref{fig:score-dist-numeric-prompt} shows the score distribution of all models using numeric prompting (compare with multiple-choice prompting shown in Figure~\ref{fig:score-dist}).
While multiple-choice prompting produces a smoother continuous score distribution, numeric prompting generally results in a small set of possible uncertainty estimates.
This arguably makes intuitive sense, as numeric prompting produces uncertainty estimates in discrete token space, while multiple-choice prompting produces uncertainty estimates in the continuous token-probability space.

\subsection{Results on additional benchmark tasks}
\label{app:results_on_additional_tasks}

The main body of the paper focuses on results on the ACSIncome prediction task.
This task is arguably the most popular for benchmarking on tabular data, as it closely mirrors the older but widely used UCI Adult dataset~\citep{ding2021retiring,misc_adult_2}.
The \folktexts package makes available natural-language versions of four additional tabular data tasks: ACSEmployment, ACSMobility, ACSTravelTime, and ACSPublicCoverage.

Tables~\ref{tab:acsemployment_results}--\ref{tab:acspubliccoverage_results} show results for the ACSEmployment, ACSMobility, ACSTravelTime, and ACSPublicCoverage tasks, respectively.
Trends discussed in the main body of the paper are broadly confirmed.
Models' moderate predictive performance is accompanied by substantial miscalibration of their risk scores.
Additionally, base models output low-variance high-uncertainty score distributions, while instruction-tuned models output high-variance low-uncertainty score distributions.
There is clear predictive signal for large models across all tasks (e.g., see the AUC of Llama 3 70B it, or Mixtral 8x22B it).
However, the extent to which models' scores are predictable varies substantially across tasks.
Most models surpass the AUC of the linear baseline (LR) on the ACSTravelTime task, as well as the main ACSIncome task; but consistently lag behind the linear baseline on the ACSMobility task.
On the ACSEmployment and ACSPublicCoverage tasks, the best performing models manage to match the linear baseline AUC.

These findings pose into question one of the main advantages of using LLMs for risk scoring: the fact that no labeled data is required.
Given the inconsistency of model performance, some small amount of testing data may always be needed to assert reliability of results.

% \begin{table}[p]
% \centering
% {\footnotesize \input{tables/both_prompting_schemes_same_rows/acsincome}}
% \caption{Zero-shot LLM results on the \textbf{ACSIncome} benchmark task, together with supervised learning baselines fitted on 1.6M samples. Rows are sorted by model size, measured by the number of active parameters.}
% \label{tab:acsincome_results}
% \end{table}

\begin{table}[p]
\centering
{\footnotesize \begin{tabular}{l|cccc|cccc}
\toprule
\multirow{3}{*}{\textbf{Model}} & \multicolumn{4}{c|}{\textbf{Multiple-choice prompting}} & \multicolumn{4}{c}{\textbf{Numeric risk prompting}} \\
 & \multirow{2}{*}{\textbf{ECE $\downarrow$}} & \multirow{2}{*}{\textbf{\shortstack{Brier \\score}} $\downarrow$} & \multirow{2}{*}{\textbf{\shortstack{AUC} $\uparrow$}} & \multirow{2}{*}{\textbf{\shortstack{Acc.} $\uparrow$}} & \multirow{2}{*}{\textbf{ECE $\downarrow$}} & \multirow{2}{*}{\textbf{\shortstack{Brier \\score}} $\downarrow$} & \multirow{2}{*}{\textbf{\shortstack{AUC} $\uparrow$}} & \multirow{2}{*}{\textbf{\shortstack{Acc.} $\uparrow$}} \\
 & & & & & & & & \\
\midrule
GPT 4o mini (it) & 0.28 & 0.29 & 0.79 & 0.65 & 0.23 & 0.23 & 0.80 & 0.73 \\
Mixtral 8x22B (it) & \cellcolor{orange!23.0} 0.38 & \cellcolor{orange!8.0} 0.39 & 0.60 & 0.51 & 0.06 & \cellcolor{cyan!21.9} 0.14 & \cellcolor{cyan!24.4} 0.87 & \cellcolor{cyan!16.5} 0.79 \\
Mixtral 8x22B & 0.21 & 0.24 & \cellcolor{cyan!25.0} 0.86 & 0.52 & 0.15 & 0.18 & 0.82 & \cellcolor{cyan!20.7} 0.80 \\
Llama 3 70B (it) & 0.17 & \cellcolor{cyan!17.1} 0.19 & \cellcolor{cyan!23.3} 0.85 & \cellcolor{cyan!15.4} 0.73 & \cellcolor{cyan!0.7} 0.05 & \cellcolor{cyan!25.0} 0.14 & \cellcolor{cyan!25.0} 0.88 & \cellcolor{cyan!25.0} 0.81 \\
Llama 3 70B & 0.25 & 0.26 & \cellcolor{cyan!2.3} 0.82 & 0.52 & 0.05 & \cellcolor{cyan!16.7} 0.15 & \cellcolor{cyan!13.6} 0.86 & \cellcolor{cyan!5.8} 0.78 \\
Mixtral 8x7B (it) & 0.22 & 0.24 & \cellcolor{cyan!6.8} 0.82 & \cellcolor{cyan!13.7} 0.73 & 0.07 & \cellcolor{cyan!16.7} 0.15 & \cellcolor{cyan!23.2} 0.87 & \cellcolor{cyan!7.2} 0.78 \\
Mixtral 8x7B & 0.30 & 0.31 & \cellcolor{cyan!1.1} 0.81 & \cellcolor{orange!24.1} 0.45 & 0.08 & 0.17 & 0.81 & 0.73 \\
Yi 34B (it) & 0.14 & 0.21 & 0.79 & 0.69 & 0.15 & 0.21 & 0.81 & 0.51 \\
Yi 34B & 0.08 & 0.23 & 0.70 & 0.62 & 0.13 & 0.23 & 0.66 & 0.50 \\
Llama 3 8B (it) & 0.07 & \cellcolor{cyan!25.0} 0.19 & 0.79 & \cellcolor{cyan!25.0} 0.74 & 0.08 & 0.17 & 0.82 & 0.77 \\
Llama 3 8B & 0.34 & 0.34 & 0.76 & \cellcolor{orange!25.0} 0.45 & 0.15 & 0.23 & 0.75 & \cellcolor{orange!25.0} 0.46 \\
Mistral 7B (it) & \cellcolor{orange!6.6} 0.35 & 0.36 & 0.72 & 0.63 & \cellcolor{cyan!7.4} 0.04 & 0.19 & 0.79 & 0.69 \\
Mistral 7B & 0.26 & 0.30 & 0.76 & \cellcolor{orange!25.0} 0.45 & 0.14 & 0.19 & 0.80 & \cellcolor{cyan!12.9} 0.79 \\
Gemma 7B (it) & \cellcolor{orange!10.0} 0.36 & 0.38 & 0.59 & 0.58 & \cellcolor{cyan!8.8} 0.04 & 0.22 & 0.71 & 0.60 \\
Gemma 7B & 0.15 & 0.25 & 0.65 & \cellcolor{orange!4.1} 0.48 & \cellcolor{orange!25.0} 0.35 & \cellcolor{orange!25.0} 0.38 & \cellcolor{orange!1.6} 0.50 & 0.51 \\
Gemma 2B (it) & \cellcolor{orange!25.0} 0.38 & \cellcolor{orange!25.0} 0.41 & \cellcolor{orange!25.0} 0.42 & \cellcolor{orange!16.3} 0.46 & 0.12 & 0.27 & \cellcolor{orange!25.0} 0.46 & \cellcolor{orange!25.0} 0.46 \\
Gemma 2B & \cellcolor{cyan!25.0} 0.01 & 0.24 & 0.63 & 0.54 & \cellcolor{cyan!25.0} 0.01 & 0.23 & 0.57 & 0.53 \\
\midrule
LR & 0.02 & 0.15 & 0.86 & 0.78 & 0.02 & 0.15 & 0.86 & 0.78 \\
% GBM & 0.00 & 0.12 & 0.91 & 0.83 & 0.00 & 0.12 & 0.91 & 0.83 \\
XGBoost & 0.00 & 0.12 & 0.91 & 0.83 & 0.00 & 0.12 & 0.91 & 0.83 \\
\bottomrule
\end{tabular}}
\caption{Zero-shot LLM results on the \textbf{ACSEmployment} benchmark task, together with supervised learning baselines fitted on 2.9M samples.}
\label{tab:acsemployment_results}
\end{table}

\begin{table}[p]
\centering
{\footnotesize \begin{tabular}{l|cccc|cccc}
\toprule
\multirow{3}{*}{\textbf{Model}} & \multicolumn{4}{c|}{\textbf{Multiple-choice prompting}} & \multicolumn{4}{c}{\textbf{Numeric risk prompting}} \\
 & \multirow{2}{*}{\textbf{ECE $\downarrow$}} & \multirow{2}{*}{\textbf{\shortstack{Brier \\score}} $\downarrow$} & \multirow{2}{*}{\textbf{\shortstack{AUC} $\uparrow$}} & \multirow{2}{*}{\textbf{\shortstack{Acc.} $\uparrow$}} & \multirow{2}{*}{\textbf{ECE $\downarrow$}} & \multirow{2}{*}{\textbf{\shortstack{Brier \\score}} $\downarrow$} & \multirow{2}{*}{\textbf{\shortstack{AUC} $\uparrow$}} & \multirow{2}{*}{\textbf{\shortstack{Acc.} $\uparrow$}} \\
 & & & & & & & & \\
\midrule
GPT 4o mini (it) & 0.26 & 0.26 & \cellcolor{cyan!0.6} 0.57 & \cellcolor{cyan!25.0} 0.73 & 0.22 & 0.25 & 0.49 & \cellcolor{cyan!25.0} 0.73 \\
Mixtral 8x22B (it) & 0.40 & 0.40 & \cellcolor{orange!12.8} 0.51 & 0.39 & 0.05 & \cellcolor{cyan!17.9} 0.20 & \cellcolor{cyan!25.0} 0.54 & \cellcolor{cyan!25.0} 0.73 \\
Mixtral 8x22B & \cellcolor{cyan!10.2} 0.11 & \cellcolor{cyan!21.2} 0.21 & 0.55 & \cellcolor{cyan!25.0} 0.73 & 0.13 & 0.22 & 0.49 & \cellcolor{cyan!25.0} 0.73 \\
Llama 3 70B (it) & 0.20 & \cellcolor{cyan!1.5} 0.25 & 0.57 & 0.58 & 0.05 & \cellcolor{cyan!10.7} 0.20 & 0.52 & \cellcolor{cyan!25.0} 0.73 \\
Llama 3 70B & 0.22 & \cellcolor{cyan!4.8} 0.24 & 0.55 & 0.53 & 0.06 & \cellcolor{cyan!7.1} 0.20 & 0.53 & \cellcolor{cyan!20.7} 0.73 \\
Mixtral 8x7B (it) & 0.26 & 0.26 & \cellcolor{cyan!12.8} 0.58 & \cellcolor{cyan!25.0} 0.73 & 0.11 & 0.21 & 0.51 & \cellcolor{cyan!25.0} 0.73 \\
Mixtral 8x7B & \cellcolor{cyan!0.8} 0.14 & \cellcolor{cyan!18.9} 0.21 & 0.57 & \cellcolor{cyan!25.0} 0.73 & 0.24 & 0.25 & \cellcolor{orange!25.0} 0.48 & \cellcolor{cyan!25.0} 0.73 \\
Yi 34B (it) & \cellcolor{cyan!18.2} 0.09 & \cellcolor{cyan!23.1} 0.20 & 0.56 & \cellcolor{cyan!15.4} 0.72 & 0.23 & 0.25 & 0.50 & \cellcolor{orange!25.0} 0.27 \\
Yi 34B & \cellcolor{cyan!25.0} 0.07 & \cellcolor{cyan!25.0} 0.20 & 0.57 & \cellcolor{cyan!20.7} 0.73 & 0.15 & 0.23 & 0.52 & 0.44 \\
Llama 3 8B (it) & 0.15 & \cellcolor{cyan!13.7} 0.22 & 0.56 & \cellcolor{cyan!5.3} 0.70 & 0.11 & 0.21 & 0.49 & \cellcolor{cyan!23.4} 0.73 \\
Llama 3 8B & \cellcolor{cyan!12.5} 0.10 & \cellcolor{cyan!21.7} 0.20 & 0.55 & \cellcolor{cyan!24.5} 0.73 & 0.14 & 0.21 & 0.51 & \cellcolor{cyan!19.1} 0.72 \\
Mistral 7B (it) & 0.26 & 0.26 & 0.57 & \cellcolor{cyan!25.0} 0.73 & 0.17 & 0.23 & 0.49 & \cellcolor{cyan!24.5} 0.73 \\
Mistral 7B & 0.20 & \cellcolor{cyan!8.6} 0.23 & 0.53 & \cellcolor{cyan!23.4} 0.73 & \cellcolor{orange!25.0} 0.27 & \cellcolor{orange!25.0} 0.27 & 0.50 & \cellcolor{cyan!25.0} 0.73 \\
Gemma 7B (it) & 0.25 & 0.26 & \cellcolor{cyan!25.0} 0.58 & \cellcolor{cyan!21.8} 0.73 & \cellcolor{orange!8.9} 0.25 & 0.26 & \cellcolor{orange!3.1} 0.49 & \cellcolor{cyan!22.3} 0.73 \\
Gemma 7B & 0.41 & 0.37 & \cellcolor{orange!25.0} 0.50 & \cellcolor{orange!25.0} 0.27 & 0.19 & 0.24 & 0.49 & \cellcolor{cyan!25.0} 0.73 \\
Gemma 2B (it) & \cellcolor{orange!25.0} 0.73 & \cellcolor{orange!25.0} 0.73 & 0.52 & \cellcolor{orange!25.0} 0.27 & \cellcolor{cyan!25.0} 0.02 & \cellcolor{cyan!25.0} 0.20 & 0.50 & \cellcolor{cyan!25.0} 0.73 \\
Gemma 2B & 0.25 & 0.26 & 0.51 & 0.34 & \cellcolor{orange!25.0} 0.27 & \cellcolor{orange!25.0} 0.27 & 0.50 & \cellcolor{cyan!25.0} 0.73 \\
\midrule
LR & 0.02 & 0.19 & 0.61 & 0.74 & 0.02 & 0.19 & 0.61 & 0.74 \\
% GBM & 0.01 & 0.17 & 0.74 & 0.76 & 0.01 & 0.17 & 0.74 & 0.76 \\
XGBoost & 0.00 & 0.16 & 0.74 & 0.76 & 0.00 & 0.16 & 0.74 & 0.76 \\
\bottomrule
\end{tabular}}
\caption{Zero-shot LLM results on the \textbf{ACSMobility} benchmark task, together with supervised learning baselines fitted on 0.6M samples.}
\label{tab:acsmobility_results}
\end{table}

\begin{table}[p]
\centering
{\footnotesize \begin{tabular}{l|cccc|cccc}
\toprule
\multirow{3}{*}{\textbf{Model}} & \multicolumn{4}{c|}{\textbf{Multiple-choice prompting}} & \multicolumn{4}{c}{\textbf{Numeric risk prompting}} \\
 & \multirow{2}{*}{\textbf{ECE $\downarrow$}} & \multirow{2}{*}{\textbf{\shortstack{Brier \\score}} $\downarrow$} & \multirow{2}{*}{\textbf{\shortstack{AUC} $\uparrow$}} & \multirow{2}{*}{\textbf{\shortstack{Acc.} $\uparrow$}} & \multirow{2}{*}{\textbf{ECE $\downarrow$}} & \multirow{2}{*}{\textbf{\shortstack{Brier \\score}} $\downarrow$} & \multirow{2}{*}{\textbf{\shortstack{AUC} $\uparrow$}} & \multirow{2}{*}{\textbf{\shortstack{Acc.} $\uparrow$}} \\
 & & & & & & & & \\
\midrule
GPT 4o mini (it) & 0.39 & 0.40 & 0.65 & 0.55 & 0.15 & 0.27 & 0.58 & 0.57 \\
Mixtral 8x22B (it) & 0.31 & 0.33 & 0.66 & \cellcolor{cyan!17.1} 0.59 & 0.12 & \cellcolor{cyan!23.7} 0.24 & \cellcolor{cyan!23.5} 0.64 & \cellcolor{cyan!25.0} 0.59 \\
Mixtral 8x22B & 0.20 & 0.28 & 0.63 & \cellcolor{orange!25.0} 0.44 & 0.30 & 0.34 & 0.57 & \cellcolor{cyan!2.4} 0.58 \\
Llama 3 70B (it) & 0.15 & \cellcolor{cyan!22.6} 0.24 & \cellcolor{cyan!25.0} 0.70 & \cellcolor{cyan!25.0} 0.60 & 0.12 & \cellcolor{cyan!22.5} 0.24 & \cellcolor{cyan!25.0} 0.64 & 0.53 \\
Llama 3 70B & 0.09 & \cellcolor{cyan!21.4} 0.24 & 0.67 & 0.55 & 0.08 & \cellcolor{cyan!12.4} 0.25 & 0.52 & 0.46 \\
Mixtral 8x7B (it) & \cellcolor{orange!25.0} 0.45 & \cellcolor{orange!25.0} 0.45 & 0.66 & 0.52 & 0.09 & \cellcolor{cyan!25.0} 0.24 & 0.61 & 0.57 \\
Mixtral 8x7B & 0.28 & 0.32 & 0.60 & \cellcolor{orange!25.0} 0.44 & 0.07 & \cellcolor{cyan!16.2} 0.25 & 0.57 & \cellcolor{cyan!2.4} 0.58 \\
Yi 34B (it) & 0.35 & 0.36 & 0.65 & 0.56 & \cellcolor{cyan!3.7} 0.06 & \cellcolor{cyan!11.2} 0.25 & 0.50 & \cellcolor{orange!25.0} 0.44 \\
Yi 34B & \cellcolor{cyan!4.5} 0.08 & \cellcolor{cyan!25.0} 0.24 & 0.62 & 0.56 & 0.14 & 0.27 & 0.53 & \cellcolor{orange!25.0} 0.44 \\
Llama 3 8B (it) & 0.19 & 0.28 & 0.60 & 0.57 & 0.11 & \cellcolor{cyan!12.4} 0.25 & 0.56 & 0.56 \\
Llama 3 8B & 0.08 & \cellcolor{cyan!7.1} 0.25 & 0.53 & 0.56 & 0.12 & 0.26 & \cellcolor{orange!25.0} 0.48 & \cellcolor{orange!25.0} 0.44 \\
Mistral 7B (it) & \cellcolor{orange!2.7} 0.41 & 0.42 & 0.59 & 0.57 & 0.11 & \cellcolor{cyan!11.2} 0.25 & 0.55 & 0.56 \\
Mistral 7B & \cellcolor{cyan!22.5} 0.05 & \cellcolor{cyan!14.2} 0.25 & 0.57 & 0.56 & \cellcolor{orange!25.0} 0.44 & \cellcolor{orange!25.0} 0.44 & 0.50 & 0.56 \\
Gemma 7B (it) & \cellcolor{orange!12.0} 0.42 & 0.43 & 0.53 & 0.56 & 0.10 & \cellcolor{cyan!4.9} 0.26 & \cellcolor{orange!4.9} 0.49 & \cellcolor{orange!25.0} 0.44 \\
Gemma 7B & \cellcolor{cyan!25.0} 0.04 & \cellcolor{cyan!19.0} 0.24 & 0.61 & 0.58 & \cellcolor{cyan!25.0} 0.03 & \cellcolor{cyan!13.7} 0.25 & 0.52 & 0.55 \\
Gemma 2B (it) & 0.34 & 0.36 & \cellcolor{orange!9.3} 0.49 & 0.56 & 0.19 & 0.28 & 0.50 & 0.56 \\
Gemma 2B & 0.09 & \cellcolor{cyan!4.7} 0.26 & \cellcolor{orange!25.0} 0.48 & \cellcolor{orange!25.0} 0.44 & \cellcolor{orange!25.0} 0.44 & \cellcolor{orange!25.0} 0.44 & 0.50 & 0.56 \\
\midrule
LR & 0.04 & 0.24 & 0.58 & 0.56 & 0.04 & 0.24 & 0.58 & 0.56 \\
% GBM & 0.02 & 0.20 & 0.75 & 0.69 & 0.02 & 0.20 & 0.75 & 0.69 \\
XGBoost & 0.02 & 0.19 & 0.77 & 0.70 & 0.02 & 0.19 & 0.77 & 0.70 \\
\bottomrule
\end{tabular}}
\caption{Zero-shot LLM results on the \textbf{ACSTravelTime} benchmark task, together with supervised learning baselines fitted on 1.3M samples.}
\label{tab:acstraveltime_results}
\end{table}

\begin{table}[p]
\centering
{\footnotesize \begin{tabular}{l|cccc|cccc}
\toprule
\multirow{3}{*}{\textbf{Model}} & \multicolumn{4}{c|}{\textbf{Multiple-choice prompting}} & \multicolumn{4}{c}{\textbf{Numeric risk prompting}} \\
 & \multirow{2}{*}{\textbf{ECE $\downarrow$}} & \multirow{2}{*}{\textbf{\shortstack{Brier \\score}} $\downarrow$} & \multirow{2}{*}{\textbf{\shortstack{AUC} $\uparrow$}} & \multirow{2}{*}{\textbf{\shortstack{Acc.} $\uparrow$}} & \multirow{2}{*}{\textbf{ECE $\downarrow$}} & \multirow{2}{*}{\textbf{\shortstack{Brier \\score}} $\downarrow$} & \multirow{2}{*}{\textbf{\shortstack{AUC} $\uparrow$}} & \multirow{2}{*}{\textbf{\shortstack{Acc.} $\uparrow$}} \\
 & & & & & & & & \\
\midrule
GPT 4o mini (it) & 0.33 & 0.34 & \cellcolor{cyan!25.0} 0.71 & 0.60 & 0.10 & 0.20 & 0.68 & \cellcolor{cyan!17.6} 0.73 \\
Mixtral 8x22B (it) & 0.24 & 0.25 & \cellcolor{cyan!11.7} 0.70 & \cellcolor{cyan!12.7} 0.72 & \cellcolor{cyan!25.0} 0.04 & \cellcolor{cyan!25.0} 0.18 & \cellcolor{cyan!11.0} 0.71 & \cellcolor{cyan!25.0} 0.75 \\
Mixtral 8x22B & 0.32 & 0.30 & 0.59 & \cellcolor{orange!25.0} 0.30 & \cellcolor{orange!15.5} 0.29 & \cellcolor{orange!3.4} 0.29 & 0.54 & \cellcolor{cyan!0.5} 0.70 \\
Llama 3 70B (it) & 0.16 & \cellcolor{cyan!16.2} 0.21 & \cellcolor{cyan!7.0} 0.69 & \cellcolor{cyan!25.0} 0.75 & 0.13 & 0.20 & \cellcolor{cyan!25.0} 0.73 & \cellcolor{cyan!24.4} 0.75 \\
Llama 3 70B & 0.18 & \cellcolor{cyan!8.3} 0.22 & 0.67 & 0.63 & 0.12 & 0.21 & 0.64 & 0.53 \\
Mixtral 8x7B (it) & 0.20 & \cellcolor{cyan!7.4} 0.23 & \cellcolor{cyan!12.7} 0.70 & \cellcolor{cyan!23.3} 0.74 & \cellcolor{cyan!0.3} 0.06 & \cellcolor{cyan!18.5} 0.19 & 0.69 & \cellcolor{cyan!22.2} 0.74 \\
Mixtral 8x7B & 0.41 & 0.37 & 0.57 & \cellcolor{orange!25.0} 0.30 & 0.20 & 0.25 & 0.56 & 0.70 \\
Yi 34B (it) & \cellcolor{cyan!16.7} 0.06 & \cellcolor{cyan!25.0} 0.19 & 0.67 & \cellcolor{cyan!19.4} 0.74 & 0.22 & 0.24 & 0.57 & \cellcolor{orange!25.0} 0.31 \\
Yi 34B & \cellcolor{cyan!25.0} 0.04 & \cellcolor{cyan!17.2} 0.21 & 0.59 & \cellcolor{cyan!2.1} 0.70 & 0.09 & 0.20 & 0.67 & 0.64 \\
Llama 3 8B (it) & 0.11 & \cellcolor{cyan!12.7} 0.21 & 0.59 & \cellcolor{cyan!7.7} 0.71 & 0.17 & 0.22 & 0.64 & 0.68 \\
Llama 3 8B & 0.41 & 0.38 & 0.55 & \cellcolor{orange!25.0} 0.30 & 0.20 & 0.25 & 0.51 & \cellcolor{orange!6.2} 0.34 \\
Mistral 7B (it) & 0.30 & 0.30 & 0.61 & \cellcolor{cyan!0.4} 0.70 & 0.07 & 0.20 & 0.67 & 0.65 \\
Mistral 7B & 0.29 & 0.30 & \cellcolor{orange!25.0} 0.45 & \cellcolor{orange!25.0} 0.30 & \cellcolor{orange!22.1} 0.30 & \cellcolor{orange!18.5} 0.30 & 0.50 & 0.70 \\
Gemma 7B (it) & 0.30 & 0.34 & \cellcolor{orange!14.6} 0.46 & 0.50 & 0.18 & 0.24 & 0.57 & 0.61 \\
Gemma 7B & 0.15 & \cellcolor{cyan!3.4} 0.23 & 0.49 & 0.49 & 0.18 & 0.26 & 0.48 & 0.70 \\
Gemma 2B (it) & \cellcolor{orange!25.0} 0.70 & \cellcolor{orange!25.0} 0.70 & 0.54 & \cellcolor{orange!25.0} 0.30 & 0.24 & \cellcolor{orange!14.2} 0.29 & \cellcolor{orange!25.0} 0.42 & 0.42 \\
Gemma 2B & 0.26 & 0.28 & 0.54 & \cellcolor{orange!25.0} 0.30 & \cellcolor{orange!25.0} 0.30 & \cellcolor{orange!25.0} 0.30 & 0.50 & 0.70 \\
\midrule
LR & 0.03 & 0.19 & 0.70 & 0.72 & 0.03 & 0.19 & 0.70 & 0.72 \\
% GBM & 0.01 & 0.14 & 0.83 & 0.80 & 0.01 & 0.14 & 0.83 & 0.80 \\
XGBoost & 0.00 & 0.14 & 0.84 & 0.80 & 0.00 & 0.14 & 0.84 & 0.80 \\
\bottomrule
\end{tabular}}
\caption{Zero-shot LLM results on the \textbf{ACSPublicCoverage} benchmark task, together with supervised learning baselines fitted on 1.0M samples.}
\label{tab:acspubliccoverage_results}
\end{table}

% \begin{table}[p]
%     \centering
%     \input{tables/multiple_choice_prompting/acsemployment}
%     \caption{Zero-shot LLM results on the \textbf{ACSEmployment} benchmark task, together with supervised learning baselines fitted on 2.9M samples.
%     See caption of Table~\ref{tab:acsincome_results} for more details.
%     }
%     \label{tab:acsemployment_results}
% \end{table}
% \begin{table}[p]
%     \centering
%     \input{tables/multiple_choice_prompting/acsmobility}
%     \caption{Zero-shot LLM results on the \textbf{ACSMobility} benchmark task, together with supervised learning baselines fitted on 0.6M samples.
%     See caption of Table~\ref{tab:acsincome_results} for more details.
%     }
%     \label{tab:acsmobility_results}
% \end{table}
% \begin{table}[p]
%     \centering
%     \input{tables/multiple_choice_prompting/acstraveltime}
%     \caption{Zero-shot LLM results on the \textbf{ACSTravelTime} benchmark task, together with supervised learning baselines fitted on 1.3M samples.
%     See caption of Table~\ref{tab:acsincome_results} for more details.
%     }
%     \label{tab:acstraveltime_results}
% \end{table}
% \begin{table}[p]
%     \centering
%     \input{tables/multiple_choice_prompting/acspubliccoverage}
%     \caption{Zero-shot LLM results on the \textbf{ACSPublicCoverage} benchmark task, together with supervised learning baselines fitted on 1.0M samples.
%     See caption of Table~\ref{tab:acsincome_results} for more details.
%     }
%     \label{tab:acspubliccoverage_results}
% \end{table}

% \section{Feature importance for LLMs}

\begin{figure}[tbp]
    \centering
    \begin{subfigure}[b]{0.38\textwidth}
        \centering
        \includegraphics[height=15em, trim=0 0 14.5em 0, clip]{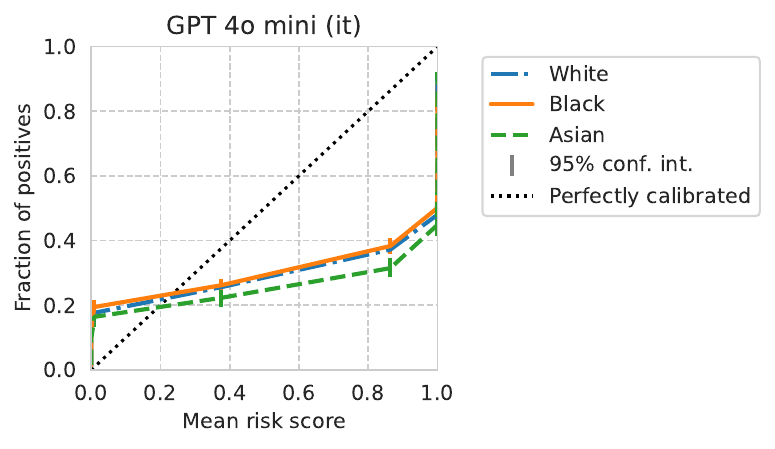}
        \caption{Multiple-choice prompting.}
        \label{fig:subgroup-calibration-gpt4o-multiple-choice}
    \end{subfigure}
    \hfill
    \begin{subfigure}[b]{0.61\textwidth}
        \centering
        \includegraphics[height=15em]{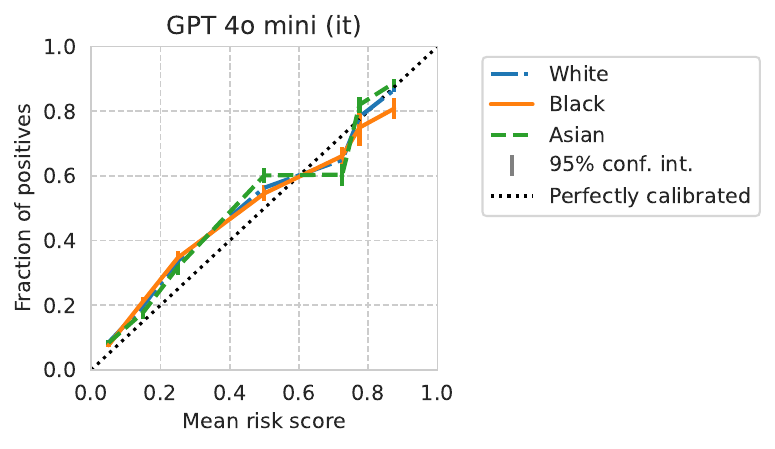}
        \caption{Numeric prompting.}
        \label{fig:subgroup-calibration-gpt4o-numeric}
    \end{subfigure}
    \caption{Calibration curves for the `GPT 4o mini' model, across different race sub-populations on the ACSIncome task, computed using $10$ quantile-based bins.
    This model is only available through a web API, and no base model variant is available. Numeric prompting (Fig.~\ref{fig:subgroup-calibration-gpt4o-numeric}) leads to reduced differences in group-conditional calibration (improved group fairness), as well as large improvements in overall calibration.
    }
    \label{fig:subgroup-calibration-gpt4o}
\end{figure}

% \clearpage

\begin{figure}[tbp]
    \centering
    \includegraphics[height=11em]{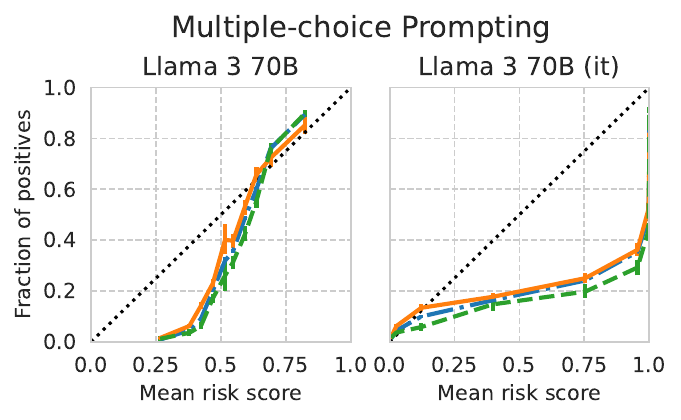}
    \includegraphics[height=11em]{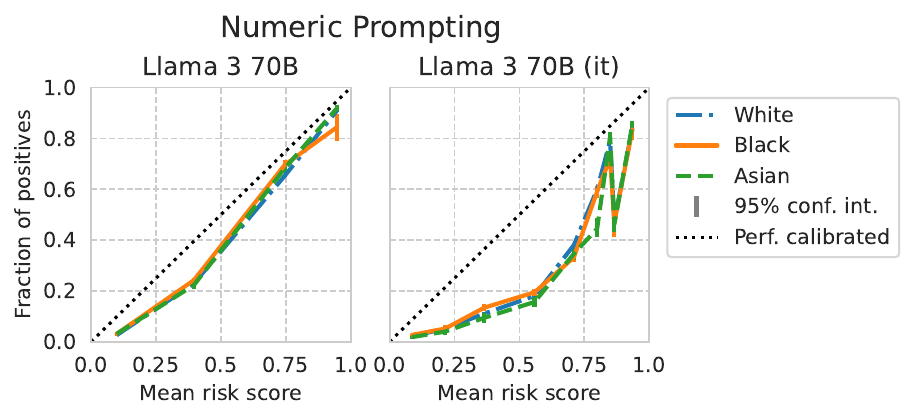}
    \includegraphics[height=11em]{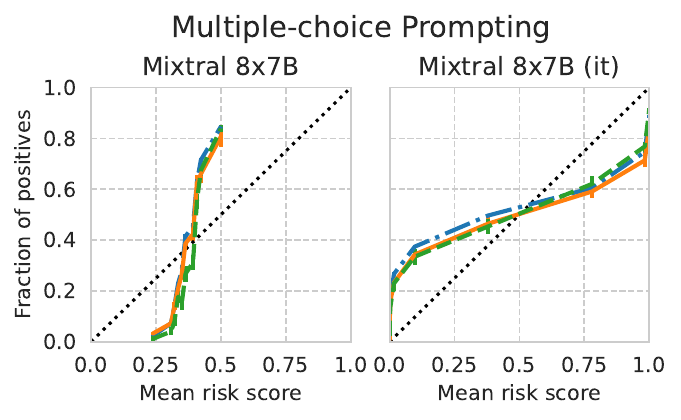}
    \includegraphics[height=11em]{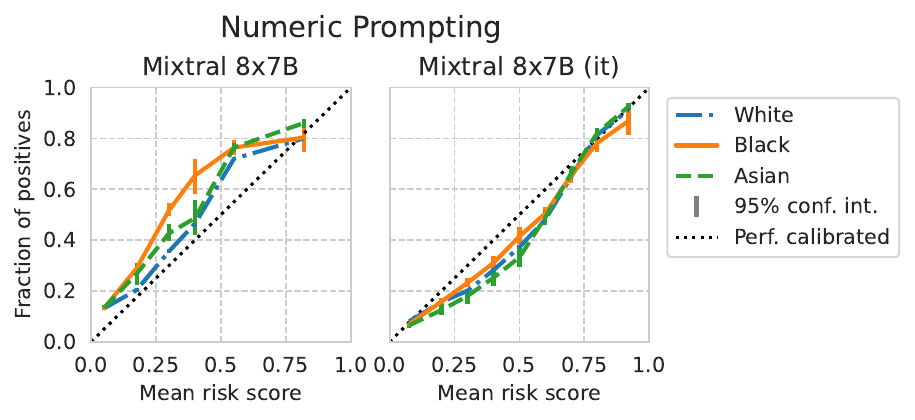}
    \includegraphics[height=11em]{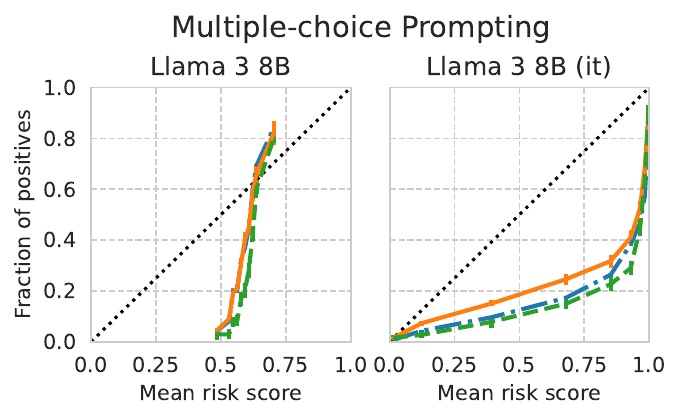}
    \includegraphics[height=11em]{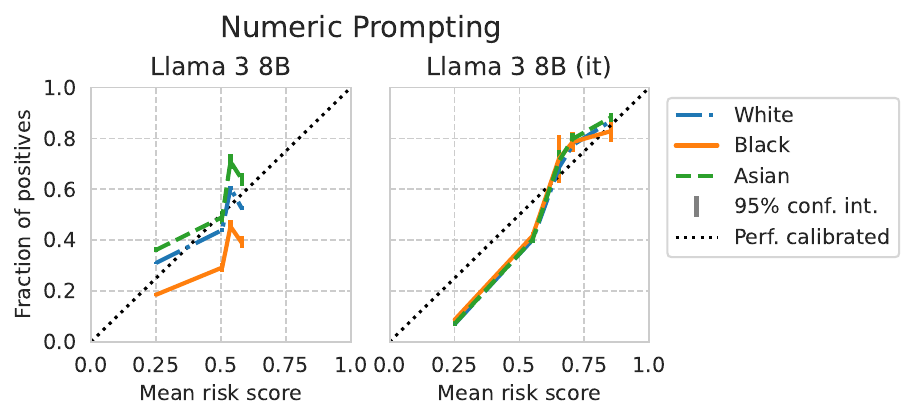}
    \includegraphics[height=11em]{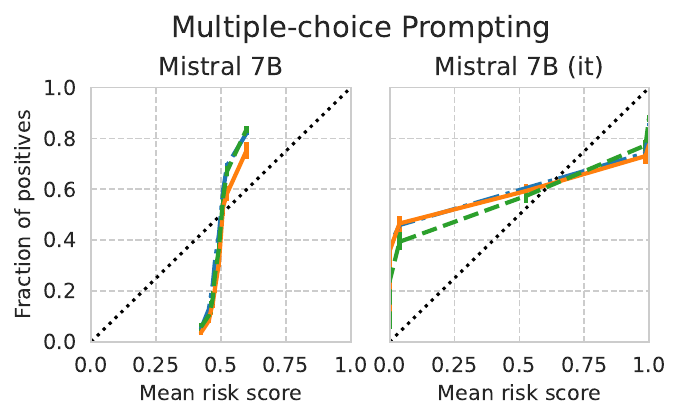}
    \includegraphics[height=11em]{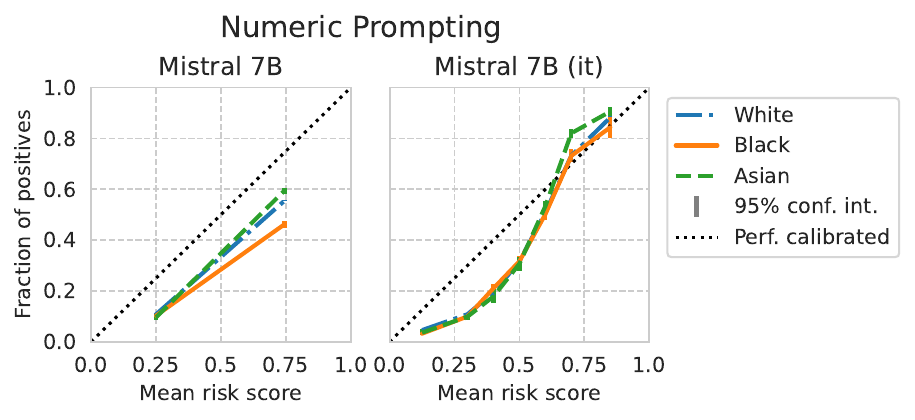}
    \caption{Calibration curves across different race sub-populations on the ACSIncome task, computed using $10$ quantile-based score bins, with $95\%$ confidence intervals.
    Using multiple-choice prompting to generate risk scores (\textit{left plots}) leads to sizeable calibration error for both base and instruction-tuned models.
    The models shown in this figure generally do not follow the same group bias trend seen in the Mixtral 8x22B and Yi 34B models (Figure~\ref{fig:subgroup-calibration-yi-mixtral}).
    }
    \label{fig:subgroup-calibration-other-models}
\end{figure}

\subsection{Additional subgroup calibration results}
\label{app:additional_subgroup_cali_plots}

This section contains additional subgroup calibration plots on the ACSIncome task.
Figure~\ref{fig:subgroup-calibration-yi-mixtral} shows the 2 worst offenders: Mixtral 8x22B and Yi 34B.
Figures~\ref{fig:subgroup-calibration-gpt4o} and~\ref{fig:subgroup-calibration-other-models} show the remaining models.
The same trend is visible but to a lesser extent: Instruction-tuned models show a bias towards lower scores for Black individuals, even after controlling for label prevalence.

We can quantify the positive score bias by evaluating the \textit{signed calibration error} (SCE):
\begin{equation}
\label{eq:signed_calibration_error}
    \mathrm{SCE} \coloneq \frac{1}{n} \sum_{m=1}^M \sum_{i\in B_m} (r_i - y_i),
\end{equation}
where $M$ is the number of score buckets, $B_m$ is the set of sample indices belonging to bucket $m$, $r_i = f_\theta(x_i)$ is the risk score given to sample $x_i$ and $y_i$ is its label.
This metric does not evaluate overall calibration, as a value of 0 does not indicate a calibrated classifier. Instead, negative/positive values indicate a bias towards lower/higher risk scores, respectively.
If higher scores are related to positive real-world outcomes (e.g., when predicting income for a loan application), then a bias towards lower scores on samples of specific protected subgroups would likely lead to unfair real-world outcomes.
Conversely, if a negative class prediction is associated with a positive outcome (e.g., when predicting risk of recidivism), then a bias towards lower scores would be beneficial for the affected group.~\looseness-1

\begin{figure}[tbp]
    \centering
    \includegraphics[width=0.47\linewidth]{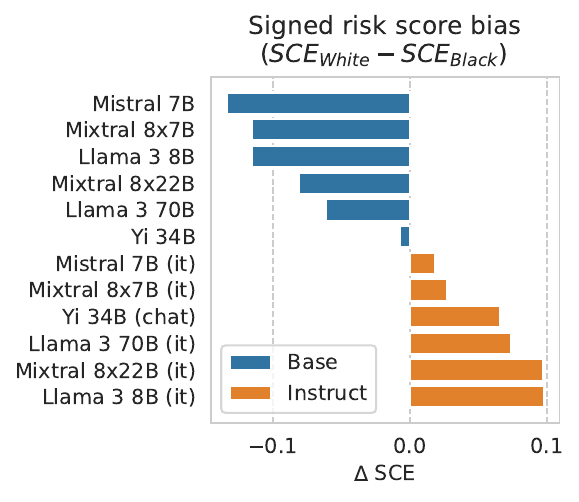}
    \hfill
    \includegraphics[width=0.47\linewidth]{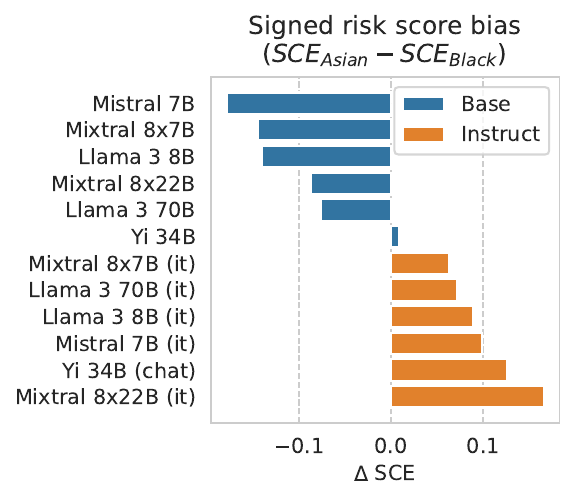}
    \caption{Racial group bias in risk score calibration error.
    When comparing score bias of group $A$ with score bias of group $B$, $\Delta_\mathrm{SCE} = \mathrm{SCE}_A - \mathrm{SCE}_B$, positive values indicate an undue score advantage of group $A$, and negative values an undue score advantage of group $B$.
    \textit{Left:} Difference between White and Black groups, $\mathrm{SCE}_{White} - \mathrm{SCE}_{Black}$.
    \textit{Right:} Difference between Asian and Black groups, $\mathrm{SCE}_{Asian} - \mathrm{SCE}_{Black}$.
    Note that the Gemma models were omitted, as their instruct versions degenerate into strictly predicting the same outcome for all samples.
    Consequently, the two instruct Gemma models are the only exceptions to the trend shown in these plots.
    }
    \label{fig:score_bias_subgroups}
\end{figure}

Figure~\ref{fig:score_bias_subgroups} shows the difference between the signed calibration error (SCE) on different group pairings, on the ACSIncome task. Positive predictions ($Y=1$) correspond to the advantaged high-income class.
Negative differences, $\Delta_{SCE} < 0$, indicates advantaged scores for Black individuals, while positive differences, $\Delta_{SCE} > 0$, indicate disadvantaged scores for Black individuals.
Interestingly, while subgroup calibration curves appear similar for base models and disadvantage Black individuals for instruction-tuned models 
(see Figures~\ref{fig:subgroup-calibration-yi-mixtral},~\ref{fig:subgroup-calibration-gpt4o}, and~\ref{fig:subgroup-calibration-other-models}), this is not entirely reflected on the SCE metric.
Indeed, a clear-cut split is visible: base models benefit the score of Black individuals, and instruct models disadvantage the score of Black individuals.
This finding could be partially explained by the fact that base models produce score distributions with low variance, and instruct models produce high-variance polarized outcomes.
Specifically, two conclusions can be drawn from the score distributions produced by base models: 
(1) models under-estimate the score of high-income earners (which are disproportionately Asian), and (2) models over-estimate the score of low-income earners (which are disproportionately Black).
This could lead base models to over-estimate the earnings of the Black population disproportionately to other groups.
The opposite is true for instruction-tuned models: high-income earners see their score over-estimated, which benefits groups with a higher prevalence of high earners.

This in no way serves as an exhaustive analysis of risk score fairness on LLMs, as it is bound to be highly task dependent and language model dependent.
We simply surface the fact that, on this high-income prediction task, risk scores are not group-calibrated~\citep{barocas-hardt-narayanan} and could lead to unfair outcomes.
Crucially, even though race has the lowest mean feature importance among all features (see Appendix~\ref{app:feature_importance}), we report and explain how different trends in risk score distributions can effectively lead to unfair outcomes.

% \begin{figure}[tbp]
%     \centering
%     \includegraphics[width=0.82\textwidth]{imgs/multiple_choice_prompting/calibration-curves-per-subgroup.gemma-2b.pdf}
%     \includegraphics[width=0.82\textwidth]{imgs/multiple_choice_prompting/calibration-curves-per-subgroup.gemma-7b.pdf}
%     \caption{Calibration curves for the Gemma family of models, across different race sub-populations on the ACSIncome task. \textit{Left}: Base models. \textit{Right}: Instruction-tuned models. Note that these are the smallest models and generally perform the worst on all tasks. The plotted calibration curves are in agreement with the lackluster performance, as they show extreme score distributions that are unlike those of other LLMs.
%     In fact, the instruction-tuned Gemma models consistently output only positive predictions, hindering their analysis.
%     }
%     \label{fig:subgroup-calibration-app2}
% \end{figure}

\subsection{Varying uncertainty}
\label{app:varying_uncertainty}

\begin{figure}[thbp]
    \centering
    \includegraphics[width=0.85\linewidth]{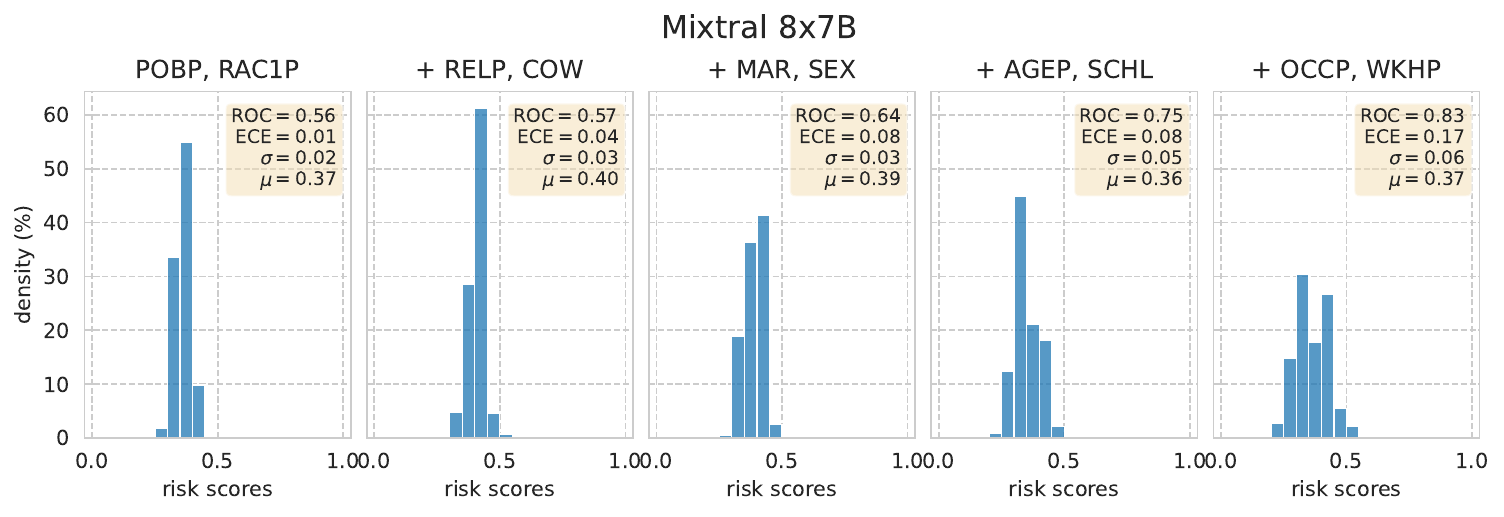}
    \includegraphics[width=0.85\linewidth]{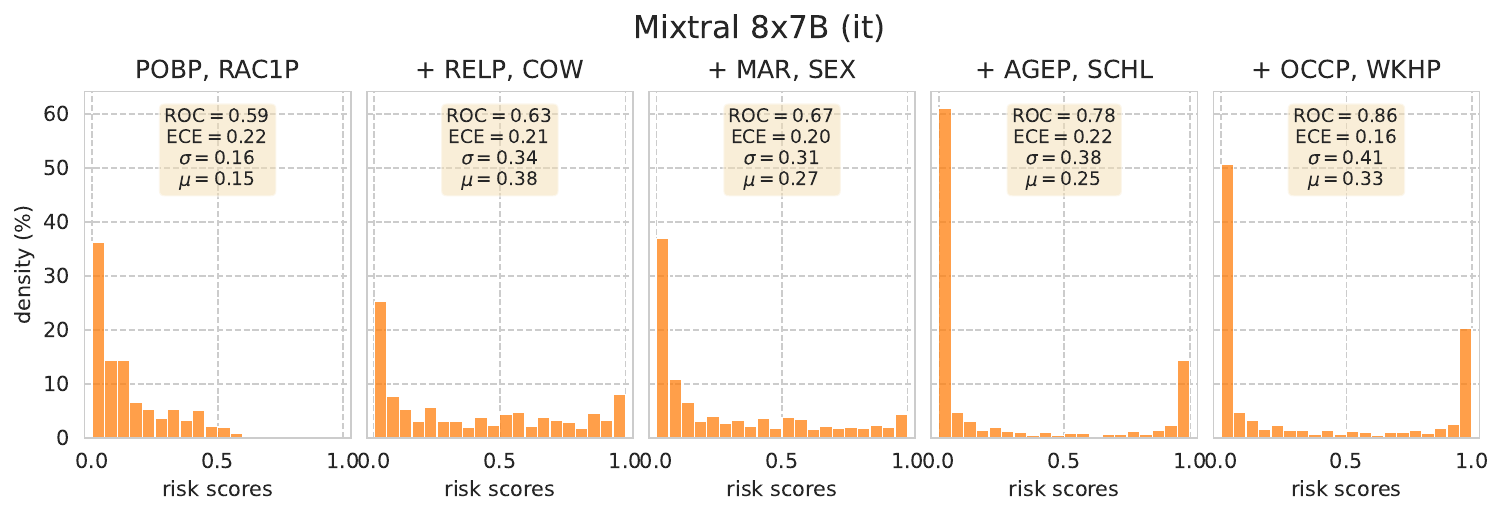}
    \caption{Shift of score distribution with increasing evidence for the Mixtral 8x7B model (which achieved the best Brier score), using multiple-choice Q\&A on the ACSIncome task.
    Features are described in Table~\ref{tab:col_to_text}.
    Score distribution gets more discriminative as more evidence is added, successfully increasing scores' predictive signal (AUC).
    The true label prevalence is $\Pr[Y=1] = 0.37$.
    }
    \label{fig:increasing_evidence}
\end{figure}

\begin{figure}[tbp]
    \centering
    \includegraphics[width=0.9\textwidth]{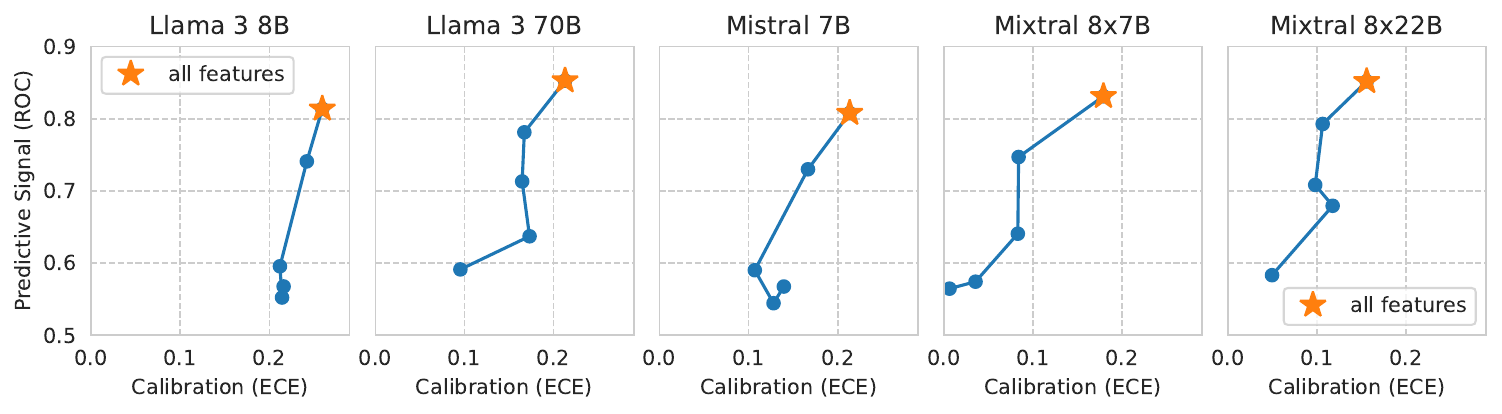}
    \includegraphics[width=0.9\textwidth]{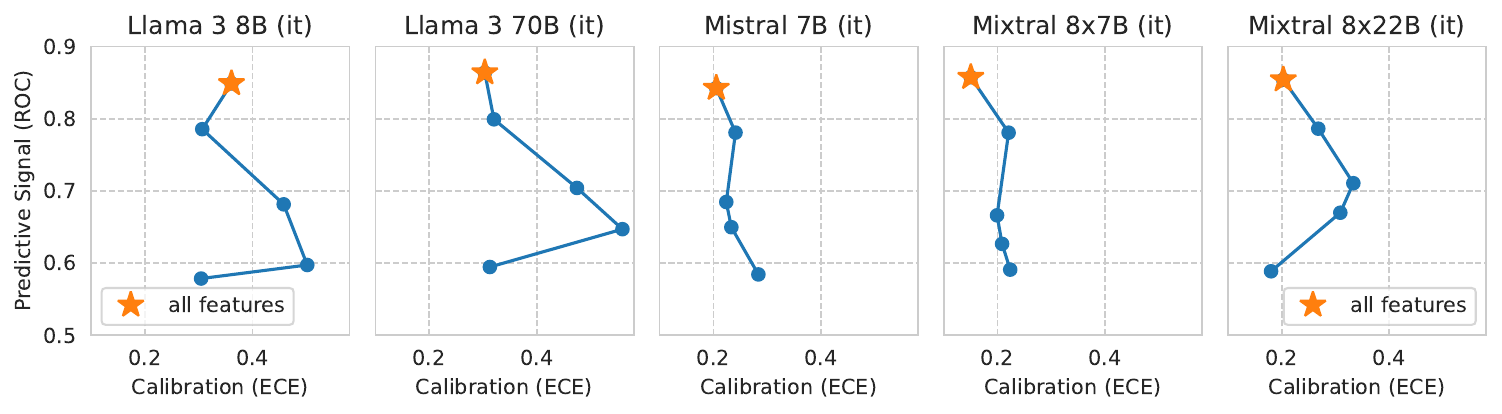}
    \caption{Evaluation of calibration (ECE) and predictive performance (AUC) on Llama and Mistral models, with an increasing number of features provided on the ACSIncome task. For each dot along the line we add two features, up to all 10 features being used in the point marked with a star.
    \textit{Top row}: base models. \textit{Bottom row}: instruction-tuned models.
    Models can successfully use each extra feature to increase predictive signal.
    Calibration trends worse the more features are added for base models, while instruct models show no clear trend.~\looseness-1
    }
    \label{fig:features_subsets_appendix}
\end{figure}

A simple API call in our package allows for selecting different subsets of attributes to include as features when using the LLM as a predictor.
Figure~\ref{fig:features_subsets_appendix} inspects the effect of increasing the feature on models' calibration and predictive power.
% We start from all 10 ACSIncome features (orange star marker) and iteratively remove one at a time.
%
Each dot along the line represents an increasing feature set used for LLM predictions, added in order of mean feature importance on all models.
%
% Features are added in the following order: AGEP, COW, SCHL, MAR, OCCP, POBP, RELP, WKHP, SEX, RAC1P.
%
Appendix~\ref{app:acsincome} details all features used in the ACSIncome task. We refer the reader to \citet{ding2021retiring} and the ACS codebook\footnote{\label{foot:codebook}See the ACS PUMS data dictionary for the full list of available variables:\\ \url{https://www.census.gov/programs-surveys/acs/microdata/documentation.html}}
% \textsuperscript{\ref{foot:codebook}} 
for an in-depth description of each categorical value a feature can take.
Predictive signal (ROC AUC) reliably increases with each added feature, for all tested models except the Gemma 2B variants.
This is expected with standard supervised learning algorithms trained and evaluated on i.i.d. data, but arguably somewhat unexpected of pre-trained models trained on a variety of datasets that are out-of-distribution relative to the evaluation set.
On the other hand, there is no clear trend for calibration: for Mistral models, it seems that calibration actually worsens for larger feature sets, while for Llama models calibration is approximately stable across all points.
% While there is no clear trend, we see that variations in calibration across feature subsets tend to be small.

This experiment show-cases one unique way of using LLMs with survey prediction tasks: while supervised learning models would have to be retrained every time a different feature set is used, LLMs can freely change the evidence they use to make a prediction.
If a model were to exhibit properties of a joint distribution with the ability to marginalize over hidden features, then calibration with respect to evidence $\cX$ implies that it is also calibrated with respect to restricted evidence $\cX'\subset \cX$. To what extent a model satisfies such properties is an interesting question for future work; we hope our package proves useful as an investigative tool.

\subsection{Feature importance}
\label{app:feature_importance}

In this section, we present feature importance results for different LLMs on the ACSIncome prediction task.
The importance value of feature $j$ is computed as the drop in AUC after permuting all values of feature $j$ across the dataset.
That is, each sample $x$, sees its value for feature $j$ randomly permuted with another sample.
This is a common feature importance implementation~\citep{breiman2001random}, as it does not rely on any internal characteristics of the model.

\begin{figure}[tbp]
    \centering
    \includegraphics[width=\textwidth]{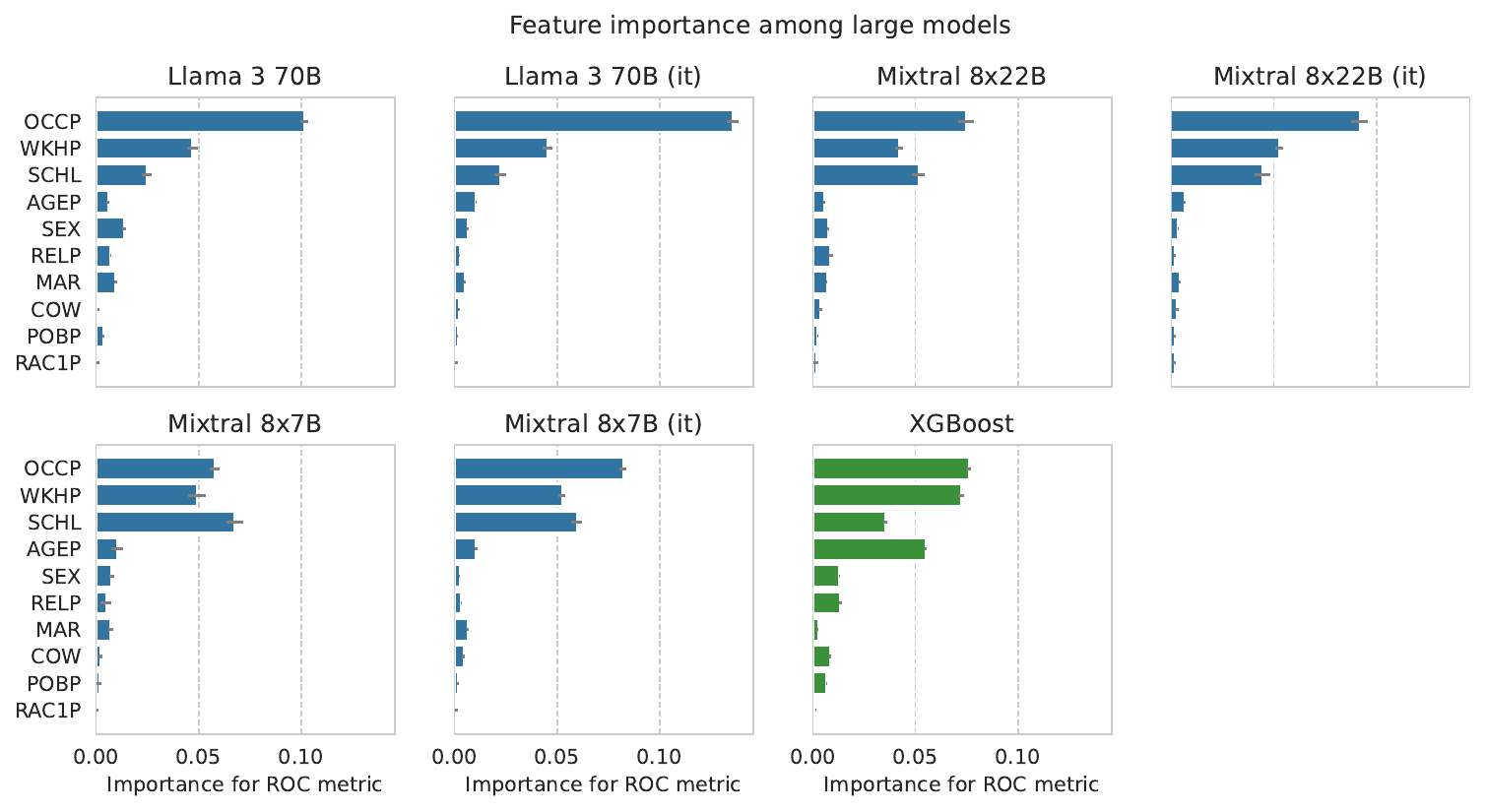}
    \caption{Feature importance among the largest language models tested, plus results for the XGBoost baseline. Feature importance values are calculated as the loss in AUC when the values of a given column are randomly permuted~\citep{breiman2001random}.}
    \label{fig:feat_imp_large_models}
\end{figure}

Figure~\ref{fig:feat_imp_large_models} shows feature importance values for the largest language models studied (above 40B active parameters). Results for the XGBoost model are also shown in green.
Note that XGBoost achieves the best result on every single metric in Table~\ref{tab:acsincome_results}.
While for supervised models, a given categorical value is nothing more than a 1 or 0, LLMs have the potential to surface the real-world meaning of such values, benefiting from the rich embedding representations of each category.
As such, we'd expect to see LLMs assigning higher importance to categorical features.
Indeed, Llama 3 models assign considerably higher importance to the occupation feature (OCCP), which is a numerically encoded categorical feature with over 500 different possible values.
Conversely, the XGBoost model assigns considerably higher importance than LLMs to `work-hours per week' (WKHP) and `age' (AGEP), both integer-encoded features.
Lastly, feature importance results indicate that the studied LLMs do not explicitly use sensitive categories such as age (AGEP), sex (SEX), or race (RAC1P) for risk score estimation.

Interestingly, feature importance is similar for base and instruct variants of the same model.
This contrasts with the score distribution and calibration curve results, where all base models followed a similar trend, distinct from their instruction-tuned versions.

% \clearpage
\section{Details on provided benchmark tasks}
\label{app:prediction_tasks}

The \folktexts package defines natural-text mappings for a variety of columns in the ACS PUMS data files. %, enabling the use of any combination of columns as features and target for an LLM prediction task.
Table~\ref{tab:col_to_text} lists and describes each implemented column-to-text mapping.
Any combination of column-to-text objects can be used to create a prediction task from ACS data, both as features and as the prediction target.
To enable straightforward comparison with existing benchmarks, we mimic the feature set and population filters used by the prediction tasks available in the popular {\tt folktables} benchmark package~\citep{ding2021retiring}.
Specifically, we put forth natural-text variants of the ACSIncome, ACSPublicCoverage, ACSMobility, ACSEmployment, and ACSTravelTime tasks.
These prediction tasks define a restricted set of columns from the ACS PUMS data files to be used as input features for machine learning models, as well as a binarized target column.
As such, we extend the use of these ACS prediction tasks to benchmark language models, enabling direct comparison with a wide-ranging set of literature works.
Although any ACS survey year could be used for benchmarking, we define the standard set of benchmark tasks as those using data from the 2018 1-year-horizon person-level survey (following~\citet{ding2021retiring}).
Notwithstanding, we welcome the addition of new column-to-text mappings by new users of the package, both for ACS data and for new datasets.
The following paragraphs detail each pre-implemented prediction task.

\paragraph{ACSIncome}
\label{app:acsincome}
The goal of the ACSIncome task is to predict whether a person's yearly income is above \$50,000, given by the PINCP column.
The ACS columns used as features are: AGEP, COW, SCHL, MAR, OCCP, POBP, RELP, WKHP, SEX, and RAC1P.
The sub-population over which the task is conducted is employed US residents with age greater than 16 years.
The ACSIncome prediction task was put-forth as the successor to the popular UCI Adult dataset~\citep{misc_adult_2}, used extensively in the algorithmic fairness literature.
This task is the default task when running the \folktexts benchmark.

\paragraph{ACSPublicCoverage}
The goal of the ACSPublicCoverage task is to predict whether an individual is covered by public health insurance, given by the PUBCOV column.
The ACS columns used as features are: AGEP, SCHL, MAR, SEX, DIS, ESP, CIT, MIG, MIL, ANC, NATIVITY, DEAR, DEYE, DREM, PINCP, ESR, ST, FER, and RAC1P.
The sub-population over which the task is conducted is US residents with age below 65 years old, and with personal income below \$30,000.

\paragraph{ACSMobility}
The goal of the ACSMobility task is to predict whether an individual has changed their home address in the last year, given by the MIG column.
The ACS columns used as features are: AGEP, SCHL, MAR, SEX, DIS, ESP, CIT, MIL, ANC, NATIVITY, RELP, DEAR, DEYE, DREM, RAC1P, COW, ESR, WKHP, JWMNP, and PINCP. %%% TODO: GCL column
The sub-population over which the task is conducted is US residents with age between 18 and 35.

\paragraph{ACSEmployment}
The goal of the ACSEmployment is to predict whether an individual is employed, given by the ESR column.
The ACS columns used as features are: AGEP, SCHL, MAR, SEX, DIS, ESP, MIG, CIT, MIL, ANC, NATIVITY, RELP, DEAR, DEYE, DREM, and RAC1P. %%% GCL
The sub-population over which the task is conducted is US residents with age between 16 and 90.

\paragraph{ACSTravelTime}
The goal of the ACSTravelTime task is to predict whether a person's commute time to work is greater than 20 minutes, given by the JWMNP column.
The ACS columns used as features are: AGEP, SCHL, MAR, SEX, DIS, ESP, MIG, RELP, RAC1P, ST, CIT, OCCP, JWTR, and POVPIP.
%%% NOTE: missing PUMA, POWPUMA, columns
The sub-population over which the task is conducted is employed US residents with age greater than 16 years.

\section{\folktexts package usage}
\label{app:folktexts}

The \folktexts package is made available to the public via its open-source code repository\textsuperscript{\ref{foot:implementation_url}} and as a standalone package to be installed via the Python Package Index (PyPI).
It is compatible with PyTorch models used locally, as well as with web-hosted models available through an API.
The main user-facing classes are \lstinline{Benchmark}, \lstinline{BenchmarkConfig}, \lstinline{LLMClassifier}, \lstinline{TaskMetadata}, \lstinline{ColumnToText}, and \lstinline{Dataset}.
The responsibilities of each class are ascribed as follows
\begin{itemize}
    \item The \lstinline{Benchmark} class is responsible for running a benchmark task, which consists in obtaining risk scores from a given LLM on a given dataset, and evaluating those predictions on a variety of benchmark metrics.
    \item The \lstinline{BenchmarkConfig} class details all configurations of a benchmark (see Figure~\ref{fig:cli_usage} for available options).
    \item The \lstinline{LLMClassifier} class is comprised of a transformers model, a tokenizer, and a task; and is responsible for producing risk scores given some tabular rows for the provided task.
    \item The \lstinline{TaskMetadata} class is responsible for defining a set of feature columns and target column, together with holding the corresponding column-to-text objects to map an entire tabular row to its natural-text representation. The benchmark ACS tasks instantiate a subclass named \lstinline{ACSTaskMetadata}.
    \item The \lstinline{ColumnToText} class is responsible for producing meaningful natural-text representations of each possible value of a numeric or categorical column.
    \item The \lstinline{Dataset} class is responsible for holding tabular data and enabling reproducible manipulation of that data, such as splitting in train/test/validation, or filtering for a specified sub-population. The data used for the benchmark ACS tasks is provided by a subclass named \lstinline{ACSDataset}.
\end{itemize}

Additionally, a command-line interface is provided to ease usability: The benchmark ACS tasks can be ran using the \lstinline{run_acs_benchmark} executable. Figure~\ref{fig:cli_usage} details each available flag. Further infromation and example notebooks can be found on github at: \url{https://github.com/socialfoundations/folktexts}.

\begin{figure}[tbh]
\centering
{\scriptsize{
\begin{lstlisting}[numbers=none,basicstyle=\footnotesize]
usage:

run_acs_benchmark [-h] --model MODEL --results-dir RESULTS_DIR --data-dir DATA_DIR [--task TASK] [--few-shot FEW_SHOT] [--batch-size BATCH_SIZE] [--context-size CONTEXT_SIZE] [--fit-threshold FIT_THRESHOLD] [--subsampling SUBSAMPLING] [--seed SEED] [--use-web-api-model] [--dont-correct-order-bias] [--numeric-risk-prompting] [--reuse-few-shot-examples] [--use-feature-subset USE_FEATURE_SUBSET]
                         [--use-population-filter USE_POPULATION_FILTER] [--logger-level {DEBUG,INFO,WARNING,ERROR,CRITICAL}]

Benchmark risk scores produced by a language model on ACS data.

options:
  -h, --help            show this help message and exit
  --model MODEL         [str] Model name or path to model saved on disk
  --results-dir RESULTS_DIR
                        [str] Directory under which this experiment's results will be saved
  --data-dir DATA_DIR   [str] Root folder to find datasets on
  --task TASK           [str] Name of the ACS task to run the experiment on
  --few-shot FEW_SHOT   [int] Use few-shot prompting with the given number of shots
  --batch-size BATCH_SIZE
                        [int] The batch size to use for inference
  --context-size CONTEXT_SIZE
                        [int] The maximum context size when prompting the LLM
  --fit-threshold FIT_THRESHOLD
                        [int] Whether to fit the prediction threshold, and on how many samples
  --subsampling SUBSAMPLING
                        [float] Which fraction of the dataset to use (if omitted will use all data)
  --seed SEED           [int] Random seed -- to set for reproducibility
  --use-web-api-model   [bool] Whether use a model hosted on a web API (instead of a local model)
  --dont-correct-order-bias
                        [bool] Whether to avoid correcting ordering bias, by default will correct it
  --numeric-risk-prompting
                        [bool] Whether to prompt for numeric risk-estimates instead of multiple-choice Q&A
  --reuse-few-shot-examples
                        [bool] Whether to reuse the same samples for few-shot prompting (or sample new ones every time)
  --use-feature-subset USE_FEATURE_SUBSET
                        [str] Optional subset of features to use for prediction, comma separated
  --use-population-filter USE_POPULATION_FILTER
                        [str] Optional population filter for this benchmark; must follow the format 'column_name=value' to filter the dataset by a specific value.
  --logger-level {DEBUG,INFO,WARNING,ERROR,CRITICAL}
                        [str] The logging level to use for the experiment
\end{lstlisting}}}
\caption{Documentation for using {\tt folktexts} package through the command-line interface. An executable named \lstinline{run\_acs\_benchmark} is made available to run the standard ACS benchmark tasks with a variety of available customization options. Detailed documentation available at \href{https://socialfoundations.github.io/folktexts/}{socialfoundations.github.io/folktexts/}
}
\label{fig:cli_usage}
\end{figure}

% \begin{landscape}
% \begin{figure}[p]
%     \centering
%     \includegraphics[width=0.9\linewidth]{diagrams/folktexts-UML.drawio.pdf}
%     \includegraphics[width=0.9\linewidth, right]{diagrams/folktexts-flow.drawio-2.pdf}
%     \caption{\textit{Top}: Diagram of public-facing API classes and modules. The main class of interest to the user is highlighted in green: {\tt LLMClassifier}. The package comes with a variety of pre-instantiated {\tt Task}, {\tt Dataset}, and {\tt ColumnToText} objects related to the American Community Survey data. \\
%     \textit{Bottom}: Illustrative diagram of information flow through \folktexts classes, and how if facilitates going from tabular data to risk scores using LLMs.
%     }
%     \label{fig:uml_diagrams}
% \end{figure}
% \end{landscape}

\begin{landscape}
\begin{table}
\centering
\begin{tabular}{lll}
\textit{\textbf{Column}} & \textit{\textbf{Description}} & \textit{\textbf{Example}} \\
%%% ACS Income
AGEP & Age & The individual's age is: {\color{greyish}\tt 42 years old}. \\
COW & Class of worker & The individual's current employment status is: {\color{greyish}\tt Working for a non-profit organization}. \\
SCHL & Educational attainment & The individual's highest grade completed is: {\color{greyish}\tt 12th grade}. \\
MAR & Marital status & The individual's marital status is: {\color{greyish}\tt Married}. \\
OCCP & Occupation & The individual's occupation is: {\color{greyish}\tt Human Resources Manager}. \\
POBP & Place of birth & The individual's place of birth is: {\color{greyish}\tt New Zealand}. \\
RELP & Relationship & The individual's relationship to the reference survey respondent in the household is: {\color{greyish}\tt Brother or sister}. \\
WKHP & Work-hours per week & The individual's usual number of hours worked per week is: {\color{greyish}\tt 40 hours}. \\
SEX & Sex & The individual's sex is: {\color{greyish}\tt Female}. \\
RAC1P & Race & The individual's race is: {\color{greyish}\tt Black or African American}. \\
PINCP & Total yearly income & The individual's total yearly income is: {\color{greyish}\tt \$75,000}. \\
%% ACS Public Coverage
CIT & Citizenship status & The individual's citizenship status is: {\color{greyish}\tt Naturalized US citizen}. \\
DIS & Disability status & The individual {\color{greyish}\tt has a disability}. \\
ESP & Employment status of parents & The individual is {\color{greyish}\tt living with two parents:~both parents in labor force}. \\
MIG & Mobility (lived here 1 year ago) & The individual {\color{greyish}\tt lived in the same house 1 year ago.} \\
MIL & Military service & The individual {\color{greyish}\tt was on active duty in the past, but not currently}. \\
PUBCOV & Public health coverage & The individual {\color{greyish}\tt is covered by public health insurance}. \\
ANC & Ancestry & The individual has {\color{greyish}\tt single ancestry}. \\
NATIVITY & Nativity & The individual is {\color{greyish}\tt foreign born}.\\
DEAR & Hearing & The individual {\color{greyish}\tt has hearing difficulty}. \\
DEYE & Vision & The individual {\color{greyish}\tt does not have vision difficulty}. \\
DREM & Cognition & The individual {\color{greyish}\tt does not have cognitive difficulties}. \\
ESR & Employment status \#2 & The individual is {\color{greyish}\tt not in the labor force}. \\
ST & State & The individual lives in {\color{greyish}\tt California}. \\
FER & Parenthood (1 year) & The individual {\color{greyish}\tt gave birth to a child within the past 12 months}. \\
%%% ACS Mobility
% GCL & Grandparents & The individual's %% TODO!
%%% ACS Travel Time
JWMNP & Commute time & The individual takes {\color{greyish}\tt 45 minutes travelling to work every day}. \\
JWTR & Means of transport & The individual's means of transport to work is {\color{greyish}\tt a bicycle}. \\
POVPIP & Income-to-poverty ratio & The individual's income to poverty ratio is {\color{greyish}\tt 150\%}. \\
\\
\end{tabular}
\caption{Description of all column-to-text mappings implemented for ACS features. The variable part of each example is shown in  {\color{greyish}\tt typeset grey font}. Details on each possible categorical value for each feature are available in the ACS PUMS data dictionary.\textsuperscript{\ref{foot:codebook}}}
\label{tab:col_to_text}
\end{table}
\end{landscape}

\end{document}